\documentclass{article}

\usepackage{amsmath,amsfonts,bm}

\def\eqref#1{equation~\ref{#1}}
\def\Eqref#1{Equation~\ref{#1}}

\def\1{\bm{1}}

\def\mW{{\mathbf{W}}}
\def\mX{{\mathbf{X}}}

\def\mmu{{\bm{\mu}}}
\def\mSigma{{\bm{\Sigma}}}

\def\mtheta{{\bm{\theta}}}

\def\vc{{\bm{c}}}

\def\vh{{\bm{h}}}

\def\vo{{\bm{o}}}

\def\vw{{\bm{w}}}
\def\vx{{\bm{x}}}
\def\vy{{\bm{y}}}
\def\vz{{\bm{z}}}

\def\mW{{\bm{W}}}
\def\mX{{\bm{X}}}

\def\mSigma{{\bm{\Sigma}}}

\DeclareMathAlphabet{\mathsfit}{\encodingdefault}{\sfdefault}{m}{sl}
\SetMathAlphabet{\mathsfit}{bold}{\encodingdefault}{\sfdefault}{bx}{n}

\def\gA{{\mathcal{A}}}

\def\gD{{\mathcal{D}}}

\def\gN{{\mathcal{N}}}

\def\gT{{\mathcal{T}}}
\def\gU{{\mathcal{U}}}

\def\gW{{\mathcal{W}}}
\def\gX{{\mathcal{X}}}

\def\gZ{{\mathcal{Z}}}

\def\sE{{\mathbb{E}}}

\def\sK{{\mathbb{K}}}
\def\sL{{\mathbb{L}}}

\def\sR{{\mathbb{R}}}

\DeclareMathOperator*{\argmin}{arg\,min}

\usepackage{microtype}
\usepackage{graphicx}
\usepackage{xcolor}
\usepackage{booktabs} 
\usepackage{multicol}
\usepackage{multirow}
\usepackage{enumitem}
\usepackage{subcaption}
\usepackage{hyperref}
\usepackage{wrapfig}

\usepackage{amsmath}
\usepackage{amssymb}
\usepackage{mathtools}
\usepackage{amsthm}

\usepackage[textsize=tiny]{todonotes}

\newcommand{\sstd}[1]{\textcolor{black}{\tiny{$\pm #1$}}}
\newcommand{\std}[1]{}
\newcommand{\highlight}[1]{\colorbox{blue!10}{#1}}

\usepackage{microtype}
\usepackage{graphicx}
\usepackage{booktabs} 

\usepackage{hyperref}

\usepackage[preprint]{icml2025}

\usepackage{amsmath}
\usepackage{amssymb}
\usepackage{mathtools}
\usepackage{amsthm}

\usepackage[nameinlink,capitalize,noabbrev]{cleveref}

\newcommand{\ie}{\textit{i.e.}}
\newcommand{\eg}{\textit{e.g.}}

\theoremstyle{plain}

\theoremstyle{definition}

\theoremstyle{remark}

\usepackage[textsize=tiny]{todonotes}
\expandafter\def\expandafter\normalsize\expandafter{%
}

\newcommand{\norm}[1]{\left\lVert#1\right\rVert}

\icmltitlerunning{Amortized In-Context Bayesian Posterior Estimation}

\begin{document}

\twocolumn[
\icmltitle{Amortized In-Context Bayesian Posterior Estimation}



\icmlsetsymbol{equal}{*}

\begin{icmlauthorlist}
\icmlauthor{Sarthak Mittal}{udem,mila}
\icmlauthor{Niels Leif Bracher}{rpi}
\icmlauthor{Guillaume Lajoie}{udem,mila}
\icmlauthor{Priyank Jaini}{gdm}
\icmlauthor{Marcus Brubaker}{gdm,york,vector}
\end{icmlauthorlist}

\icmlaffiliation{udem}{Universit\'e de Montreal}
\icmlaffiliation{mila}{Mila}
\icmlaffiliation{york}{York University}
\icmlaffiliation{gdm}{Google DeepMind}
\icmlaffiliation{vector}{Vector Institute}
\icmlaffiliation{rpi}{Rensselaer Polytechnic Institute}

\icmlcorrespondingauthor{Sarthak Mittal}{sarthmit@gmail.com}

\icmlkeywords{Machine Learning, ICML}

\vskip 0.3in
]



\printAffiliationsAndNotice{}

\begin{abstract}
Bayesian inference provides a natural way of incorporating prior beliefs and assigning a probability measure to the space of hypotheses. Current solutions rely on iterative routines like Markov Chain Monte Carlo (MCMC) sampling and Variational Inference (VI), which need to be re-run whenever new observations are available. Amortization, through conditional estimation, is a viable strategy to alleviate such difficulties and has been the guiding principle behind simulation-based inference, neural processes and in-context methods using pre-trained models. In this work, we conduct a thorough comparative analysis of amortized in-context Bayesian posterior estimation methods from the lens of different optimization objectives and architectural choices. Such methods train an amortized estimator to perform posterior parameter inference by conditioning on a set of data examples passed as context to a sequence model such as a transformer. In contrast to language models, we leverage permutation invariant architectures as the true posterior is invariant to the ordering of context examples. Our empirical study includes generalization to out-of-distribution tasks, cases where the assumed underlying model is misspecified, and transfer from simulated to real problems. Subsequently, it highlights the superiority of the reverse KL estimator for predictive problems, especially when combined with the transformer architecture and normalizing flows.
\end{abstract}

\vspace{-9mm}
\section{Introduction}
\vspace{-1mm}
\label{sec:intro}
\looseness=-1
Bayesian analysis of data has become increasingly popular and is widely used in numerous scientific disciplines.
In politics, predictive models based on public polling and other factors play a crucial role in the discourse around the state of a campaign.
Throughout the COVID-19 pandemic, models that estimate the infectiousness of the virus, the efficacy of public health measures, and the future course of the pandemic became critical to government planning and the public's understanding of the pandemic~\citep{Cooper2020-covidmodel}. In cryogenic electron microscopy (cryo-EM), the posterior over an unknown 3D atomic-resolution molecular structure is explored given image observations~\citep{Glaeser2021-cryoem}. 

\looseness=-1
While recent years have made such methods more accessible \citep{bingham2019pyro,carpenter2017stan,StatSoftReview}, they still remain computationally burdensome.
Further, in practical contexts where new observations are continuously available, the analysis must be re-run every time new data becomes available, e.g., when new case counts become available, previous measurements are corrected, or when applied to different geographic regions.
As a result practitioners adopt approximations~\citep{welling2011bayesian,gelfand2000gibbs,brooks1998markov}, simplify their models~\citep{hoffman2013stochastic,blei2017variational} or reduce the frequency with which they perform their analyses.

\looseness=-1
A common thread is that the probabilistic model defining the relationship between its parameters and the observations is fixed. Poll aggregation models use hierarchical time series models~\citep{Athanasopoulos-hierarchicalts,chen2023-pollaggregation}, infectious diseases are studied using variants on compartment models~\citep{Tang2020-diseasemodels}, and cryo-EM uses a linear image formation model~\citep{Glaeser2021-cryoem}. This makes these applications ideal candidates for amortized inference~\citep{morris2013recognition,paige2016inference,kingma2013auto,rezende2014stochastic,stuhlmuller2013learning}.

\looseness=-1
Multiple approaches leverage neural networks to learn functions that map an observed \emph{dataset} directly to a posterior distribution~\citep{garnelo2018neural,cranmer2020sbireview} or model the posterior predictive directly~\citep{garnelo2018conditional,muller2021transformers,garg2022can,hollmann2022tabpfn}. They sidestep the need for iterative procedures, e.g., Markov chain Monte Carlo (MCMC) sampling~\citep{gelfand2000gibbs,hoffman2014no} or standard variational inference (VI) and efficiently handle permutation invariance stemming from \textit{iid} observations using Transformers and DeepSets~\citep{Zaheer2017deepsets,vaswani2017attention,lee2019set}. If learned properly, this mapping allows generalization to new datasets passed in context in zero-shot.

However, analysis into evaluating different in-context posterior estimation objectives is currently lacking. The goal of such estimators is to model the posterior distribution by leveraging observations in context as opposed to invoking iterative estimation procedures again from scratch. We provide a rigorous analysis into different training objectives, i.e. forward and reverse KL objectives, where the former is equivalent to neural posterior estimation in simulation-based inference \citep{cranmer2020sbireview} and the latter has connections to neural processes \citep{garnelo2018neural}. However, NP only model the posterior over some unstructured latent variable (akin to \citet{kingma2013auto,rezende2014stochastic}) with the objective being a proxy to maximum likelihood while we are interested in a fully Bayesian treatment of all parameters defining the likelihood.

Our benchmark considers a wide variety of probabilistic models and evaluates different design choices in inferring the posterior over their parameters. We look at different permutation invariant architectures, parametrizations for the approximate density, as well as training objectives. Our evaluation criteria tests for both in-distribution (ID) and out-of-distribution (OoD) generalization, and relies on a simple masking procedure to amortize posterior estimation over datasets with a variable number of features, inching closer towards a generalist in-context Bayesian learner $-$ as evidenced by its generalization capabilities on real-world tasks zero-shot through only pre-training on synthetic data.

Generally, real-world datasets do not exactly follow standard models, e.g., while practitioners often rely on linear models, data rarely follows them exactly. We further evaluate the estimators on tasks where the assumed probabilistic model is incorrect (misspecification), or where we only have access to samples but not underlying parameters, which is a common paradigm in most machine learning tasks.
%
Our detailed experiments provide clear insights into the architectural choices that lead to better amortized posterior estimation, through the lens of both predictive and sample-based metrics. Our contributions include
\begin{itemize}[topsep=0pt,parsep=1pt,partopsep=0pt,leftmargin=3mm] 
  \setlength\itemsep{0pt}
    \item Providing a general framework for in-context Bayesian posterior estimation with different training objectives.
    \item Benchmarking various design choices like architectural backbones, parametrizations of approximate density and training objectives through extensive ablations.
    \item Evaluating the ability of estimators to generalize OoD when the modeling assumption is different from the underlying true model class (misspecification), especially to real-world tasks when trained only on synthetic data.
\end{itemize}
\begin{figure*}
     \centering
     \captionsetup[subfigure]{font=tiny}
     \begin{subfigure}[b]{0.16\textwidth}
         \centering
         \includegraphics[width=\textwidth]{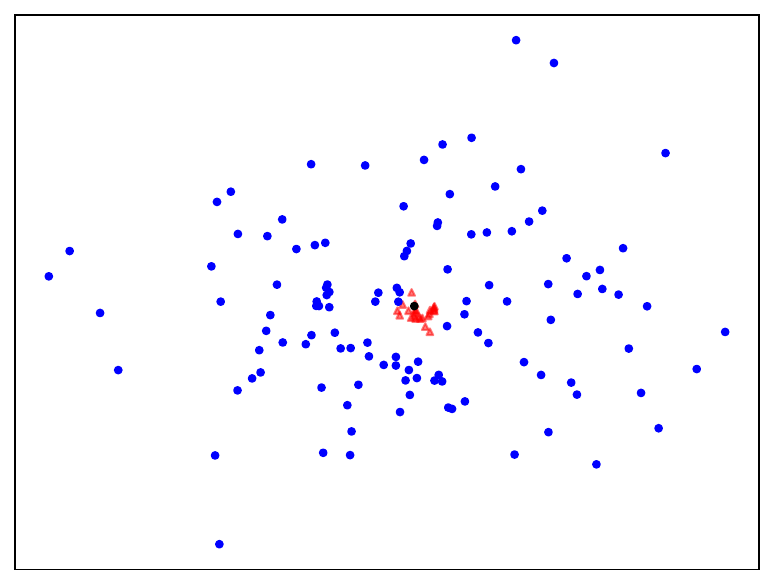}
         \caption*{Mean of Gaussian}
         \label{fig:gaussian}
     \end{subfigure}
     \hfill
     \begin{subfigure}[b]{0.16\textwidth}
         \centering
         \includegraphics[width=\textwidth]{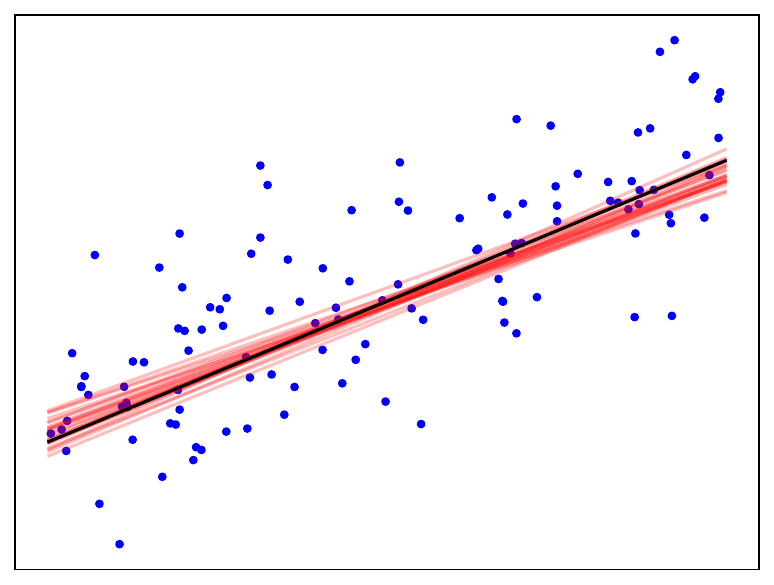}
         \caption*{Linear Regression}
         \label{fig:linear_regression}
     \end{subfigure}
     \hfill
     \begin{subfigure}[b]{0.16\textwidth}
         \centering
         \includegraphics[width=\textwidth]{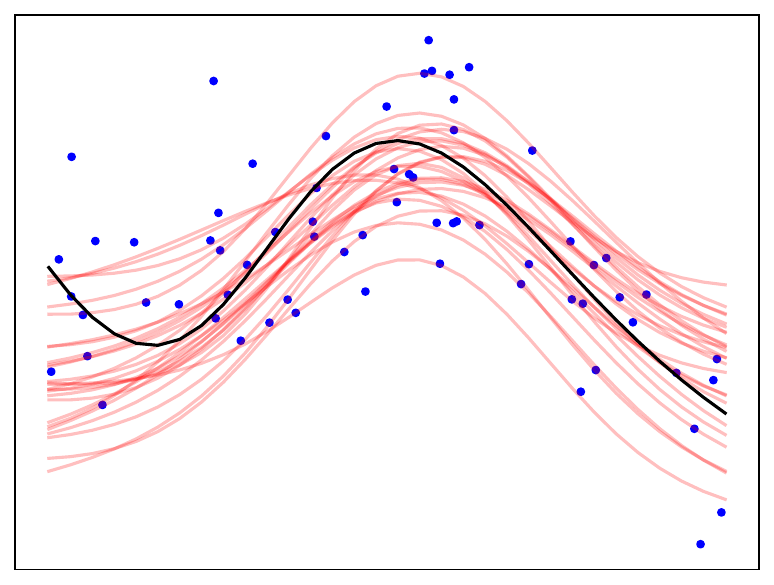}
         \caption*{Nonlinear Regression}
         \label{nonlinear_regression}
     \end{subfigure}
     \hfill
     \begin{subfigure}[b]{0.16\textwidth}
         \centering
         \includegraphics[width=\textwidth]{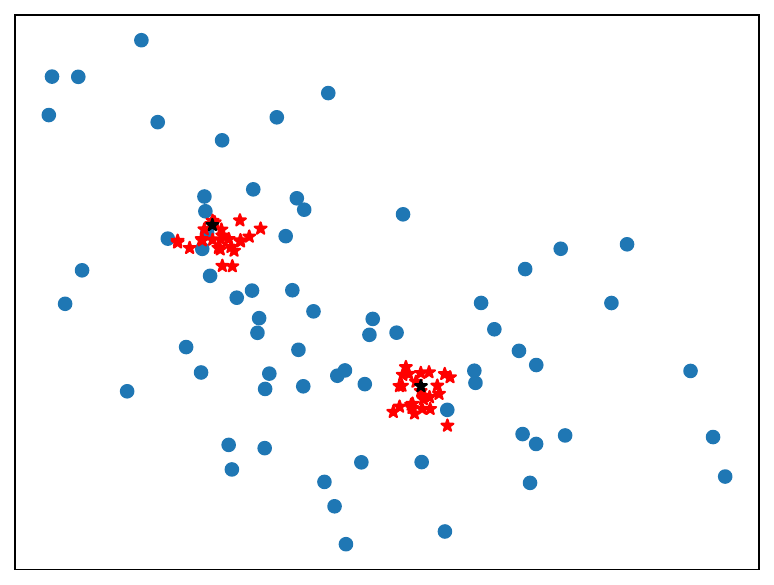}
         \caption*{Gaussian Mixture}
         \label{fig:gmm}
     \end{subfigure}
     \hfill
     \begin{subfigure}[b]{0.16\textwidth}
         \centering
         \includegraphics[width=\textwidth]{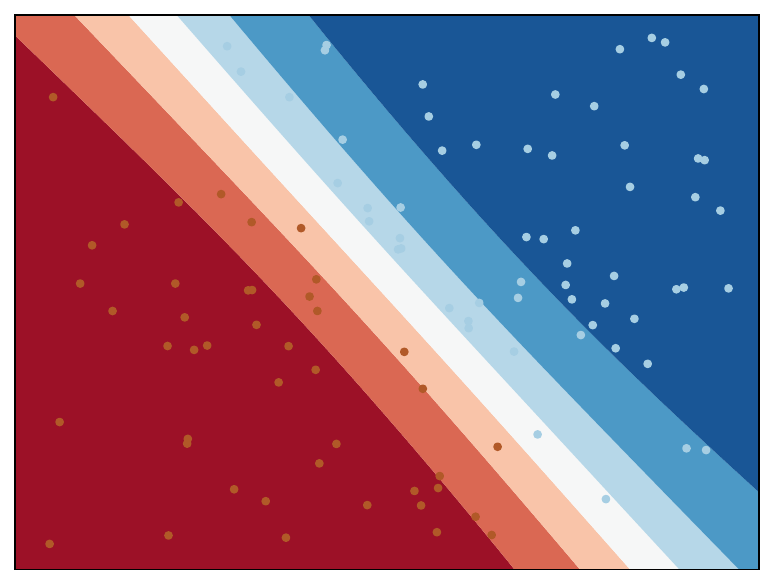}
         \caption*{Linear Classification}
         \label{fig:linear_classification}
     \end{subfigure}
     \hfill
     \begin{subfigure}[b]{0.16\textwidth}
         \centering
         \includegraphics[width=\textwidth]{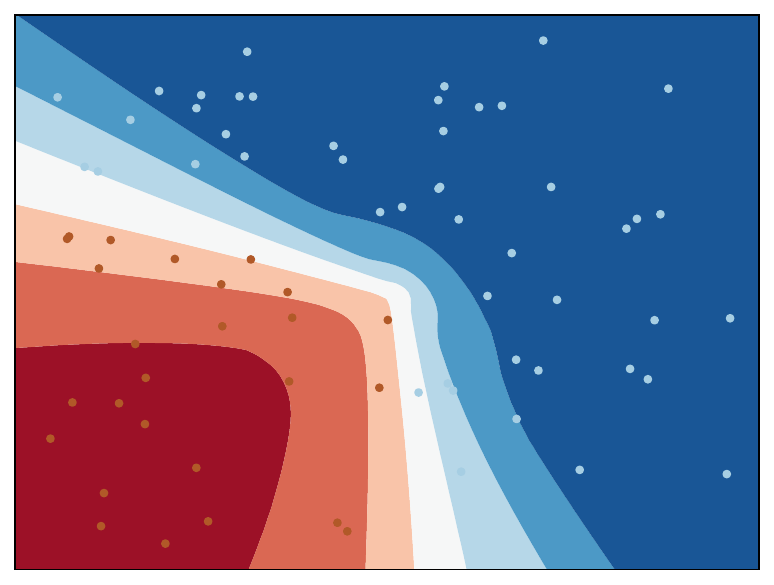}
         \caption*{Nonlinear Classification}
         \label{fig:nonlinear_classification}
     \end{subfigure}
    \vspace{-4mm}
    \caption{\textbf{Amortized Bayesian Posterior Estimation:} Illustration of predictions from the reverse KL in-context estimator. Model predictions, true predictions and sample points are shown in red, black and blue respectively. Additionally for classification, we label sample points with their ground-truth class, and draw the decision boundary according to the model.}
    \vspace{-6mm}
    \label{fig:fixed_dim}
\end{figure*}

\vspace{-3mm}
\section{Background}
\label{sec:prelim}
\vspace{-2mm}
\looseness=-1
We first cover some of the important preliminaries below.

\looseness=-1
\textbf{Bayesian Inference}. Let $\vx \in \mathbb{R}^d$ denote the outcome of an experiment observed through a set of independent and identically distributed (\textit{iid}) samples $\gD := \{\vx_1, ..., \vx_N\} \subseteq \mathbb{R}^d$. Given these observations, we are interested in either quantifying the certainty of or generating potential future observations $\vx_*$. Bayesian Inference provides a natural methodology of quantifying $p(\vx_* | \gD)$ by prescribing a space of hypotheses $\mtheta \in \mathbb{R}^k$ and a \textit{prior} belief $p(\mtheta)$ over it. These hypotheses define the \textit{likelihood} of observing an outcome, i.e., $p(\vx | \mtheta)$. The likelihood and prior are then combined through Bayes rule to infer the \textit{posterior} $p(\mtheta | \gD)$, through which the quantity of interest can then be easily expressed as 
\begin{align}
    p(\vx_* | \gD) = \int_\mtheta p(\vx_* | \mtheta) p(\mtheta | \gD) d\mtheta
\end{align}
This poses two challenges: (a) the \textit{posterior}, often a quantity of interest in itself, is not known, and (b) the integration can be intractable which is often resolved through Monte Carlo estimation
\begin{align}
    p(\vx_* | \gD) = \mathbb{E}_{\mtheta | \gD} \left[p(\vx_* | \mtheta)\right]
    \approx \frac{1}{M} \sum_{m=1}^M p(\vx_* | \mtheta^{(m)})
\end{align}
where $\mtheta^{(m)} \sim p(\mtheta | \gD)$. The quantity $p(\mtheta | \gD)$ can be obtained through an application of Bayes rule
\begin{align}
    \label{eq:bayes_rule}
    p(\mtheta | \gD) = \frac{p(\gD | \mtheta)\;p(\mtheta)}{p(\gD)} 
    = \frac{p(\mtheta)}{p(\gD)} \prod_{n=1}^N p(\vx_n | \mtheta)
\end{align}
Given the form of the \textit{likelihood} and \textit{prior}, the above distribution is often difficult to sample from, especially with the added complexity of the marginal $p(\gD) = \int_\mtheta p(\gD | \mtheta)\, p(\mtheta)$ being intractable. Additionally, the posterior itself is often of interest on its own, especially in cases where $\mtheta$ is interpretable, \eg if we model the bias of a coin based on multiple tosses. We refer the readers to~\citet{bishop2006pattern} for additional applications of Bayesian Inference.

\textbf{Approximate Bayesian Inference}. To bypass the intractability of the posterior distribution, or at least the difficulty to sample from it, approximate methods are used. 

Sampling based methods provide ways of sampling from the true posterior distribution based on easy access to an unnormalized density function, e.g. rejection sampling. More advanced methods like  MCMC construct a chain of updates $\mtheta_1, \mtheta_2, \ldots$ such that asymptotically the samples converge to samples from the true posterior. Such sampling methods rely on transition kernels $\gT(\mtheta_{t+1} | \mtheta_t)$ and often some acceptance criteria $\gA(\mtheta_{t+1}, \mtheta_t)$, a key example of which is the Metropolis-Hastings algorithm. We refer the readers to~\citep{hoffman2014no,welling2011bayesian} for a detailed analysis into different MCMC methods like Langevin and Hamiltonian Monte Carlo which rely on gradient of the log density as additional signal for better convergence.

In contrast, another class of methods approximate the true posterior with a parametric family $q_\varphi(\mtheta)$ and convert the estimation problem into the following optimization problem
\begin{align}
    \varphi^* = \arg\min_\varphi \mathbb{D}\left(p(\cdot | \gD), q_\varphi(\cdot)\right)
\end{align} 
where $\mathbb{D}$ is a notion of divergence between two distributions. Once the optimal parameters $\varphi^*$ are obtained, $q_{\varphi^*}$ can be used to substitute the true posterior wherever needed. The above optimization procedure finds a member in the family of variational distributions $\{q_\varphi\}_\varphi$ that is closest to the true posterior under $\mathbb{D}$. An example of this is Variational Inference, where the reverse KL divergence is used
\begin{align}
\label{eq:rkl}
\mathbb{D}_{\text{R-}\mathbb{KL}}\left(p(\cdot | \gD), q_\varphi(\cdot)\right) &= \mathbb{E}_{\mtheta \sim q_\varphi(\cdot)} \left[\log \frac{q_\varphi(\mtheta)}{p(\mtheta | \gD)}\right]
\end{align}
which is equivalent to optimizing the well known Evidence Lower-Bound (ELBO)~\citep{Gelman2013-bayesdata}
\begin{align}
    \varphi^* &= \arg\max_\varphi \sE_{\mtheta \sim q_\varphi(\cdot)}\left[\log \frac{p(\gD, \mtheta)}{q_\varphi(\mtheta)}\right]
\end{align}
Another example is Expected Propagation \citep[EP;][]{minka2013expectation} which relies on the forward KL divergence
\begin{align}
\label{eq:fkl}
\mathbb{D}_{\text{F-}\mathbb{KL}}\left(p(\cdot | \gD), q_\varphi(\cdot)\right) &= \mathbb{E}_{\mtheta \sim p(\cdot | \gD)} \left[\log \frac{p(\mtheta | \gD)}{q_\varphi(\mtheta)}\right]
\end{align}
Once the optimal parameters $\varphi^*$ are obtained, the \textit{posterior predictive} distribution $p(\vx_* | \gD)$ can be approximated as
\begin{align}
    p(\vx_* | \gD) \approx \mathbb{E}_{q_{\varphi^*}(\mtheta)}\left[p(\vx_* | \mtheta)\right]
\end{align}
\looseness=-1
The family of distributions $q_\varphi$ is chosen such that it is easy to sample from. Typical choices include independent multivariate Gaussian distribution (mean-field approximation) or normalizing flows~\citep{rezende2015variational,papamakarios2021normalizing,freia}.

\begin{table*}[t]
    \centering
    \small
    \setlength{\tabcolsep}{2pt}
    \begin{tabular}{@{}lcr c c c cc c cc}
        \toprule
         &  &  & \multicolumn{5}{c}{\textit{$L_2$ Loss} ($\downarrow$)} & \multicolumn{3}{c}{\textit{Accuracy} ($\uparrow$)}\\
         \cmidrule(lr){4-8}\cmidrule(lr){9-11}
        \textbf{Objective} & $q_\varphi$ & \textbf{Model} & \multicolumn{1}{c}{\textbf{Gaussian}} & \multicolumn{1}{c}{\textbf{GMM}} & \multicolumn{1}{c}{\textbf{LR}} & \multicolumn{2}{c}{\textbf{NLR}} & \multicolumn{1}{c}{\textbf{LC}} & \multicolumn{2}{c}{\textbf{NLC}} \\
        \cmidrule(lr){4-4}\cmidrule(lr){5-5}\cmidrule(lr){6-6}\cmidrule(lr){7-8}\cmidrule(lr){9-9}\cmidrule(lr){10-11}
        & & & \textit{100D} & \textit{5D 2 cl} & \textit{100D} & 
        \textit{1D} & \textit{25D} & \textit{100D} &  \textit{2D} & \textit{25D}\\
        \midrule

\multirow{4}{*}{Baseline} & - & Random & $301.06$\sstd{$0.35$} & $5.00$\sstd{$0.04$} & $202.6$\sstd{$0.3$} & $65.94$\sstd{$0.91$} & $831.6$\sstd{$8.7$} & $50.0$\sstd{$0.0$} & $50.3$\sstd{$0.6$} & $50.0$\sstd{$0.3$} \\
& - & Optimization & $101.24$\sstd{$0.00$} & $0.42$\sstd{$0.00$} & $25.1$\sstd{$0.0$} & $0.36$\sstd{$0.00$} & $104.0$\sstd{$0.1$} & $70.3$\sstd{$0.0$} & $96.9$\sstd{$0.0$} & $77.9$\sstd{$0.0$} \\
& - & Langevin & $102.35$\sstd{$0.03$} & $0.45$\sstd{$0.01$} & $23.3$\sstd{$0.7$} & $0.31$\sstd{$0.00$} & $132.4$\sstd{$1.0$} & $65.1$\sstd{$0.4$} & $96.0$\sstd{$0.3$} & $73.2$\sstd{$0.3$} \\
& - & HMC & $102.41$\sstd{$0.03$} & $0.48$\sstd{$0.01$} & $18.7$\sstd{$0.2$} & $0.37$\sstd{$0.00$} & $98.1$\sstd{$0.7$} & $62.1$\sstd{$0.2$} & $91.8$\sstd{$0.2$} & $70.4$\sstd{$0.1$} \\
\cmidrule{2-11}

\multirow{3}{*}{Fwd-KL} & \multirow{6}{*}{\rotatebox[origin=c]{90}{Gaussian}} & GRU &$102.64$\sstd{$0.01$} & $2.43$\sstd{$0.03$} & $124.8$\sstd{$0.1$} & $49.33$\sstd{$0.95$} & $671.6$\sstd{$10.5$} & $59.7$\sstd{$0.1$} & $59.5$\sstd{$0.4$} & $56.9$\sstd{$0.3$} \\
& & DeepSets &$103.22$\sstd{$0.05$} & $2.44$\sstd{$0.04$} & $123.1$\sstd{$1.1$} & $49.86$\sstd{$0.98$} & $684.9$\sstd{$2.6$} & $50.0$\sstd{$0.1$} & $59.4$\sstd{$0.2$} & $56.8$\sstd{$0.2$} \\
& & Transformer &$102.78$\sstd{$0.00$} & $2.50$\sstd{$0.03$} & $45.9$\sstd{$1.3$} & $49.68$\sstd{$0.94$} & $680.9$\sstd{$5.8$} & $63.0$\sstd{$0.1$} & $59.6$\sstd{$0.4$} & $57.1$\sstd{$0.4$} \\
\cmidrule{3-11}

\multirow{3}{*}{Fwd-KL} & & GRU &$102.51$\sstd{$0.01$} & \highlight{$0.47$\sstd{$0.01$}} & $60.2$\sstd{$0.9$} & $0.43$\sstd{$0.00$} & $106.0$\sstd{$0.6$} & $63.5$\sstd{$0.3$} & $92.4$\sstd{$0.2$} & $72.5$\sstd{$0.0$} \\
& & DeepSets &$102.60$\sstd{$0.04$} & $0.50$\sstd{$0.02$} & $62.8$\sstd{$0.6$} & $0.43$\sstd{$0.00$} & $125.9$\sstd{$0.8$} & $60.9$\sstd{$0.3$} & $92.5$\sstd{$0.1$} & $59.8$\sstd{$0.3$} \\
& & Transformer &$102.54$\sstd{$0.03$} & \highlight{$0.49$\sstd{$0.02$}} & \highlight{$28.7$\sstd{$0.3$}} & $0.42$\sstd{$0.01$} & \highlight{$102.3$\sstd{$1.8$}} & $68.2$\sstd{$0.0$} & $92.6$\sstd{$0.4$} & \highlight{$75.2$\sstd{$0.1$}} \\
\cmidrule{2-11}

\multirow{3}{*}{Fwd-KL} & \multirow{6}{*}{\rotatebox[origin=c]{90}{Flow}} & GRU &$102.66$\sstd{$0.02$} & $0.67$\sstd{$0.09$} & $119.1$\sstd{$0.2$} & $15.78$\sstd{$0.21$} & $539.0$\sstd{$4.3$} & $59.9$\sstd{$0.2$} & $76.9$\sstd{$0.3$} & $58.3$\sstd{$0.0$} \\
& & DeepSets &$103.34$\sstd{$0.03$} & $0.65$\sstd{$0.08$} & $125.7$\sstd{$3.7$} & $15.05$\sstd{$0.12$} & $548.5$\sstd{$3.3$} & $50.1$\sstd{$0.0$} & $72.3$\sstd{$1.8$} & $58.1$\sstd{$0.1$} \\
& & Transformer &$102.77$\sstd{$0.02$} & $0.62$\sstd{$0.07$} & $43.3$\sstd{$2.7$} & $16.11$\sstd{$0.31$} & $539.3$\sstd{$4.3$} & $64.3$\sstd{$0.1$} & $77.3$\sstd{$0.2$} & $58.3$\sstd{$0.1$} \\
\cmidrule{3-11}

\multirow{3}{*}{Rev-KL} & & GRU &\highlight{$102.49$\sstd{$0.01$}} & \highlight{$0.47$\sstd{$0.00$}} & $61.3$\sstd{$1.0$} & $0.41$\sstd{$0.01$} & $106.0$\sstd{$0.4$} & $64.7$\sstd{$0.2$} & $93.4$\sstd{$0.1$} & $72.0$\sstd{$0.5$} \\
& & DeepSets &$102.67$\sstd{$0.05$} & $0.52$\sstd{$0.01$} & $76.4$\sstd{$2.0$} & $0.40$\sstd{$0.00$} & $128.2$\sstd{$1.5$} & $58.4$\sstd{$0.8$} & $93.3$\sstd{$0.2$} & $60.9$\sstd{$0.2$} \\
& & Transformer &\highlight{$102.53$\sstd{$0.05$}} & \highlight{$0.47$\sstd{$0.01$}} & \highlight{$29.4$\sstd{$1.6$}} & \highlight{$0.39$\sstd{$0.00$}} & \highlight{$102.6$\sstd{$0.9$}} & \highlight{$68.7$\sstd{$0.1$}} & \highlight{$93.6$\sstd{$0.1$}} & \highlight{$75.0$\sstd{$0.5$}} \\
\bottomrule
    \end{tabular}
    \vspace{-2mm}
    \caption{\textbf{Fixed-dimensional In-Context Posterior Estimation} for estimating the mean of a Gaussian (Gaussian), means of a Gaussian mixture (GMM), parameters of (non-)linear regression (NLR/LR) and (non-)linear binary classification (NLC/LC). We ablate over different architectures and density parametrizations and use the expected predictive $L_2$ loss and accuracy as the metrics. 
    }
    \vspace{-6mm}
    \label{tab:fixed_dim}
\end{table*}
\textbf{Estimators and Amortization}. A core benefit of deep learning is its ability to generalize well. Amortized approaches leverage this ability by training conditional models to solve a family of problems in an efficient and scalable manner, as opposed to independently solving each problem. For \eg, the encoder in Variational Autoencoders~\citep[VAEs;][]{kingma2013auto,rezende2014stochastic} is tasked with estimating the posterior distribution $p(\vz | \vx_i)$ for each $\vx_i \in \gD$. Here, $p(\vz)$ is the standard normal prior and $p(\vx_i | \vz)$ is the decoder defining the trainable likelihood. Instead of separate optimization problems $q_{\varphi_i^*}(\vz)$ for each posterior $p(\vz | \vx_i)$, VAEs rely on amortization to train a shared network $q_\varphi(\vz | \vx)$, where $\varphi$ now represents the parameters of a neural network and takes $\vx$ explicitly as input, allowing zero-shot generalization to new $\vx_*$ at inference. 

Amortization plays a key role in multiple domains of machine learning, beyond VAEs. For \eg, score-based diffusion models \citep{song2020score} amortize training of a time-conditioned score model, while Neural Processes (NPs) \citep{garnelo2018conditional,garnelo2018neural} and neural posterior estimation in Simulation-Based Inference (SBI) \citep{cranmer2020sbireview} amortize dataset-conditioned VAE-styled encoders under $\mathbb{D}_{\text{R-}\mathbb{KL}}$ and $\mathbb{D}_{\text{F-}\mathbb{KL}}$ respectively. Even further, in-context learning (ICL)~\citep{von2023transformers,muller2021transformers} can also be seen as amortizing the posterior predictive distribution $p(y_* | \vx_*, \gD)$ based on context examples $\gD$ as an emergent phenomena owing to the shared modality of language tying different tasks, where $y_*$ defines the label.

\begin{table*}[t]
    \centering
    \small
    \setlength{\tabcolsep}{2pt}
    \begin{tabular}{@{}lcr  c  c  c  cc  c  cc}
        \toprule
         &  &  & \multicolumn{5}{c}{\textit{$L_2$ Loss} ($\downarrow$)} & \multicolumn{3}{c}{\textit{Accuracy} ($\uparrow$)}\\
         \cmidrule(lr){4-8}\cmidrule(lr){9-11}
        \textbf{Objective} & $q_\varphi$ & \textbf{Model} & \multicolumn{1}{c}{\textbf{Gaussian}} & \multicolumn{1}{c}{\textbf{GMM}} & \multicolumn{1}{c}{\textbf{LR}} & \multicolumn{2}{c}{\textbf{NLR}} & \multicolumn{1}{c}{\textbf{LC}} & \multicolumn{2}{c}{\textbf{NLC}} \\
        \cmidrule(lr){4-4}\cmidrule(lr){5-5}\cmidrule(lr){6-6}\cmidrule(lr){7-8}\cmidrule(lr){9-9}\cmidrule(lr){10-11}
        & & & \textit{100D} & \textit{5D 2 cl} & \textit{100D} & 
        \textit{1D} & \textit{50D} & \textit{100D} &  \textit{2D} & \textit{50D}\\
        \midrule

\multirow{4}{*}{Baseline} & - & Random & $298.24$\sstd{$0.23$} & $4.66$\sstd{$0.03$} & $200.8$\sstd{$0.6$} & $73.01$\sstd{$0.17$} & $1704.3$\sstd{$9.3$} & $50.0$\sstd{$0.1$} & $50.0$\sstd{$0.3$} & $49.9$\sstd{$0.3$} \\
& - & Optimization & $100.88$\sstd{$0.00$} & $0.43$\sstd{$0.00$} & $20.1$\sstd{$0.0$} & $0.36$\sstd{$0.00$} & $309.2$\sstd{$0.2$} & $71.2$\sstd{$0.0$} & $96.8$\sstd{$0.0$} & $76.1$\sstd{$0.0$} \\
& - & Langevin & $101.92$\sstd{$0.04$} & $0.44$\sstd{$0.00$} & $21.8$\sstd{$1.0$} & $0.31$\sstd{$0.00$} & \textsc{N/A} & $65.5$\sstd{$0.5$} & $96.1$\sstd{$0.0$} & $70.1$\sstd{$0.2$} \\
& - & HMC & $102.01$\sstd{$0.01$} & $0.46$\sstd{$0.01$} & $17.8$\sstd{$0.1$} & $0.38$\sstd{$0.01$} & $303.9$\sstd{$2.5$} & $62.6$\sstd{$0.2$} & $91.7$\sstd{$0.2$} & $68.0$\sstd{$0.4$} \\
\cmidrule{2-11}

\multirow{3}{*}{Fwd-KL} & \multirow{6}{*}{\rotatebox[origin=c]{90}{Gaussian}} & GRU &$133.22$\sstd{$0.58$} & $2.36$\sstd{$0.02$} & $139.4$\sstd{$1.0$} & $51.45$\sstd{$0.03$} & $1346.5$\sstd{$6.8$} & $57.9$\sstd{$0.2$} & $59.6$\sstd{$0.2$} & $58.6$\sstd{$0.2$} \\
& & DeepSets &$129.69$\sstd{$0.74$} & $2.35$\sstd{$0.02$} & $149.8$\sstd{$0.8$} & $51.90$\sstd{$1.54$} & $1357.5$\sstd{$5.3$} & $50.8$\sstd{$0.1$} & $49.9$\sstd{$0.3$} & $49.9$\sstd{$0.3$} \\
& & Transformer &$108.98$\sstd{$0.10$} & $2.40$\sstd{$0.02$} & $64.3$\sstd{$3.7$} & $50.81$\sstd{$0.53$} & $1319.9$\sstd{$12.2$} & $62.4$\sstd{$0.0$} & $59.9$\sstd{$0.2$} & $58.8$\sstd{$0.2$} \\
\cmidrule{3-11}

\multirow{3}{*}{Rev-KL} & & GRU &$105.14$\sstd{$0.10$} & \highlight{$0.46$\sstd{$0.01$}} & $62.6$\sstd{$0.1$} & $2.31$\sstd{$0.13$} & $316.3$\sstd{$6.2$} & $63.4$\sstd{$0.2$} & $88.8$\sstd{$0.5$} & $68.4$\sstd{$0.3$} \\
& & DeepSets &$105.06$\sstd{$0.21$} & $0.48$\sstd{$0.02$} & $64.1$\sstd{$0.2$} & $0.98$\sstd{$0.12$} & $451.9$\sstd{$2.8$} & $61.3$\sstd{$0.1$} & \highlight{$91.0$\sstd{$0.5$}} & $61.7$\sstd{$0.1$} \\
& & Transformer & \highlight{$104.71$\sstd{$0.12$}} & \highlight{$0.47$\sstd{$0.01$}} & \highlight{$32.0$\sstd{$0.5$}} & \highlight{$0.81$\sstd{$0.02$}} & $278.3$\sstd{$1.1$} & $67.7$\sstd{$0.1$} & $90.0$\sstd{$0.2$} & \highlight{$73.7$\sstd{$0.3$}} \\
\cmidrule{2-11}

\multirow{3}{*}{Fwd-KL} & \multirow{6}{*}{\rotatebox[origin=c]{90}{Flow}} & GRU &$125.84$\sstd{$1.98$} & $0.60$\sstd{$0.07$} & $138.3$\sstd{$1.0$} & $38.41$\sstd{$0.36$} & $1097.4$\sstd{$9.5$} & $58.0$\sstd{$0.1$} & $61.2$\sstd{$0.8$} & $60.2$\sstd{$0.1$} \\
& & DeepSets &$133.23$\sstd{$1.93$} & $0.58$\sstd{$0.03$} & $153.2$\sstd{$0.8$} & $43.31$\sstd{$2.06$} & $1120.0$\sstd{$5.5$} & $50.5$\sstd{$0.1$} & $49.6$\sstd{$0.2$} & $50.1$\sstd{$0.1$} \\
& & Transformer &$108.48$\sstd{$0.16$} & $0.59$\sstd{$0.08$} & $63.1$\sstd{$2.0$} & $39.70$\sstd{$0.52$} & $1073.3$\sstd{$1.5$} & $63.6$\sstd{$0.1$} & $60.9$\sstd{$0.3$} & $60.3$\sstd{$0.1$} \\
\cmidrule{3-11}

\multirow{3}{*}{Rev-KL} & & GRU &$105.19$\sstd{$0.03$} & \highlight{$0.47$\sstd{$0.01$}} & $71.3$\sstd{$1.3$} & $2.31$\sstd{$0.41$} & $302.9$\sstd{$5.6$} & $63.4$\sstd{$0.1$} & \highlight{$90.4$\sstd{$0.7$}} & $66.2$\sstd{$0.1$} \\
& & DeepSets &$105.09$\sstd{$0.06$} & $0.49$\sstd{$0.01$} & $76.8$\sstd{$1.8$} & \highlight{$0.83$\sstd{$0.02$}} & $454.1$\sstd{$10.2$} & $59.1$\sstd{$0.5$} & $89.1$\sstd{$0.3$} & $62.9$\sstd{$0.1$} \\
& & Transformer &$104.91$\sstd{$0.11$} & \highlight{$0.46$\sstd{$0.00$}} & $33.1$\sstd{$0.3$} & $0.99$\sstd{$0.07$} & \highlight{$274.0$\sstd{$1.3$}} & \highlight{$68.1$\sstd{$0.2$}} & \highlight{$91.1$\sstd{$0.2$}} & $72.6$\sstd{$0.1$} \\
   \bottomrule
    \end{tabular}
    \vspace{-2mm}
    \caption{\textbf{Variable-Dimensional In-Context Posterior Estimation} for estimating the mean of a Gaussian (Gaussian), means of a Gaussian mixture model (GMM), (non-)linear regression (NLR/LR) and (non-)linear binary classification (NLC/LC). For each task, a single model is trained to estimate the posterior for a variable number of features; e.g. the same model estimates both \textit{1D} and \textit{50D} NLR parameters.}
    \vspace{-6mm}
    \label{tab:variable_dim}
\end{table*}
We refer to \Cref{appdx:related_work} for details about related work.
\vspace{-3mm}
\section{Posterior Estimation from Data in Context}
\label{sec:method}
\vspace{-1mm}
As described earlier, standard in-context approaches are predominantly concerned with prediction and model the posterior predictive directly, taking $\gD$ as input. However, they can be tweaked to perform posterior estimation instead. We discuss ways to train such an in-context estimator and showcase its connections to existing amortization methods.

Given any modeling assumption defined via a probabilistic model $p(\cdot | \mtheta)$ with parameters $\mtheta$, we are interested in estimating the full Bayesian posterior over the parameters $p(\mtheta | \gD)$ after obtaining some observations $\gD$, in a manner that allows fast and scalable approximation. \Cref{eq:rkl,eq:fkl} showcase two different methodologies of performing posterior estimation, however, both the methods train a new $q_\varphi$ every time new observations $\gD$ are obtained. However, such methods can be easily amortized by leveraging in-context learning with a goal towards posterior estimation instead of prediction, i.e. training a model to approximate the posterior distribution based on in-context examples. Mathematically, this is obtained by considering an approximate density $q_\varphi(\cdot | \gD)$\footnote{We term this conditional model \emph{in-context posterior estimator}.} which is explicitly conditioned on the set of observations $\gD$ and trained over multiple such sets
\begin{align}
    \varphi^* = \arg\min_\varphi \mathbb{E}_{\gD \sim \chi} \mathbb{D}\left(p(\cdot | \gD), q_\varphi(\cdot|\gD)\right)
\end{align}
where $\chi$ denotes some distribution over observations $\gD$. 

If the measure of divergence is the forward KL $\mathbb{D}_{F\text{-}\mathbb{KL}}$, it leads to the neural posterior estimation methodology of simulation-based inference (SBI-NPE) as long as an additional constraint is satisfied, i.e. $\chi$ defines sampling from the assumed underlying model $p$, \ie
\begin{align}
    \label{eq:chi_sim}
    \chi(\gD) = \int p(\mtheta) \prod_{\vx_n \in \gD} p(\vx_n | \mtheta) d\mtheta
\end{align}
The importance of this constraint is that it leads to a simpler gradient-based optimization procedure as opposed to EP
\begin{align}
    \varphi^*_{F\text{-}\mathbb{KL}} &= \arg\min_\varphi \mathbb{E}_{\gD \sim \chi} \mathbb{E}_{\mtheta \sim p(\cdot | \gD)} \left[\log \frac{p(\mtheta | \gD)}{q_\varphi(\mtheta | \gD)}\right] \\
    \label{eq:afkl}
    &= \arg\min_\varphi \mathbb{E}_{\mtheta} \mathbb{E}_{\gD \sim p(\cdot | \mtheta)} \left[-\log q_\varphi(\mtheta | \gD)\right]
\end{align}
where we focus our attention to the change in expectations which is only possible when $\gD$ is sampled according to $p$. This removes the requirement of sampling or evaluating the true posterior, a known hurdle with EP methods. 

\begin{table*}
\centering
\small
    \setlength{\tabcolsep}{15pt}
    \begin{tabular}{@{}lcr cccc}
        \toprule
         &  &  & \multicolumn{2}{c}{\textit{$L_2$ Loss} ($\downarrow)$} & \multicolumn{2}{c}{\textit{Accuracy} ($\uparrow$)}\\
        \cmidrule(lr){4-5}\cmidrule(lr){6-7}
        \textbf{Objective} & $q_\varphi$ & \textbf{Model} & LR & NLR & LC & NLC \\
        \midrule
Baseline & - & Random & $23.52$\sstd{$0.42$} & $209.35$\sstd{$9.92$} & $50.10$\sstd{$0.17$} & $50.84$\sstd{$1.00$} \\
\cmidrule{2-7}

\multirow{3}{*}{Fwd-KL} & \multirow{6}{*}{\rotatebox[origin=c]{90}{Gaussian}} & GRU &$8.95$\sstd{$0.47$} & $84.63$\sstd{$3.96$} & $76.34$\sstd{$1.58$} & $60.00$\sstd{$0.98$} \\
& & DeepSets &$10.81$\sstd{$0.08$} & $97.51$\sstd{$2.96$} & $68.26$\sstd{$0.31$} & $50.84$\sstd{$0.99$} \\
& & Transformer &$9.35$\sstd{$0.99$} & $111.07$\sstd{$6.55$} & $63.97$\sstd{$3.21$} & $60.39$\sstd{$0.53$} \\
\cmidrule{3-7}

\multirow{3}{*}{Rev-KL} & & GRU &$9.78$\sstd{$2.75$} & $17.01$\sstd{$5.16$} & \highlight{$79.71$\sstd{$1.12$}} & $77.19$\sstd{$0.21$} \\
& & DeepSets &$9.26$\sstd{$0.10$} & \highlight{$8.03$\sstd{$0.23$}} & $77.40$\sstd{$0.14$} & $72.63$\sstd{$0.14$} \\
& & Transformer & \highlight{$7.30$\sstd{$0.21$}} & \highlight{$8.23$\sstd{$1.26$}} & $76.96$\sstd{$1.58$} & $71.33$\sstd{$5.44$} \\
\cmidrule{2-7}

\multirow{3}{*}{Fwd-KL} & \multirow{6}{*}{\rotatebox[origin=c]{90}{Flow}} & GRU &$8.28$\sstd{$0.33$} & $52.04$\sstd{$3.39$} & $76.45$\sstd{$1.43$} & $61.10$\sstd{$0.61$} \\
& & DeepSets &$12.94$\sstd{$0.41$} & $71.69$\sstd{$2.38$} & $67.99$\sstd{$2.15$} & $49.74$\sstd{$0.76$} \\
& & Transformer &$9.64$\sstd{$0.50$} & $84.45$\sstd{$7.88$} & $68.26$\sstd{$3.23$} & $61.65$\sstd{$1.22$} \\
\cmidrule{3-7}

\multirow{3}{*}{Rev-KL} & & GRU &$8.17$\sstd{$0.25$} & $17.35$\sstd{$8.02$} & $66.30$\sstd{$2.81$} & \highlight{$78.55$\sstd{$1.27$}} \\
& & DeepSets &$11.05$\sstd{$0.35$} & $8.54$\sstd{$0.49$} & $78.18$\sstd{$0.13$} & $71.69$\sstd{$0.18$} \\
& & Transformer &\highlight{$7.48$\sstd{$0.26$}} & $8.80$\sstd{$1.80$} & $72.75$\sstd{$5.73$} & \highlight{$78.23$\sstd{$0.89$}} \\
\bottomrule        
    \end{tabular}
    \vspace{-2mm}
    \caption{\textbf{Tabular Experiments}: Zero-shot performance of the amortized variable-dimensional models across (non-)linear regression and classification real-world tabular tasks. All the models are trained solely on simulated data, and evaluated zero-shot on real-world data with varying number of both features and training (observations that are fed as in-context amortization) observations.}
    \vspace{-6mm}
    \label{tab:tabular}
\end{table*}
In contrast, one could also take motivation from VAEs and NP which predominantly work under the reverse KL divergence $\mathbb{D}_{R\text{-}\mathbb{KL}}$ minimization. An amortized in-context learner in this setting can be mathematically formalized as
\begin{align}
\varphi^*_{R\text{-}\mathbb{KL}} &= \arg\min_\varphi \mathbb{E}_{\gD \sim \chi} \mathbb{E}_{\mtheta \sim q_\varphi(\cdot | \gD)} \left[\log \frac{q_\varphi(\mtheta | \gD)}{p(\mtheta | \gD)}\right] \\
\label{eq:arkl}
&= \arg\min_\varphi \mathbb{E}_{\gD \sim \chi} \mathbb{E}_{\mtheta \sim q_\varphi(\cdot | \gD)} \left[\log \frac{q_\varphi(\mtheta | \gD)}{p(\gD, \mtheta)}\right]
\end{align}
It is important to note that unlike forward KL, reverse KL provides the freedom of choosing any arbitrary $\chi$ while still maintaining ease in training, i.e. the in-context estimator can be trained on datasets that come from a different distribution than $p$. Such flexibility is important because we often want to generalize well to data whose underlying true probabilistic model is unknown, and thus we often prescribe a likelihood based on our best belief. For \eg, practitioners often model regression problems as linear even when the data might come from a nonlinear process like a Gaussian Process. Reverse KL allows us to train the in-context learner on a steady stream of data even when the likelihood is misspecified, while the forward KL objective in this case will have to be trained on simulated linear data.

The estimators defined in \Cref{eq:afkl,eq:arkl} rely on a parameterization of $q_\varphi(\cdot | \gD)$, which involves two components: a flexible parametric form of density and an architecture that can be conditioned on entire datasets. In this work, we provide an in-depth analysis into the use of a diagonal Gaussian and discrete-time normalizing flows for the former, and Gated Recurrent Units (GRU), DeepSets and Transformers for the latter. We note that the conditioning architecture should respect permutation invariance of the approximate posterior to the ordering in the observations, 
which is respected in DeepSets and Transformers but not in GRUs. See \Cref{appdx:exchangeability} for architectural choice details.

Finally, these training procedures naturally introduce a dependency on the dataset generating distribution $\chi$. Since we are working with a known probabilistic model, an obvious choice of $\chi$ is to treat this probabilistic model as a black-box simulator and generate samples using ancestral sampling.

In the following sections, we provide a comparative analysis between the two approaches to in-context posterior estimation, as well as the different architectural choices and parametrizations of the density $q_\varphi$. While this has been independently studied in the SBI and NP framework, a rigorous comparative study between them through the lens of predictive and sample-based metrics has been lacking. Further, we aim to understand the impact of misspecification in such in-context learners, i.e. when real data may not come from the $p$ defined by the likelihood and the prior. To study this rigorously, we further evaluate the in-context estimators on observations coming from different underlying known and unknown processes in \Cref{subsec:misspecification,subsec:tabular}.
\vspace{-3mm}
\section{Experiments}
\label{sec:experiments}
\vspace{-1mm}
\looseness=-1
To provide a fair and comprehensive evaluation of the different estimators and modeling choices, we consider a variety of well-known probabilistic models encompassing supervised and unsupervised scenarios. In particular, we look at the problem of estimating the Bayesian posterior over the (a) mean of a Gaussian distribution (GM), (b) means of a Gaussian mixture model (GMM), (c) parameters of a (non-)linear regression model (NLR/LR), and (d) parameters of a (non-)linear classification model (NLC/LC). We refer the readers to \Cref{appdx:probabilistic_models} for particulars about the probabilistic models, including their likelihoods and priors considered.

\textbf{Baselines}. We consider dataset-specific baselines to compare different amortized in-context posterior estimators with. In particular, we use the prior (Random), perform maximum likelihood estimation using gradient-based optimization (Optimization) as well as an approximate Bayesian inference procedure through Langevin and Hamiltonian based MCMC sampling. Such baselines rely on iterative procedures and must be run independently for different datasets.

\textbf{Metrics}. We consider two different types of metrics: predictive and sample-based. For the former, we consider $L_2$ loss and accuracy as applicable, in the following manner
\begin{align}
    \mathbb{E}_{(\vx_*,y_*), \gD \sim \chi} \mathbb{E}_{\mtheta \sim q_\varphi(\cdot | \gD)}\textsc{Metric}\left(\hat{y}, y_*\right)
\end{align}
where $\hat{y}$ is the mode of the distribution $p(\cdot | \vx_*, \mtheta)$ and $\textsc{Metric}$ is $L_2$ loss or accuracy for regression and classification respectively. For unsupervised learning settings, we consider a similar $L_2$ based metric defining distance from the mean of the Gaussian or the closest mean in the GMM. For sample-based metrics, we leverage the 
\begin{align}
    \gW_2^2 &= \inf_\pi \iint \norm{\mtheta_q - \mtheta_p}^2 d\pi(q, p)
\end{align}
where $\pi$ denotes a joint distribution over $(\mtheta_q, \mtheta_p)$ with marginals $q_\varphi(\cdot | \gD)$ and $p(\cdot | \gD)$ respectively. This is called the 2-Wasserstein metric which can be computed with finite samples from each, where we use samples from MCMC as reference for $p$. We also leverage the symmetric KL divergence as a metric whenever the true posterior is available. 

We refer to \Cref{appdx:experiment,appdx:metrics,appdx:results} for details about the experiments, metrics and additional results respectively.

\vspace{-2mm}
\subsection{Zero-Shot Posterior Approximation}
\vspace{-1mm}
\label{subsec:fixed-dim}
\looseness=-1
We first test the in-context estimators' ability to succeed at novel tasks solely at inference over $q_\varphi(\cdot | \gD)$. To do so, we train the estimators on datasets being generated as $\gD_{\text{train}} \sim p$, and are then evaluated on new $\gD_{\text{test}} \sim p$. Mathematically this is equivalent to setting $\chi$ according to \cref{eq:chi_sim} where the number of observations, $|\gD|$, is varied in some range both during training and evaluation. 

\begin{figure*}
    \centering
    \captionsetup[subfigure]{font=scriptsize}
    \includegraphics[width=\textwidth]{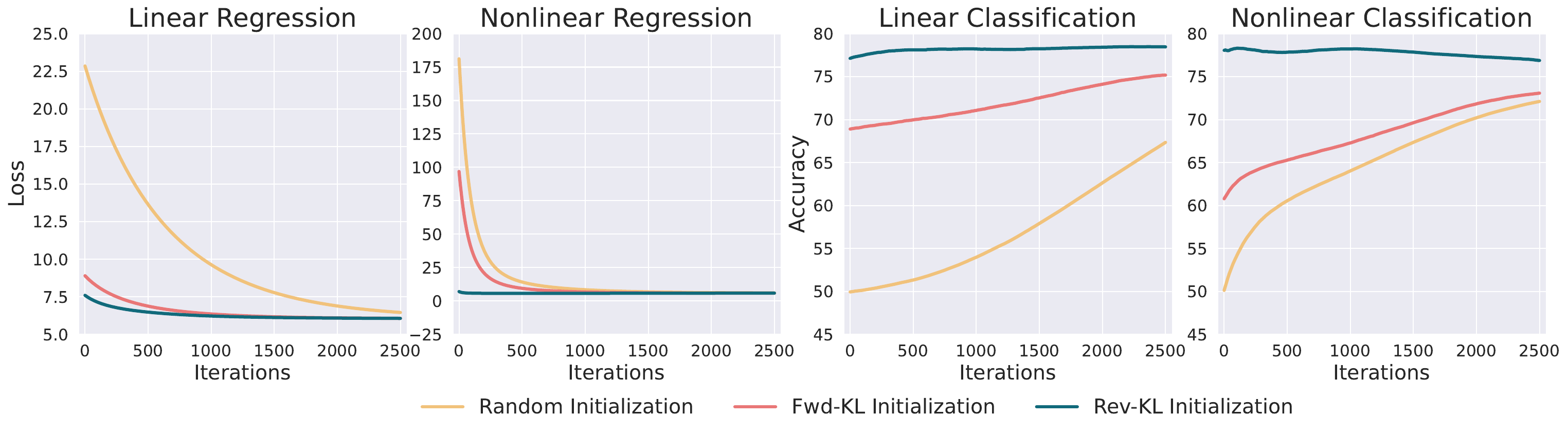}
    \vspace{-10mm}
    \caption{\textbf{Tabular Experiments:} Initializing parameters from the proposed amortized model leads to good zero-shot performance and often optimal initialization across (non-)linear regression and classification tasks.}
    \vspace{-5mm}
    \label{fig:tabular}
\end{figure*}
\Cref{fig:fixed_dim} visualizes the amortized estimators in low-dimensional problems, showing that they learn meaningful distributions over the parameters zero-shot on new tasks. Next, we turn our attention to quantitative assessment of the different estimators on more complex, high-dimensional counterparts of the same probabilistic models. \Cref{tab:fixed_dim} shows that the in-context estimators are often comparable to optimization and MCMC baselines, with reverse KL objective combined with normalizing flows and the transformer architecture outperforming other design choices. In particular, we see that for high-dimensional multi-modal problems like NLR/NLC, forward KL approach does not fare well potentially due to its mode averaging property. Surprisingly, we also see non permutation invariant architectures like GRUs perform well, and often better than DeepSets.

\vspace{-3mm}
\subsection{Generalizing to Variable Feature Dimensions}
\vspace{-1mm}
\label{subsec:variable-dim}
So far, we only considered amortization over datasets for the same underlying likelihood model, which fixes the dimensionality of the problem. For example, a different in-context estimator has to be trained for a $2$-dimensional and $5$-dimensional Bayesian linear regression model since the dimensionality of $\mtheta$ changes. It is important to note that a deep learning-based approach leaves hopes of generalizing to new datasets of different dimensionalities since the underlying functional form of the solution remains constant across different datasets, irrespective of the number of features, and is given by the solution obtained from Equation~\ref{eq:bayes_rule}. 

Alternatively, we can see that a low-dimensional problem can just be embedded into a high-dimensional space, with the extra features and parameters set to $0$, akin to the procedure of masking unnecessary dimensions, similar to \cite{hollmann2022tabpfn}. This simple but strong insight allows us to amortize $q_\varphi$ over datasets with varying dimensionalities.

We embed all low-dimensional problems in a $100$-dimensional space by masking the unnecessary dimensions. Our experiments in \Cref{tab:variable_dim} indicate that the same in-context learner can generalize to novel datasets \textit{with a variable number of feature dimensions} zero-shot.

\vspace{-3mm}
\subsection{Model Misspecification}
\vspace{-1mm}
\label{subsec:misspecification}
\looseness=-1
The true likelihood model underlying a data-generating process is often unknown. Practitioners address this by assuming a likelihood model and fitting its parameters to best explain the data. For example, while the true model for classifying emails as spam or not is unknown, one can assume a linear model to approximate the problem. This introduces model misspecification—a mismatch between the assumed and true model.

\looseness=-1
As discussed in \Cref{sec:method}, forward KL methods train only on simulated data from the assumed model, whereas reverse KL methods can use real-world data. Consequently, forward KL approaches struggle with sim-to-real transfer because they cannot incorporate real data during training. In contrast, reverse KL methods leverage real data, leading to more robust predictions in practical settings.

\begin{table}
    \centering
    \small
    \setlength{\tabcolsep}{0.5pt}
    \begin{tabular}{@{}lccccc}
    \toprule
     \multirow{2}{*}{$\chi_{real}\;(\rightarrow)$} & \multirow{3}{*}{$q_\varphi$} & \multirow{2}{*}{\textit{Data}} & \multirow{2}{*}{Linear} & MLP & GP \\
     & & & & Nonlinear & Nonlinear \\
    \cmidrule{3-6}
    $\chi_{sim}\;(\rightarrow)$& &  \textit{Model} & NLR & LR & NLR \\
    \midrule

Fwd-KL & \multirow{3}{*}{\rotatebox[origin=c]{90}{Gaussian}} && $15.454$\sstd{$0.246$} & $2.216$\sstd{$0.097$} & $14.733$\sstd{$0.513$} \\
Rev-KL &&& $0.382$\sstd{$0.003$} & $1.892$\sstd{$0.113$} & $0.155$\sstd{$0.006$} \\
\;\textit{+ switched data} &&& $0.367$\sstd{$0.006$} & \highlight{$1.226$\sstd{$0.001$}} & $0.066$\sstd{$0.004$} \\
\midrule
Fwd-KL & \multirow{3}{*}{\rotatebox[origin=c]{90}{Flow}} && $7.949$\sstd{$0.419$} & $1.632$\sstd{$0.070$} & $8.557$\sstd{$0.561$} \\
Rev-KL &&& \highlight{$0.347$\sstd{$0.001$}} & $1.471$\sstd{$0.016$} & $0.120$\sstd{$0.005$} \\
\;\textit{+ switched data} &&& \highlight{$0.346$\sstd{$0.002$}} & \highlight{$1.226$\sstd{$0.004$}} & \highlight{$0.055$\sstd{$0.002$}} \\
\bottomrule
    \end{tabular}
    \vspace{-3.5mm}
    \caption{\textbf{Misspecification}. OoD evaluation under predictive $L_2$ metric when the true data generating process, $\chi_{real}$, is not known, while $\chi_{sim}$ denotes the simulated one according to the (wrongly) assumed model. Estimators are trained on $\chi_{sim}$, with switched data denoting training on $\chi_{real}$, and evaluation is solely on $\chi_{real}$.}
    \vspace{-7mm}
    \label{tab:misspecification}
\end{table}
\looseness=-1
Mathematically, let $\chi_{sim}$ from \Cref{eq:chi_sim} denote simulated data and $\chi_{real}$ the actual target data. \Cref{tab:misspecification} shows that reverse KL methods outperform forward KL when trained on $\chi_{sim}$ but tested on $\chi_{real}$. Moreover, reverse KL methods trained directly on $\chi_{real}$ generalize even better (see rows labeled “+ switched data”).
For experimental details and additional results on model misspecification, see Appendices~\ref{appdx:details_misspecification} and~\ref{appdx:results_missspecification}.

\vspace{-3mm}
\subsection{Application to Tabular Benchmarks}
\vspace{-1mm}
\label{subsec:tabular}
To evaluate the efficacy of the trained in-context estimators and their ability to generalize out-of-distribution, we test the models trained in \Cref{subsec:variable-dim} on a suite of regression and classification problems chosen from the OpenML platform. These tasks have varying number of feature dimensions and inherently have different data statistics than the ones obtained from $\chi$ during training. \Cref{tab:tabular} shows the zero-shot performance of the parameters inferred from the in-context estimators, and highlights that they perform considerably better than chance, with transformer models and reverse KL methods outperforming other modeling choices.

We also look at finetuning the inferred parameters from the in-context estimators with a maximum-a-posteriori (\textit{MAP}) objective and compare its performance with a corresponding model initialized from the prior. Our results in \Cref{fig:tabular} highlights that in-context estimators lead to much faster convergence, with reverse KL methods being superior in complex, multi-modal and nonlinear tasks.

\begin{table}
    \centering
    \small	    
    \setlength{\tabcolsep}{0.5pt}
    \begin{tabular}{@{}lr cc cc}
        \toprule
         &  & \multicolumn{4}{c}{\textit{Symmetric KL Divergence} ($\downarrow)$} \\
         \cmidrule(lr){3-6}
        & \textbf{Model} & \multicolumn{2}{c}{\textbf{Gaussian Mean}} & \multicolumn{2}{c}{\textbf{LR}} \\
        \cmidrule(lr){3-4}\cmidrule(lr){5-6}
        & & \textit{2D} & \textit{100D} & \textit{1D} & \textit{100D} \\
        \midrule
    Baseline & Random & $44.32$ & $46.78$ & $179.2$ & $186.8$ \\
        \cmidrule{2-6}
\multirow{3}{*}{Fwd-KL} & GRU & \highlight{$0.018$\sstd{$0.007$}} & $0.07$\sstd{$0.00$} & $0.04$\sstd{$0.01$} & $81.8$\sstd{$0.2$} \\
& DeepSets & $0.037$\sstd{$0.015$} & $0.22$\sstd{$0.01$} & $0.06$\sstd{$0.00$} & $82.0$\sstd{$0.4$} \\
& Transformer & $0.030$\sstd{$0.008$} & $0.06$\sstd{$0.00$} & $0.04$\sstd{$0.01$} & \highlight{$20.6$\sstd{$1.0$}} \\
\cmidrule{2-6}
\multirow{3}{*}{Rev-KL} & GRU & \highlight{$0.017$\sstd{$0.005$}} & $0.08$\sstd{$0.00$} & \highlight{$0.03$\sstd{$0.00$}} & $78.7$\sstd{$1.5$} \\
& DeepSets & $0.029$\sstd{$0.002$} & $0.19$\sstd{$0.01$} & $0.05$\sstd{$0.00$} & $104.5$\sstd{$8.8$} \\
& Transformer & $0.035$\sstd{$0.013$} & \highlight{$0.05$\sstd{$0.00$}} & \highlight{$0.03$\sstd{$0.00$}} & $32.7$\sstd{$1.0$} \\
        \bottomrule
    \end{tabular}
    \vspace{-3.5mm}
    \caption{\textbf{Normalized Symmetric KL Divergence}. Amortized models with Gaussian $q_\varphi$ approximate the true posterior well in tasks with tractable posteriors, when compared to the prior.}
    \vspace{-8mm}
    \label{tab:posterior_metrics}

\end{table}
Finally, we look at a suite of problems that are extremely out-of-distribution from the $\chi$ used during training. In particular, we look at a suite of regression and classification tasks from the OpenML platform which consist of tasks with varying number of features. We refer the readers to Appendix~\ref{appdx:results_tabular} for results on individual datasets with different $q_\varphi$, as well as Appendix~\ref{appdx:details_tabular} for implementation details.

\begin{figure*}
    \centering
    \captionsetup[subfigure]{font=scriptsize}
    \begin{subfigure}[b]{0.45\textwidth}
          \centering
          \begin{subfigure}[b]{0.5\textwidth}
                  \centering
                  \includegraphics[width=\textwidth]{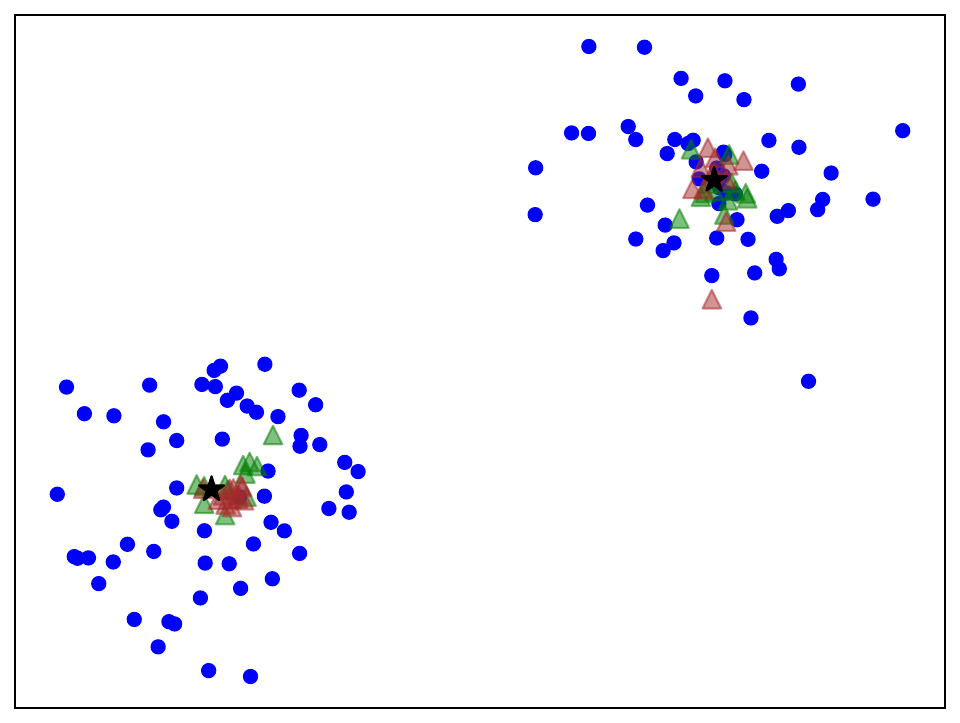}
                  \caption{Forward-KL}
          \end{subfigure}%
          \begin{subfigure}[b]{0.5\textwidth}
                  \centering
                  \includegraphics[width=\textwidth]{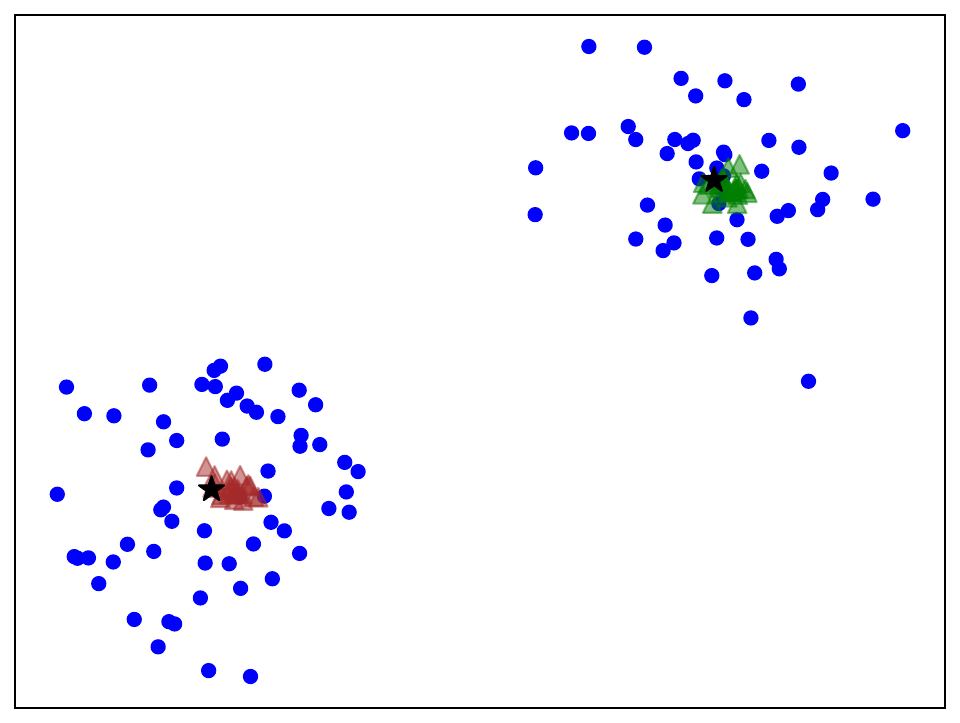}
                  \caption{Reverse-KL}
          \end{subfigure}
    \end{subfigure}
    \hspace{1mm}
    \vline
    \hspace{1mm}
    \begin{subfigure}[b]{0.45\textwidth}
          \centering
          \begin{subfigure}[b]{0.43\textwidth}
                  \centering
                  \includegraphics[width=\textwidth]{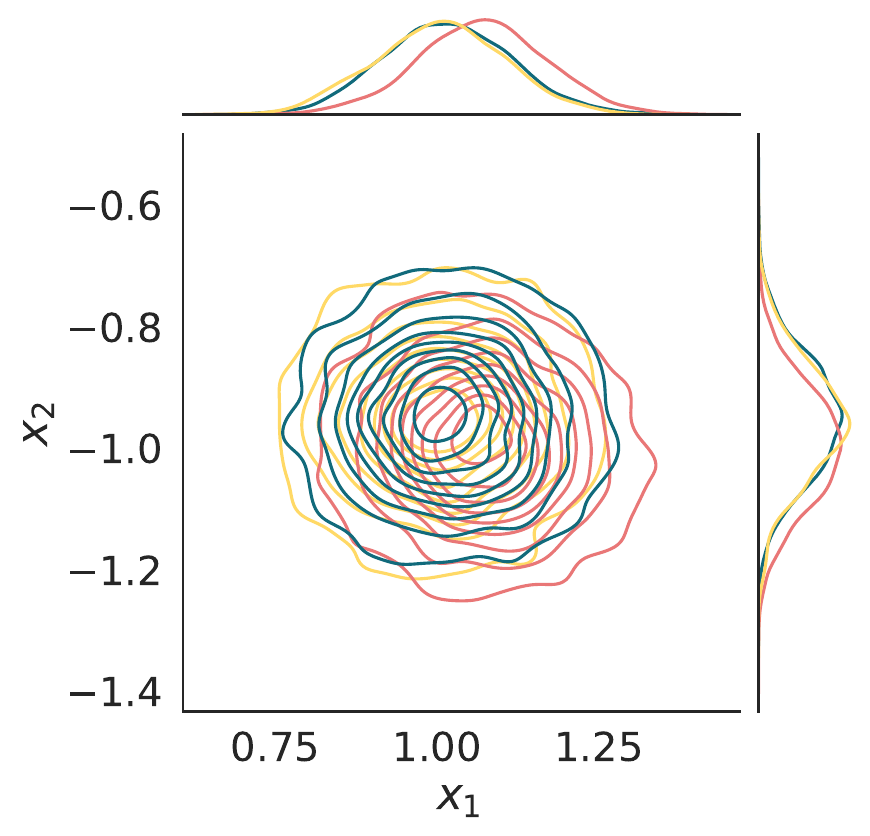}
                  \caption{Mean of Gaussian}
          \end{subfigure}
          \begin{subfigure}[b]{0.43\textwidth}
                  \centering
                  \includegraphics[width=\textwidth]{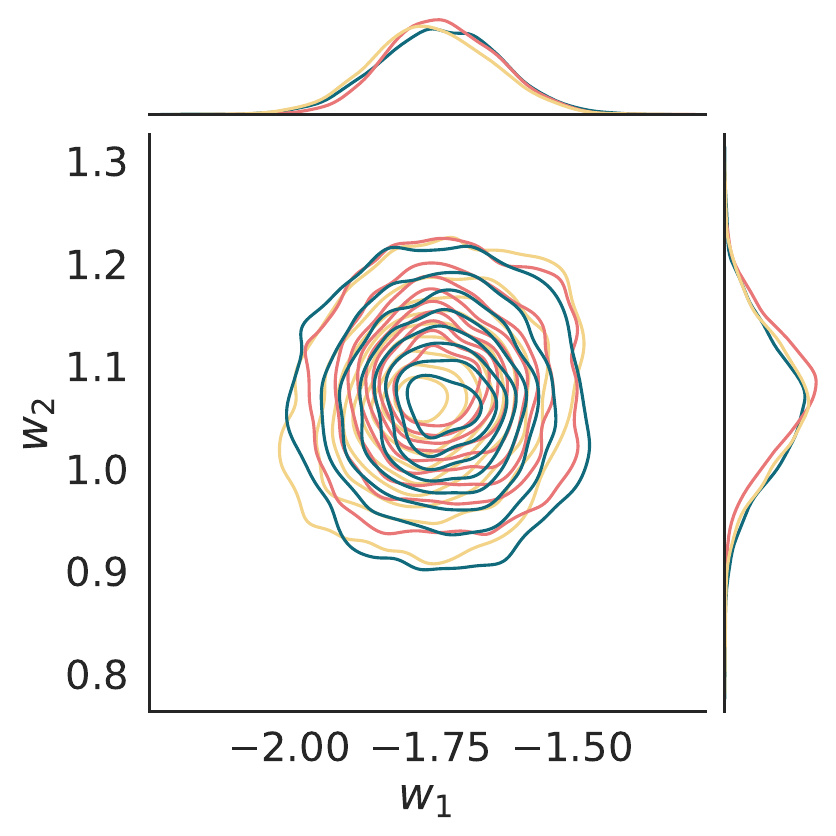}
                  \vspace*{-4.5mm}
                  \caption{Linear Regression}
          \end{subfigure}
          \begin{subfigure}[b]{0.11\textwidth}
                  \hspace*{-3mm}
                  \includegraphics[trim=14cm 10cm 0mm 0mm, clip, width=2.25\textwidth]
                  {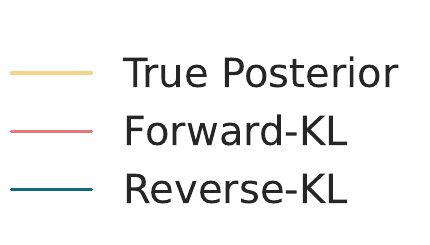}
                  \vspace*{18mm}
          \end{subfigure}
    \end{subfigure}
    \vspace{-3mm}
    \caption{\textbf{Left}: Estimation of the means of a GMM, where red and green samples denote the first and second mean vectors. Unlike in reverse KL, the cluster labels switch in forward KL, highlighting its ability to capture underlying multi-modality. \textbf{Right}: Kernel density estimation of the true posterior, overlaid with estimates from forward and reverse KL systems, for different probabilistic models.}
    \vspace{-2mm}
    \label{fig:mode_switching}
\end{figure*}
\begin{table*}[t]
    \centering
    \small
    \setlength{\tabcolsep}{1pt}
    \begin{tabular}{lcr c c c cc c cc}
        \toprule
         &  &  & \multicolumn{8}{c}{\textit{$\mathcal{W}_2^2$} ($\downarrow$)} \\
         \cmidrule(lr){4-8}\cmidrule(lr){9-11}
        \textbf{Objective} & $q_\varphi$ & \textbf{Model} & \multicolumn{1}{c}{\textbf{Gaussian}} & \multicolumn{1}{c}{\textbf{GMM}} & \multicolumn{1}{c}{\textbf{LR}} & \multicolumn{2}{c}{\textbf{NLR}} & \multicolumn{1}{c}{\textbf{LC}} & \multicolumn{2}{c}{\textbf{NLC}} \\
        \cmidrule(lr){4-4}\cmidrule(lr){5-5}\cmidrule(lr){6-6}\cmidrule(lr){7-8}\cmidrule(lr){9-9}\cmidrule(lr){10-11}
        & & & \textit{100D} & \textit{5D 2 cl} & \textit{100D} & 
        \textit{1D} & \textit{25D} & \textit{100D} &  \textit{2D} & \textit{25D}\\
        \midrule

Baseline & - & Random & $13.96$\sstd{$0.00$} & $4.01$\sstd{$0.00$} & $13.53$\sstd{$0.00$} & $11.21$\sstd{$0.00$} & $36.46$\sstd{$0.00$} & $16.72$\sstd{$0.00$} & $14.71$\sstd{$0.00$} & $36.67$\sstd{$0.00$} \\
\cmidrule{2-11}

\multirow{3}{*}{Fwd-KL} & \multirow{6}{*}{\rotatebox[origin=c]{90}{Gaussian}} & GRU &$1.37$\sstd{$0.00$} & $2.35$\sstd{$0.01$} & $10.42$\sstd{$0.02$} & $11.10$\sstd{$0.00$} & $36.33$\sstd{$0.01$} & $15.25$\sstd{$0.01$} & $14.67$\sstd{$0.00$} & $36.66$\sstd{$0.00$} \\
& & DeepSets &$1.55$\sstd{$0.01$} & $2.35$\sstd{$0.01$} & $10.39$\sstd{$0.03$} & $11.11$\sstd{$0.01$} & $36.41$\sstd{$0.01$} & $16.72$\sstd{$0.00$} & $14.68$\sstd{$0.00$} & $36.66$\sstd{$0.00$} \\
& & Transformer &$1.41$\sstd{$0.00$} & $2.40$\sstd{$0.01$} & $5.83$\sstd{$0.09$} & \highlight{$11.09$\sstd{$0.01$}} & $36.32$\sstd{$0.01$} & $14.70$\sstd{$0.01$} & $14.67$\sstd{$0.00$} & $36.66$\sstd{$0.00$} \\
\cmidrule{3-11}

\multirow{3}{*}{Rev-KL} & & GRU &$1.34$\sstd{$0.00$} & $2.98$\sstd{$0.01$} & $7.39$\sstd{$0.03$} & $11.31$\sstd{$0.01$} & $35.92$\sstd{$0.32$} & \highlight{$12.31$\sstd{$0.01$}} & $14.14$\sstd{$0.01$} & $35.16$\sstd{$0.00$} \\
& & DeepSets &$1.38$\sstd{$0.02$} & $2.98$\sstd{$0.02$} & $7.58$\sstd{$0.05$} & $11.31$\sstd{$0.02$} & \highlight{$35.58$\sstd{$0.20$}} & $12.93$\sstd{$0.06$} & \highlight{$14.10$\sstd{$0.01$}} & \highlight{$35.05$\sstd{$0.00$}} \\
& & Transformer & \highlight{$1.34$\sstd{$0.01$}} & $2.98$\sstd{$0.03$} & \highlight{$4.84$\sstd{$0.03$}} & $11.38$\sstd{$0.01$} & $35.86$\sstd{$0.08$} & $12.84$\sstd{$0.04$} & $14.17$\sstd{$0.02$} & $35.36$\sstd{$0.00$} \\
\cmidrule{2-11}

\multirow{3}{*}{Fwd-KL} & \multirow{6}{*}{\rotatebox[origin=c]{90}{Flow}} & GRU &$1.37$\sstd{$0.00$} & $1.71$\sstd{$0.16$} & $10.18$\sstd{$0.04$} & \highlight{$11.09$\sstd{$0.02$}} & $36.35$\sstd{$0.01$} & $15.29$\sstd{$0.02$} & $14.66$\sstd{$0.01$} & $36.66$\sstd{$0.00$} \\
& & DeepSets &$1.58$\sstd{$0.01$} & $1.81$\sstd{$0.08$} & $10.48$\sstd{$0.14$} & \highlight{$11.08$\sstd{$0.01$}} & $36.41$\sstd{$0.00$} & $16.72$\sstd{$0.00$} & $14.67$\sstd{$0.01$} & $36.66$\sstd{$0.00$} \\
& & Transformer &$1.40$\sstd{$0.01$} & \highlight{$1.20$\sstd{$0.26$}} & $5.64$\sstd{$0.23$} & \highlight{$11.08$\sstd{$0.00$}} & $36.32$\sstd{$0.02$} & $14.66$\sstd{$0.01$} & $14.65$\sstd{$0.01$} & $36.66$\sstd{$0.00$} \\
\cmidrule{3-11}

\multirow{3}{*}{Rev-KL} & & GRU & \highlight{$1.33$\sstd{$0.00$}} & $2.99$\sstd{$0.02$} & $7.48$\sstd{$0.06$} & $11.15$\sstd{$0.04$} & $35.97$\sstd{$0.04$} & $13.51$\sstd{$0.02$} & $14.33$\sstd{$0.01$} & $35.79$\sstd{$0.01$} \\
& & DeepSets &$1.41$\sstd{$0.02$} & $2.96$\sstd{$0.01$} & $8.40$\sstd{$0.09$} & $11.15$\sstd{$0.04$} & $36.02$\sstd{$0.09$} & $13.61$\sstd{$0.05$} & $14.32$\sstd{$0.01$} & $35.69$\sstd{$0.01$} \\
& & Transformer & \highlight{$1.33$\sstd{$0.01$}} & $3.00$\sstd{$0.04$} & \highlight{$4.90$\sstd{$0.15$}} & $11.21$\sstd{$0.02$} & $36.00$\sstd{$0.17$} & $13.63$\sstd{$0.02$} & $14.37$\sstd{$0.01$} & $35.88$\sstd{$0.02$} \\
\bottomrule
    \end{tabular}
    \vspace{-1mm}
    \caption{\textbf{Sample Based Metrics}. We compute the $2$-Wasserstein metric between samples from the approximate posterior and MCMC.}
    \vspace{-6mm}
    \label{tab:w2}
\end{table*}
\vspace{-3mm}
\subsection{Evaluating Posterior Quality}
\vspace{-1mm}
While comparing with the true posterior is hard due to its intractability, it is still available when estimating the mean of a Gaussian distribution or performing Bayesian Linear Regression. Figure~\ref{fig:mode_switching} (\textbf{Right}) shows the kernel density estimate of the samples from the true posterior, amortized forward, and reverse KL model, showing that both estimators efficiently capture the true posterior. We further quantify it through the symmetric KL divergence in Table~\ref{tab:posterior_metrics}. 

For more complex problems, we compute the squared Wasserstein metric $\mathcal{W}_2^2$ between samples from the amortized posterior and multiple chains of Langevin MCMC in \cref{tab:w2}. Our results indicate that reverse KL approaches do slightly better in high-dimensional setups, while for low-dimensional multi-modal scenarios (eg. GMM), forward KL approaches fare better. Importantly, we note that this metric only provides a crude proxy to the quality of the posterior, since MCMC methods only provide asymptotic guarantees.


\vspace{-3mm}
\section{Discussion and Conclusion}
\vspace{-2mm}
We show that Bayesian posterior inference can be amortized for a broad class of probabilistic models and explore a variety of design decisions associated with it. Some key conclusions from our analysis are described below.

\textbf{Forward vs Reverse KL}. Our GMM experiments (\cref{fig:mode_switching}; \textbf{Left}) indicate that forward KL is more amenable to learning multimodal solutions compared to reverse KL in low-dimensional problems. However, the latter outperforms the former in both predictive and sample-based metrics when the parameter space is high-dimensional. Further, reverse KL methods do not require access to $(\mtheta, \gD)$ samples during training and thus show improvements in misspecification and simulation to real transfer.

\looseness=-1
\textbf{Architectural Choices}. We compare permutation invariant architectures like DeepSets and Transformers with non invariant architecture like GRU and see that GRU outperforms DeepSets even though the latter is permutation invariant. We hypothesize that this could be due to limited expressivity of DeepSets and their reliance on a fixed pooling operator. In contrast, GRUs can learn to be approximately permutation invariant through training. We further see that Transformers outperform both DeepSets and GRUs as they do not rely on fixed aggregation schemes but still respect the invariant structure of the posterior.

\looseness=-1
\textbf{Capacity of $q_\varphi$}. Increasing the capacity of $q_\varphi$ using normalizing flows substantially helps for forward KL but only marginally for the reverse KL objective. We hypothesize that because of the mode-seeking tendency of reverse KL, even with the capacity to model different modes, the algorithm latches to a single one. However, in forward KL setup without additional capacity the model overestimates the variance a lot.

We provide a rigorous comparison of different in-context posterior estimators, especially in the presence of misspecification and generalization to real-world problems. It provides an exciting direction of research which could reduce the load of real-world, complex, and iterative approximations through quick and cheap inference over a trained amortized network -- providing a direction into learning a generalist in-context Bayesian estimator. We believe that scaling our approach to more complex probabilistic models, leveraging better modeling choices for high-dimensional problems~\citep{bengio2021flow,zhang2021path,vargas2023denoising}, and training a single model for multiple probabilistic models are important directions of future work.
\clearpage
\section*{Acknowledgements}
The authors would like to acknowledge the computing resources provided by the Mila cluster to enable the experiments outlined in this work. SM acknowledges the support of UNIQUE's scholarship.
GL acknowledges the support of the Canada CIFAR AI Chair program, NSERC Discovery Grant RGPIN-2018-04821, and a Canada Research Chair in Neural Computations and Interfacing. 
MAB acknowledges the support of the Canada First Research Excellence Fund (CFREF) for the Vision: Science to Applications (VISTA) program, NSERC Discovery Grant RGPIN-2017-05638 and Google.
The authors also thank NVIDIA for computing resources.

\section*{Impact Statement}
We provide a comprehensive evaluation of different approaches and design choices in performing Bayesian posterior estimation for a wide variety of probabilistic models. We believe that analysis into such amortized estimators could lead to more efficient and scalable Bayesian methods that can lead to robust predictions and better OoD generalization. Thus, we believe that our work generally advances the field of machine learning through careful and thorough benchmarking. There are many potential societal
consequences of our work, none of which we feel must be
specifically highlighted here.
\bibliography{references}

\begin{thebibliography}{70}
\providecommand{\natexlab}[1]{#1}
\providecommand{\url}[1]{\texttt{#1}}
\expandafter\ifx\csname urlstyle\endcsname\relax
  \providecommand{\doi}[1]{doi: #1}\else
  \providecommand{\doi}{doi: \begingroup \urlstyle{rm}\Url}\fi

\bibitem[Ardizzone et~al.(2018)Ardizzone, Bungert, Draxler, K{\"o}the, Kruse, Schmier, and Sorrenson]{Ardizzone2018freia}
Ardizzone, L., Bungert, T., Draxler, F., K{\"o}the, U., Kruse, J., Schmier, R., and Sorrenson, P.
\newblock {{FrEIA}}: Framework for easily invertible architectures, 2018.
\newblock URL \url{https://github.com/vislearn/FrEIA}.

\bibitem[Ardizzone et~al.(2018-2022)Ardizzone, Bungert, Draxler, Köthe, Kruse, Schmier, and Sorrenson]{freia}
Ardizzone, L., Bungert, T., Draxler, F., Köthe, U., Kruse, J., Schmier, R., and Sorrenson, P.
\newblock {Framework for Easily Invertible Architectures (FrEIA)}, 2018-2022.
\newblock URL \url{https://github.com/vislearn/FrEIA}.

\bibitem[Arenz et~al.(2022)Arenz, Dahlinger, Ye, Volpp, and Neumann]{arenz2022unified}
Arenz, O., Dahlinger, P., Ye, Z., Volpp, M., and Neumann, G.
\newblock A unified perspective on natural gradient variational inference with gaussian mixture models.
\newblock \emph{arXiv preprint arXiv:2209.11533}, 2022.

\bibitem[Athanasopoulos et~al.(2023)Athanasopoulos, Hyndman, Kourentzes, and Panagiotelis]{Athanasopoulos-hierarchicalts}
Athanasopoulos, G., Hyndman, R.~J., Kourentzes, N., and Panagiotelis, A.
\newblock Forecast reconciliation: A review.
\newblock \emph{International Journal of Forecasting}, 2023.
\newblock ISSN 0169-2070.
\newblock \doi{https://doi.org/10.1016/j.ijforecast.2023.10.010}.
\newblock URL \url{https://www.sciencedirect.com/science/article/pii/S0169207023001097}.

\bibitem[Bengio et~al.(2021)Bengio, Jain, Korablyov, Precup, and Bengio]{bengio2021flow}
Bengio, E., Jain, M., Korablyov, M., Precup, D., and Bengio, Y.
\newblock Flow network based generative models for non-iterative diverse candidate generation.
\newblock \emph{Advances in Neural Information Processing Systems}, 34:\penalty0 27381--27394, 2021.

\bibitem[Bingham et~al.(2019)Bingham, Chen, Jankowiak, Obermeyer, Pradhan, Karaletsos, Singh, Szerlip, Horsfall, and Goodman]{bingham2019pyro}
Bingham, E., Chen, J.~P., Jankowiak, M., Obermeyer, F., Pradhan, N., Karaletsos, T., Singh, R., Szerlip, P., Horsfall, P., and Goodman, N.~D.
\newblock Pyro: Deep universal probabilistic programming.
\newblock \emph{The Journal of Machine Learning Research}, 20\penalty0 (1):\penalty0 973--978, 2019.

\bibitem[Bischl et~al.(2019)Bischl, Casalicchio, Feurer, Hutter, Lang, Mantovani, van Rijn, and Vanschoren]{bischl2019openmlcc18}
Bischl, B., Casalicchio, G., Feurer, M., Hutter, F., Lang, M., Mantovani, R.~G., van Rijn, J.~N., and Vanschoren, J.
\newblock Openml benchmarking suites.
\newblock \emph{arXiv:1708.03731v2 [stat.ML]}, 2019.

\bibitem[Bishop \& Nasrabadi(2006)Bishop and Nasrabadi]{bishop2006pattern}
Bishop, C.~M. and Nasrabadi, N.~M.
\newblock \emph{Pattern recognition and machine learning}, volume~4.
\newblock Springer, 2006.

\bibitem[Bitzer et~al.(2023)Bitzer, Meister, and Zimmer]{bitzer2023amortized}
Bitzer, M., Meister, M., and Zimmer, C.
\newblock Amortized inference for gaussian process hyperparameters of structured kernels.
\newblock \emph{arXiv preprint arXiv:2306.09819}, 2023.

\bibitem[Blei et~al.(2017)Blei, Kucukelbir, and McAuliffe]{blei2017variational}
Blei, D.~M., Kucukelbir, A., and McAuliffe, J.~D.
\newblock Variational inference: A review for statisticians.
\newblock \emph{Journal of the American statistical Association}, 112\penalty0 (518):\penalty0 859--877, 2017.

\bibitem[Brooks(1998)]{brooks1998markov}
Brooks, S.
\newblock Markov chain monte carlo method and its application.
\newblock \emph{Journal of the royal statistical society: series D (the Statistician)}, 47\penalty0 (1):\penalty0 69--100, 1998.

\bibitem[Carpenter et~al.(2017)Carpenter, Gelman, Hoffman, Lee, Goodrich, Betancourt, Brubaker, Guo, Li, and Riddell]{carpenter2017stan}
Carpenter, B., Gelman, A., Hoffman, M.~D., Lee, D., Goodrich, B., Betancourt, M., Brubaker, M.~A., Guo, J., Li, P., and Riddell, A.
\newblock Stan: A probabilistic programming language.
\newblock \emph{Journal of statistical software}, 76, 2017.

\bibitem[Chauhan et~al.(2023)Chauhan, Zhou, Lu, Molaei, and Clifton]{chauhan2023hyperreview}
Chauhan, V.~K., Zhou, J., Lu, P., Molaei, S., and Clifton, D.~A.
\newblock A brief review of hypernetworks in deep learning.
\newblock \emph{arXiv preprint arxiv:2306.06955}, 2023.

\bibitem[Chen et~al.(2023)Chen, Garnett, and Montgomery]{chen2023-pollaggregation}
Chen, Y., Garnett, R., and Montgomery, J.~M.
\newblock Polls, context, and time: A dynamic hierarchical bayesian forecasting model for us senate elections.
\newblock \emph{Political Analysis}, 31\penalty0 (1):\penalty0 113–133, 2023.
\newblock \doi{10.1017/pan.2021.42}.

\bibitem[Cooper et~al.(2020)Cooper, Mondal, and Antonopoulos]{Cooper2020-covidmodel}
Cooper, I., Mondal, A., and Antonopoulos, C.~G.
\newblock A sir model assumption for the spread of covid-19 in different communities.
\newblock \emph{Chaos, Solitons \& Fractals}, 139:\penalty0 110057, 2020.
\newblock ISSN 0960-0779.
\newblock \doi{https://doi.org/10.1016/j.chaos.2020.110057}.
\newblock URL \url{https://www.sciencedirect.com/science/article/pii/S0960077920304549}.

\bibitem[Cranmer et~al.(2020)Cranmer, Brehmer, and Louppe]{cranmer2020sbireview}
Cranmer, K., Brehmer, J., and Louppe, G.
\newblock The frontier of simulation-based inference.
\newblock \emph{Proceedings of the National Academy of Sciences}, 117\penalty0 (48):\penalty0 30055–30062, May 2020.
\newblock ISSN 1091-6490.
\newblock \doi{10.1073/pnas.1912789117}.
\newblock URL \url{http://dx.doi.org/10.1073/pnas.1912789117}.

\bibitem[Dinh et~al.(2017)Dinh, Sohl-Dickstein, and Bengio]{Dinh2017rnvp}
Dinh, L., Sohl-Dickstein, J., and Bengio, S.
\newblock Density estimation using real {NVP}.
\newblock \emph{5th International Conference on Learning Representations, ICLR 2017 - Conference Track Proceedings}, 2017.
\newblock URL \url{http://arxiv.org/abs/1605.08803}.

\bibitem[Finn et~al.(2017)Finn, Abbeel, and Levine]{finn2017model}
Finn, C., Abbeel, P., and Levine, S.
\newblock Model-agnostic meta-learning for fast adaptation of deep networks.
\newblock In \emph{International conference on machine learning}, pp.\  1126--1135. PMLR, 2017.

\bibitem[Fischer et~al.(2023)Fischer, Feurer, and Bischl]{fischer2023openmlctr23}
Fischer, S.~F., Feurer, M., and Bischl, B.
\newblock Open{ML}-{CTR}23 {\textendash} a curated tabular regression benchmarking suite.
\newblock In \emph{AutoML Conference 2023 (Workshop)}, 2023.
\newblock URL \url{https://openreview.net/forum?id=HebAOoMm94}.

\bibitem[Gardner et~al.(2018)Gardner, Pleiss, Weinberger, Bindel, and Wilson]{gardner2018gpytorch}
Gardner, J., Pleiss, G., Weinberger, K.~Q., Bindel, D., and Wilson, A.~G.
\newblock Gpytorch: Blackbox matrix-matrix gaussian process inference with gpu acceleration.
\newblock \emph{Advances in neural information processing systems}, 31, 2018.

\bibitem[Garg et~al.(2022)Garg, Tsipras, Liang, and Valiant]{garg2022can}
Garg, S., Tsipras, D., Liang, P.~S., and Valiant, G.
\newblock What can transformers learn in-context? a case study of simple function classes.
\newblock \emph{Advances in Neural Information Processing Systems}, 35:\penalty0 30583--30598, 2022.

\bibitem[Garnelo et~al.(2018{\natexlab{a}})Garnelo, Rosenbaum, Maddison, Ramalho, Saxton, Shanahan, Teh, Rezende, and Eslami]{garnelo2018conditional}
Garnelo, M., Rosenbaum, D., Maddison, C., Ramalho, T., Saxton, D., Shanahan, M., Teh, Y.~W., Rezende, D., and Eslami, S.~A.
\newblock Conditional neural processes.
\newblock In \emph{International conference on machine learning}, pp.\  1704--1713. PMLR, 2018{\natexlab{a}}.

\bibitem[Garnelo et~al.(2018{\natexlab{b}})Garnelo, Schwarz, Rosenbaum, Viola, Rezende, Eslami, and Teh]{garnelo2018neural}
Garnelo, M., Schwarz, J., Rosenbaum, D., Viola, F., Rezende, D.~J., Eslami, S., and Teh, Y.~W.
\newblock Neural processes.
\newblock \emph{arXiv preprint arXiv:1807.01622}, 2018{\natexlab{b}}.

\bibitem[Geffner et~al.(2023)Geffner, Papamakarios, and Mnih]{geffner2023compositional}
Geffner, T., Papamakarios, G., and Mnih, A.
\newblock Compositional score modeling for simulation-based inference.
\newblock 2023.

\bibitem[Gelfand(2000)]{gelfand2000gibbs}
Gelfand, A.~E.
\newblock Gibbs sampling.
\newblock \emph{Journal of the American statistical Association}, 95\penalty0 (452):\penalty0 1300--1304, 2000.

\bibitem[Gelman et~al.(2013)Gelman, Carlin, Stern, Dunson, Vehtari, and Rubin]{Gelman2013-bayesdata}
Gelman, A., Carlin, J.~B., Stern, H.~S., Dunson, D.~B., Vehtari, A., and Rubin, D.~B.
\newblock \emph{{Bayesian Data Analysis, Third Edition}}.
\newblock CRC Press, November 2013.
\newblock ISBN 9781439840955.
\newblock URL \url{https://play.google.com/store/books/details?id=ZXL6AQAAQBAJ}.

\bibitem[Glaeser et~al.(2021)Glaeser, Nogales, and Chiu]{Glaeser2021-cryoem}
Glaeser, R.~M., Nogales, E., and Chiu, W.
\newblock \emph{Single-particle Cryo-EM of Biological Macromolecules}.
\newblock 2053-2563. IOP Publishing, 2021.
\newblock ISBN 978-0-7503-3039-8.
\newblock \doi{10.1088/978-0-7503-3039-8}.
\newblock URL \url{https://dx.doi.org/10.1088/978-0-7503-3039-8}.

\bibitem[Gordon et~al.(2019)Gordon, Bruinsma, Foong, Requeima, Dubois, and Turner]{gordon2019convolutional}
Gordon, J., Bruinsma, W.~P., Foong, A.~Y., Requeima, J., Dubois, Y., and Turner, R.~E.
\newblock Convolutional conditional neural processes.
\newblock \emph{arXiv preprint arXiv:1910.13556}, 2019.

\bibitem[Grant et~al.(2018)Grant, Finn, Levine, Darrell, and Griffiths]{grant2018recasting}
Grant, E., Finn, C., Levine, S., Darrell, T., and Griffiths, T.
\newblock Recasting gradient-based meta-learning as hierarchical bayes.
\newblock \emph{arXiv preprint arXiv:1801.08930}, 2018.

\bibitem[Higgins et~al.(2017)Higgins, Matthey, Pal, Burgess, Glorot, Botvinick, Mohamed, and Lerchner]{higgins2017betavae}
Higgins, I., Matthey, L., Pal, A., Burgess, C., Glorot, X., Botvinick, M., Mohamed, S., and Lerchner, A.
\newblock beta-{VAE}: Learning basic visual concepts with a constrained variational framework.
\newblock In \emph{International Conference on Learning Representations}, 2017.
\newblock URL \url{https://openreview.net/forum?id=Sy2fzU9gl}.

\bibitem[Hoffman et~al.(2013)Hoffman, Blei, Wang, and Paisley]{hoffman2013stochastic}
Hoffman, M.~D., Blei, D.~M., Wang, C., and Paisley, J.
\newblock Stochastic variational inference.
\newblock \emph{Journal of Machine Learning Research}, 2013.

\bibitem[Hoffman et~al.(2014)Hoffman, Gelman, et~al.]{hoffman2014no}
Hoffman, M.~D., Gelman, A., et~al.
\newblock The no-u-turn sampler: adaptively setting path lengths in hamiltonian monte carlo.
\newblock \emph{J. Mach. Learn. Res.}, 15\penalty0 (1):\penalty0 1593--1623, 2014.

\bibitem[Hollmann et~al.(2022)Hollmann, M{\"u}ller, Eggensperger, and Hutter]{hollmann2022tabpfn}
Hollmann, N., M{\"u}ller, S., Eggensperger, K., and Hutter, F.
\newblock Tabpfn: A transformer that solves small tabular classification problems in a second.
\newblock \emph{arXiv preprint arXiv:2207.01848}, 2022.

\bibitem[Hospedales et~al.(2022)Hospedales, Antoniou, Micaelli, and Storkey]{hospedales2022metareview}
Hospedales, T., Antoniou, A., Micaelli, P., and Storkey, A.
\newblock Meta-learning in neural networks: A survey.
\newblock \emph{IEEE Transactions on Pattern Analysis \& Machine Intelligence}, 44\penalty0 (09):\penalty0 5149--5169, sep 2022.
\newblock ISSN 1939-3539.
\newblock \doi{10.1109/TPAMI.2021.3079209}.

\bibitem[Kim et~al.(2019)Kim, Mnih, Schwarz, Garnelo, Eslami, Rosenbaum, Vinyals, and Teh]{kim2019attentive}
Kim, H., Mnih, A., Schwarz, J., Garnelo, M., Eslami, A., Rosenbaum, D., Vinyals, O., and Teh, Y.~W.
\newblock Attentive neural processes.
\newblock \emph{arXiv preprint arXiv:1901.05761}, 2019.

\bibitem[Kingma \& Ba(2014)Kingma and Ba]{kingma2014adam}
Kingma, D.~P. and Ba, J.
\newblock Adam: A method for stochastic optimization.
\newblock \emph{arXiv preprint arXiv:1412.6980}, 2014.

\bibitem[Kingma \& Dhariwal(2018)Kingma and Dhariwal]{kingma2018glow}
Kingma, D.~P. and Dhariwal, P.
\newblock Glow: Generative flow with invertible 1x1 convolutions.
\newblock \emph{Advances in neural information processing systems}, 31, 2018.

\bibitem[Kingma \& Welling(2013)Kingma and Welling]{kingma2013auto}
Kingma, D.~P. and Welling, M.
\newblock Auto-encoding variational bayes.
\newblock \emph{arXiv preprint arXiv:1312.6114}, 2013.

\bibitem[Kingma et~al.(2019)Kingma, Welling, et~al.]{kingma2019introduction}
Kingma, D.~P., Welling, M., et~al.
\newblock An introduction to variational autoencoders.
\newblock \emph{Foundations and Trends{\textregistered} in Machine Learning}, 12\penalty0 (4):\penalty0 307--392, 2019.

\bibitem[Kobyzev et~al.(2020)Kobyzev, Prince, and Brubaker]{kobyzev2020normalizing}
Kobyzev, I., Prince, S.~J., and Brubaker, M.~A.
\newblock Normalizing flows: An introduction and review of current methods.
\newblock \emph{IEEE transactions on pattern analysis and machine intelligence}, 43\penalty0 (11):\penalty0 3964--3979, 2020.

\bibitem[Koch et~al.(2015)Koch, Zemel, Salakhutdinov, et~al.]{koch2015siamese}
Koch, G., Zemel, R., Salakhutdinov, R., et~al.
\newblock Siamese neural networks for one-shot image recognition.
\newblock In \emph{ICML deep learning workshop}, volume~2. Lille, 2015.

\bibitem[Krueger et~al.(2017)Krueger, Huang, Islam, Turner, Lacoste, and Courville]{krueger2017hyperbayesian}
Krueger, D., Huang, C.-W., Islam, R., Turner, R., Lacoste, A., and Courville, A.
\newblock Bayesian hypernetworks.
\newblock \emph{arXiv preprint arxiv:1710.04759}, 2017.

\bibitem[Lee et~al.(2019)Lee, Lee, Kim, Kosiorek, Choi, and Teh]{lee2019set}
Lee, J., Lee, Y., Kim, J., Kosiorek, A., Choi, S., and Teh, Y.~W.
\newblock Set transformer: A framework for attention-based permutation-invariant neural networks.
\newblock In \emph{International conference on machine learning}, pp.\  3744--3753. PMLR, 2019.

\bibitem[Lin et~al.(2020)Lin, Schmidt, and Khan]{lin2020handling}
Lin, W., Schmidt, M., and Khan, M.~E.
\newblock Handling the positive-definite constraint in the bayesian learning rule.
\newblock In \emph{International conference on machine learning}, pp.\  6116--6126. PMLR, 2020.

\bibitem[Liu et~al.(2020)Liu, Sun, Ramadge, and Adams]{liu2020task}
Liu, S., Sun, X., Ramadge, P.~J., and Adams, R.~P.
\newblock Task-agnostic amortized inference of gaussian process hyperparameters.
\newblock \emph{Advances in Neural Information Processing Systems}, 33:\penalty0 21440--21452, 2020.

\bibitem[Lorch et~al.(2022)Lorch, Sussex, Rothfuss, Krause, and Sch{\"o}lkopf]{lorch2022amortized}
Lorch, L., Sussex, S., Rothfuss, J., Krause, A., and Sch{\"o}lkopf, B.
\newblock Amortized inference for causal structure learning.
\newblock \emph{Advances in Neural Information Processing Systems}, 35:\penalty0 13104--13118, 2022.

\bibitem[Minka(2013)]{minka2013expectation}
Minka, T.~P.
\newblock Expectation propagation for approximate bayesian inference.
\newblock \emph{arXiv preprint arXiv:1301.2294}, 2013.

\bibitem[Morris(2013)]{morris2013recognition}
Morris, Q.
\newblock Recognition networks for approximate inference in bn20 networks.
\newblock \emph{arXiv preprint arXiv:1301.2295}, 2013.

\bibitem[M{\"u}ller et~al.(2021)M{\"u}ller, Hollmann, Arango, Grabocka, and Hutter]{muller2021transformers}
M{\"u}ller, S., Hollmann, N., Arango, S.~P., Grabocka, J., and Hutter, F.
\newblock Transformers can do bayesian inference.
\newblock \emph{arXiv preprint arXiv:2112.10510}, 2021.

\bibitem[Paige \& Wood(2016)Paige and Wood]{paige2016inference}
Paige, B. and Wood, F.
\newblock Inference networks for sequential monte carlo in graphical models.
\newblock In \emph{International Conference on Machine Learning}, pp.\  3040--3049. PMLR, 2016.

\bibitem[Pakman et~al.(2020)Pakman, Wang, Mitelut, Lee, and Paninski]{pakman2020neural}
Pakman, A., Wang, Y., Mitelut, C., Lee, J., and Paninski, L.
\newblock Neural clustering processes.
\newblock In \emph{International Conference on Machine Learning}, pp.\  7455--7465. PMLR, 2020.

\bibitem[Papamakarios et~al.(2021)Papamakarios, Nalisnick, Rezende, Mohamed, and Lakshminarayanan]{papamakarios2021normalizing}
Papamakarios, G., Nalisnick, E., Rezende, D.~J., Mohamed, S., and Lakshminarayanan, B.
\newblock Normalizing flows for probabilistic modeling and inference.
\newblock \emph{The Journal of Machine Learning Research}, 22\penalty0 (1):\penalty0 2617--2680, 2021.

\bibitem[Radev et~al.(2020)Radev, Mertens, Voss, Ardizzone, and K{\"o}the]{radev2020bayesflow}
Radev, S.~T., Mertens, U.~K., Voss, A., Ardizzone, L., and K{\"o}the, U.
\newblock Bayesflow: Learning complex stochastic models with invertible neural networks.
\newblock \emph{IEEE transactions on neural networks and learning systems}, 33\penalty0 (4):\penalty0 1452--1466, 2020.

\bibitem[Rezende \& Mohamed(2015)Rezende and Mohamed]{rezende2015variational}
Rezende, D. and Mohamed, S.
\newblock Variational inference with normalizing flows.
\newblock In \emph{International conference on machine learning}, pp.\  1530--1538. PMLR, 2015.

\bibitem[Rezende et~al.(2014)Rezende, Mohamed, and Wierstra]{rezende2014stochastic}
Rezende, D.~J., Mohamed, S., and Wierstra, D.
\newblock Stochastic backpropagation and approximate inference in deep generative models.
\newblock In \emph{International conference on machine learning}, pp.\  1278--1286. PMLR, 2014.

\bibitem[Simpson et~al.(2021)Simpson, Davies, Lalchand, Vullo, Durrande, and Rasmussen]{simpson2021kernel}
Simpson, F., Davies, I., Lalchand, V., Vullo, A., Durrande, N., and Rasmussen, C.~E.
\newblock Kernel identification through transformers.
\newblock \emph{Advances in Neural Information Processing Systems}, 34:\penalty0 10483--10495, 2021.

\bibitem[Song et~al.(2020)Song, Sohl-Dickstein, Kingma, Kumar, Ermon, and Poole]{song2020score}
Song, Y., Sohl-Dickstein, J., Kingma, D.~P., Kumar, A., Ermon, S., and Poole, B.
\newblock Score-based generative modeling through stochastic differential equations.
\newblock \emph{arXiv preprint arXiv:2011.13456}, 2020.

\bibitem[Stuhlm{\"u}ller et~al.(2013)Stuhlm{\"u}ller, Taylor, and Goodman]{stuhlmuller2013learning}
Stuhlm{\"u}ller, A., Taylor, J., and Goodman, N.
\newblock Learning stochastic inverses.
\newblock \emph{Advances in neural information processing systems}, 26, 2013.

\bibitem[Sun et~al.(2017)Sun, Ozay, and Okatani]{sun2017hypercnn}
Sun, Z., Ozay, M., and Okatani, T.
\newblock Hypernetworks with statistical filtering for defending adversarial examples.
\newblock \emph{arXiv preprint arxiv:1711.01791}, 2017.

\bibitem[Sung et~al.(2018)Sung, Yang, Zhang, Xiang, Torr, and Hospedales]{sung2018learning}
Sung, F., Yang, Y., Zhang, L., Xiang, T., Torr, P.~H., and Hospedales, T.~M.
\newblock Learning to compare: Relation network for few-shot learning.
\newblock In \emph{Proceedings of the IEEE conference on computer vision and pattern recognition}, pp.\  1199--1208, 2018.

\bibitem[Tang et~al.(2020)Tang, Zhou, Wang, Purkayastha, Zhang, He, Wang, and Song]{Tang2020-diseasemodels}
Tang, L., Zhou, Y., Wang, L., Purkayastha, S., Zhang, L., He, J., Wang, F., and Song, P. X.-K.
\newblock A review of multi-compartment infectious disease models.
\newblock \emph{International Statistical Review}, 88\penalty0 (2):\penalty0 462--513, 2020.
\newblock \doi{https://doi.org/10.1111/insr.12402}.
\newblock URL \url{https://onlinelibrary.wiley.com/doi/abs/10.1111/insr.12402}.

\bibitem[Vargas et~al.(2023)Vargas, Grathwohl, and Doucet]{vargas2023denoising}
Vargas, F., Grathwohl, W., and Doucet, A.
\newblock Denoising diffusion samplers.
\newblock \emph{arXiv preprint arXiv:2302.13834}, 2023.

\bibitem[Vaswani et~al.(2017)Vaswani, Shazeer, Parmar, Uszkoreit, Jones, Gomez, Kaiser, and Polosukhin]{vaswani2017attention}
Vaswani, A., Shazeer, N., Parmar, N., Uszkoreit, J., Jones, L., Gomez, A.~N., Kaiser, {\L}., and Polosukhin, I.
\newblock Attention is all you need.
\newblock \emph{Advances in neural information processing systems}, 30, 2017.

\bibitem[Vinyals et~al.(2016)Vinyals, Blundell, Lillicrap, Wierstra, et~al.]{vinyals2016matching}
Vinyals, O., Blundell, C., Lillicrap, T., Wierstra, D., et~al.
\newblock Matching networks for one shot learning.
\newblock \emph{Advances in neural information processing systems}, 29, 2016.

\bibitem[Von~Oswald et~al.(2023)Von~Oswald, Niklasson, Randazzo, Sacramento, Mordvintsev, Zhmoginov, and Vladymyrov]{von2023transformers}
Von~Oswald, J., Niklasson, E., Randazzo, E., Sacramento, J., Mordvintsev, A., Zhmoginov, A., and Vladymyrov, M.
\newblock Transformers learn in-context by gradient descent.
\newblock In \emph{International Conference on Machine Learning}, pp.\  35151--35174. PMLR, 2023.

\bibitem[von Oswald et~al.(2023)von Oswald, Niklasson, Schlegel, Kobayashi, Zucchet, Scherrer, Miller, Sandler, Vladymyrov, Pascanu, et~al.]{von2023uncovering}
von Oswald, J., Niklasson, E., Schlegel, M., Kobayashi, S., Zucchet, N., Scherrer, N., Miller, N., Sandler, M., Vladymyrov, M., Pascanu, R., et~al.
\newblock Uncovering mesa-optimization algorithms in transformers.
\newblock \emph{arXiv preprint arXiv:2309.05858}, 2023.

\bibitem[Welling \& Teh(2011)Welling and Teh]{welling2011bayesian}
Welling, M. and Teh, Y.~W.
\newblock Bayesian learning via stochastic gradient langevin dynamics.
\newblock In \emph{Proceedings of the 28th international conference on machine learning (ICML-11)}, pp.\  681--688, 2011.

\bibitem[Zaheer et~al.(2017)Zaheer, Kottur, Ravanbhakhsh, P{\'o}czos, Salakhutdinov, and Smola]{Zaheer2017deepsets}
Zaheer, M., Kottur, S., Ravanbhakhsh, S., P{\'o}czos, B., Salakhutdinov, R., and Smola, A.~J.
\newblock Deep sets.
\newblock In \emph{Advances in Neural Information Processing Systems}, volume 2017-December, 2017.

\bibitem[Zhang \& Chen(2021)Zhang and Chen]{zhang2021path}
Zhang, Q. and Chen, Y.
\newblock Path integral sampler: a stochastic control approach for sampling.
\newblock \emph{arXiv preprint arXiv:2111.15141}, 2021.

\bibitem[Štrumbelj et~al.(2023)Štrumbelj, Bouchard-Côté, Corander, Gelman, Rue, Murray, Pesonen, Plummer, and Vehtari]{StatSoftReview}
Štrumbelj, E., Bouchard-Côté, A., Corander, J., Gelman, A., Rue, H., Murray, L., Pesonen, H., Plummer, M., and Vehtari, A.
\newblock Past, present, and future of software for bayesian inference, 2023.
\newblock URL \url{http://hdl.handle.net/10754/694575}.

\end{thebibliography}
\bibliographystyle{icml2025}

\clearpage
\appendix
\onecolumn
\section*{\LARGE Appendix}
\section{Related Work}
\label{appdx:related_work}
In this section, we draw parallels of our work to various approaches that have been proposed to tackle the problem of either providing a good initialization for different tasks, performing implicit optimization to model predictive distributions for new tasks, or estimating the posterior through a different objective.

\subsection{Variational Autoencoders}
VAEs~\citep{kingma2013auto,rezende2014stochastic,rezende2015variational,kingma2019introduction} are latent variable models which model observations $\vx$ conditioned on latent variables $\vz$ through the joint distribution $p_\theta(\vx, \vz) = p_\theta(\vx | \vz) p(\vz)$ where $p(\vz)$ is generally chosen as $\mathcal{N}(\mathbf{0}, \mathbf{I})$. Training the model is done through VI where $q_\varphi(\vz)$ is obtained by explicit amortization over the data point, that is, $q_\varphi(\vz | \vx) = \gN\left(\mmu_\varphi(\vx), \mSigma_\varphi(\vx)\right)$. Training this system on a dataset $\gD$ is done by similarly optimizing the Evidence Lower-Bound, which boils down to the following optimization problem
\begin{align}
    \arg\max_{\theta, \varphi} \sE_{\vx \sim \gD}\sE_{\vz \sim q(\cdot | \vx)}\left[\log \frac{p_\theta(\vx, \vz)}{q_\varphi(\vz | \vx)}\right]
\end{align}
This objective can easily be optimized using gradient-based learning and the reparameterization trick. While typically, a diagonal Gaussian distribution is considered for $q_\varphi$, more complex distributions utilizing normalizing flows can also be used.

\subsection{Hypernetworks}
Hypernetworks are neural networks that generate weights for another neural network, used in tasks such as uncertainty quantification, zero-shot learning, etc. We refer for a comprehensive overview to~\cite{chauhan2023hyperreview}. Based on experiments on predicting the weights of a compact MLP (section \ref{sec:experiments}), our work shows similarities with studies in this area but also has significant differences. 
Regarding uncertainty quantification, hypernetworks are instrumental in creating an ensemble of models by generating multiple weight vectors for the primary network. Each model within this ensemble possesses distinct parameter configurations, enabling robust estimation of uncertainty in model predictions. This feature is precious in safety-critical domains like healthcare, where confidence in predictions is essential. Multiple weight sets can be generated through techniques like dropout within hypernetworks or sampling from a noise distribution.
The latter~\citep{krueger2017hyperbayesian} is based on a Bayesian framework where weights can be sampled using invertible network architecture, such as normalizing flows. However, while we amortize posterior inference, the weights sampled from the hypernetwork are not conditioned on information from the currently observed input data during inference time but indirectly solely on the dataset available during training, and retraining would need to be done given a new dataset. 
Departing from the Bayesian framework, \cite{sun2017hypercnn} have shown data-specific discriminative weight prediction, which aligns well with their specific objective of defending a convolutional neural network against adversarial attacks.
Combining the ability to sample a new set of weights dataset-specifically but also handling dataset exchangeability, even in the more realistic case of missing information, our work has a distinctly different focus but also can be seen as an extension to hypernetwork research. 

\subsection{In-Context Learning}
Amortized inference has close links to in-context learning (ICL), which has been gaining popularity, especially in natural language modeling. Various works show how in-context learning can be seen as performing implicit optimization based on the context examples, with some constructions showing exact equivalence with gradient descent in linear regression~\citep{von2023transformers,von2023uncovering}. Other works have shown how such systems can be seen as implicitly modeling the Bayesian posterior predictive distribution~\citep{muller2021transformers}. In a similar vein, there have been additional works aimed at directly modeling the posterior predictive distribution by providing the training data as ``context" to a Transformer model and training it based on the maximum log-likelihood principle~\citep{hollmann2022tabpfn}. While such approaches have been seeing tremendous success, they cannot be directly applied to cases where we care about and want to analyze the solution space as the solution space is only modeled implicitly, and thus, recovering it is not possible. For example, if our goal is to learn a linear regression model, an ICL model could end up learning a nonlinear model and would provide no information about the actual parameters used for prediction. As opposed to this, we obtain parameters explicitly. We thus can answer questions like the relevance of a particular feature (which corresponds to its weight in the output, and we know the weight vector explicitly). Even further, many systems grounded in physics and economics only admit a constrained solution space; for example, the movement of a human arm lies on a particular manifold, or the configuration of molecules and proteins cannot be arbitrary. Thus, performing predictions through an implicit solution space, which may violate several constraints, is not ideal. Furthermore, explicitly modeling the solution space and encoding the constraints present can be done through the prior and the parametric distribution used for modeling. 

\begin{table}
    \centering
    \small
    \def\arraystretch{1.25}
    \setlength{\tabcolsep}{3pt}
\begin{tabular}{lcr cc cccc}
        \toprule
         &  &  & \multicolumn{6}{c}{\textit{$L_2$ Loss} ($\downarrow$)} \\
        \cmidrule(lr){4-9}
        \textbf{Objective} & $q_\varphi$ & \textbf{Model} & \multicolumn{2}{c}{\textbf{Gaussian}} & \multicolumn{4}{c}{\textbf{GMM}} \\
        \cmidrule(lr){4-5}\cmidrule(lr){6-9}
        & & & \textit{2D} & \textit{100D} & \textit{2D-2cl} & \textit{2D-5cl} & \textit{5D-2cl} & \textit{5D-5cl} \\
        \midrule
\multirow{4}{*}{Baseline} & - & Random & $5.839$\sstd{$0.015$} & $301.065$\sstd{$0.346$} & $1.887$\sstd{$0.031$} & $0.730$\sstd{$0.004$} & $5.001$\sstd{$0.037$} & $1.670$\sstd{$0.008$} \\
& - & Optimization & $1.989$\sstd{$0.000$} & $101.243$\sstd{$0.000$} & $0.169$\sstd{$0.000$} & $0.119$\sstd{$0.001$} & $0.425$\sstd{$0.000$} & $0.308$\sstd{$0.000$} \\
& - & Langevin & $2.013$\sstd{$0.004$} & $102.346$\sstd{$0.031$} & $0.173$\sstd{$0.001$} & $0.125$\sstd{$0.001$} & $0.448$\sstd{$0.009$} & $0.352$\sstd{$0.005$} \\
& - & HMC & $2.018$\sstd{$0.008$} & $102.413$\sstd{$0.028$} & $0.174$\sstd{$0.001$} & $0.135$\sstd{$0.001$} & $0.479$\sstd{$0.007$} & $0.449$\sstd{$0.002$} \\
\cmidrule{2-9}

\multirow{3}{*}{Fwd-KL} & \multirow{6}{*}{\rotatebox[origin=c]{90}{Gaussian}} & GRU &$2.014$\sstd{$0.001$} & $102.641$\sstd{$0.011$} & $0.921$\sstd{$0.013$} & $0.522$\sstd{$0.001$} & $2.430$\sstd{$0.034$} & $1.235$\sstd{$0.011$} \\
\bottomrule
    \end{tabular}
    \caption{\textbf{Fixed-Dimensional}. Results for estimating the mean of a Gaussian (Gaussian) and means of a Gaussian mixture model (GMM) with the expected $L_2$ loss according to the posterior predictive as metric.}
    \vspace{-4mm}
    \label{tab:apdx_gaussian}
\end{table}

\begin{table*}[t]
    \centering
    \small
    \def\arraystretch{1.25}
    \setlength{\tabcolsep}{3pt}
    \begin{tabular}{lcr cc cccc}
        \toprule
         &  &  & \multicolumn{2}{c}{\textit{$L_2$ Loss} ($\downarrow$)} & \multicolumn{2}{c}{\textit{Accuracy} ($\uparrow$)} \\
         \cmidrule(lr){4-5}\cmidrule(lr){6-9}
         \textbf{Objective} & $q_\varphi$ & \textbf{Model} & \multicolumn{2}{c}{\textbf{Linear Regression}} & \multicolumn{4}{c}{\textbf{Linear Classification}} \\
         \cmidrule(lr){4-5}\cmidrule(lr){6-9}
         & & & \textit{2D} & \textit{100D} & \textit{2D-2cl} & \textit{2D-5cl} & \textit{100D-2cl} & \textit{100D-5cl} \\
         \midrule

\multirow{4}{*}{Baseline} & - & Random & $4.178$\sstd{$0.018$} & $202.601$\sstd{$0.321$} & $50.498$\sstd{$0.357$} & $19.891$\sstd{$0.028$} & $50.046$\sstd{$0.047$} & $20.054$\sstd{$0.053$} \\
& - & Optimization & $0.257$\sstd{$0.000$} & $25.083$\sstd{$0.006$} & $96.982$\sstd{$0.000$} & $93.449$\sstd{$0.002$} & $70.258$\sstd{$0.012$} & $41.338$\sstd{$0.012$} \\
& - & Langevin & $0.263$\sstd{$0.002$} & $23.340$\sstd{$0.689$} & $95.034$\sstd{$0.412$} & $88.277$\sstd{$0.290$} & $65.123$\sstd{$0.370$} & $32.544$\sstd{$0.422$} \\
& - & HMC & $0.263$\sstd{$0.001$} & $18.659$\sstd{$0.189$} & $92.659$\sstd{$0.344$} & $82.169$\sstd{$0.518$} & $62.145$\sstd{$0.245$} & $29.582$\sstd{$0.371$} \\
\cmidrule{2-9}

\multirow{3}{*}{Fwd-KL} & \multirow{6}{*}{\rotatebox[origin=c]{90}{Gaussian}} & GRU & \highlight{$0.264$\sstd{$0.001$}} & $124.823$\sstd{$0.135$} & $81.170$\sstd{$0.389$} & $71.170$\sstd{$0.275$} & $59.740$\sstd{$0.102$} & $23.042$\sstd{$0.246$} \\
\bottomrule
    \end{tabular}
    \caption{\textbf{Fixed-Dimensional}. Results for estimating the parameters of linear regression (LR) and classification (LC) models with the expected $L_2$ loss and accuracy according to the posterior predictive as metrics.}
    \vspace{-4mm}
    \label{tab:}
\end{table*}

\begin{table*}[t]
    \centering
    \small
    \def\arraystretch{1.25}
    \setlength{\tabcolsep}{5pt}
    \begin{tabular}{lcr cccc}
        \toprule
         &  &  & \multicolumn{4}{c}{\textit{$L_2$ Loss} ($\downarrow$)} \\
         \cmidrule(lr){4-7}
         \textbf{Objective} & $q_\varphi$ & \textbf{Model} & \multicolumn{4}{c}{\textbf{Nonlinear Regression} $|$ \textbf{ReLU}} \\
         \cmidrule(lr){4-7}
         & & & \multicolumn{2}{c}{\textit{1-layer}} & \multicolumn{2}{c}{\textit{2-layers}} \\
         \cmidrule(lr){4-5}\cmidrule(lr){6-7}
         & & & \textit{1D} & \textit{25D} & \textit{1D} & \textit{25D} \\
         \midrule
\multirow{4}{*}{Baseline} & - & Random & $65.936$\sstd{$0.913$} & $831.595$\sstd{$8.696$} & $1029.407$\sstd{$11.542$} & $12067.691$\sstd{$183.598$} \\
& - & Optimization & $0.360$\sstd{$0.001$} & $103.967$\sstd{$0.110$} & $2.370$\sstd{$0.015$} & $1894.574$\sstd{$4.266$} \\
& - & Langevin & $0.308$\sstd{$0.000$} & $132.391$\sstd{$0.992$} & \textsc{N/A} & \textsc{N/A} \\
& - & HMC & $0.374$\sstd{$0.002$} & $98.061$\sstd{$0.730$} & $22.314$\sstd{$0.814$} & $3903.510$\sstd{$5.377$} \\
\bottomrule
    \end{tabular}
    \caption{\textbf{Fixed-Dimensional}. Results for estimating the parameters of nonlinear regression models with ReLU activation function, with the expected $L_2$ loss according to the posterior predictive as metric.}
    \vspace{-4mm}
    \label{tab:}
\end{table*}
\begin{table*}[t]
    \centering
    \small
    \def\arraystretch{1.25}
    \setlength{\tabcolsep}{5pt}
    \begin{tabular}{lcr cccc}
        \toprule
         &  &  & \multicolumn{4}{c}{\textit{$L_2$ Loss} ($\downarrow$)} \\
         \cmidrule(lr){4-7}
         \textbf{Objective} & $q_\varphi$ & \textbf{Model} & \multicolumn{4}{c}{\textbf{Nonlinear Regression} $|$ \textbf{TanH}} \\
         \cmidrule(lr){4-7}
         & & & \multicolumn{2}{c}{\textit{1-layer}} & \multicolumn{2}{c}{\textit{2-layers}} \\
         \cmidrule(lr){4-5}\cmidrule(lr){6-7}
         & & & \textit{1D} & \textit{25D} & \textit{1D} & \textit{25D} \\
         \midrule

\multirow{4}{*}{Baseline} & - & Random & $31.448$\sstd{$0.186$} & $52.644$\sstd{$0.173$} & $52.735$\sstd{$1.122$} & $52.583$\sstd{$0.132$} \\
& - & Optimization & $0.366$\sstd{$0.001$} & $13.352$\sstd{$0.005$} & $0.651$\sstd{$0.002$} & $30.176$\sstd{$0.056$} \\
& - & Langevin & $0.296$\sstd{$0.003$} & $17.221$\sstd{$0.130$} & $0.363$\sstd{$0.003$} & $28.528$\sstd{$0.115$} \\
& - & HMC & $0.398$\sstd{$0.003$} & $12.607$\sstd{$0.192$} & $0.733$\sstd{$0.021$} & $23.571$\sstd{$0.346$} \\
\bottomrule
    \end{tabular}
    \caption{\textbf{Fixed-Dimensional}. Results for estimating the parameters of nonlinear regression models with TanH activation function, with the expected $L_2$ loss according to the posterior predictive as metric.}
    \vspace{-4mm}
    \label{tab:}
\end{table*}
\begin{table*}[t]
    \centering
    \small
    \def\arraystretch{1.25}
    \setlength{\tabcolsep}{5pt}
    \begin{tabular}{lcr cc cc}
        \toprule
         &  &  & \multicolumn{4}{c}{\textit{Accuracy} ($\uparrow$)} \\
         \cmidrule(lr){4-7}
         \textbf{Objective} & $q_\varphi$ & \textbf{Model} & \multicolumn{4}{c}{\textbf{Nonlinear Classification} $|$ \textbf{ReLU - 2 class}} \\
         \cmidrule(lr){4-7}
         & & & \multicolumn{2}{c}{\textit{1-layer}} & \multicolumn{2}{c}{\textit{2-layers}} \\
         \cmidrule(lr){4-5}\cmidrule(lr){6-7}
         & & & \textit{2D} & \textit{25D} & \textit{2D} & \textit{25D} \\
         \midrule
\multirow{4}{*}{Baseline} & - & Random & $50.306$\sstd{$0.590$} & $50.008$\sstd{$0.326$} & $50.394$\sstd{$0.190$} & $49.846$\sstd{$0.635$} \\
& - & Optimization & $96.879$\sstd{$0.028$} & $77.896$\sstd{$0.023$} & $96.770$\sstd{$0.062$} & $82.073$\sstd{$0.200$} \\
& - & Langevin & $95.971$\sstd{$0.313$} & $73.165$\sstd{$0.282$} & $96.645$\sstd{$0.101$} & $76.541$\sstd{$0.139$} \\
& - & HMC & $91.763$\sstd{$0.163$} & $70.395$\sstd{$0.110$} & $91.797$\sstd{$0.048$} & $76.445$\sstd{$0.372$} \\
\bottomrule
    \end{tabular}
    \caption{\textbf{Fixed-Dimensional}. Results for estimating the parameters of nonlinear classification models with ReLU activation function and two classes, with the expected accuracy according to the posterior predictive as metric.}
    \vspace{-4mm}
    \label{tab:}
\end{table*}
\begin{table*}[t]
    \centering
    \small
    \def\arraystretch{1.25}
    \setlength{\tabcolsep}{5pt}
    \begin{tabular}{lcr cc cc}
        \toprule
         &  &  & \multicolumn{4}{c}{\textit{Accuracy} ($\uparrow$)} \\
         \cmidrule(lr){4-7}
         \textbf{Objective} & $q_\varphi$ & \textbf{Model} & \multicolumn{4}{c}{\textbf{Nonlinear Classification} $|$ \textbf{ReLU - 5 class}} \\
         \cmidrule(lr){4-7}
         & & & \multicolumn{2}{c}{\textit{1-layer}} & \multicolumn{2}{c}{\textit{2-layers}} \\
         \cmidrule(lr){4-5}\cmidrule(lr){6-7}
         & & & \textit{2D} & \textit{25D} & \textit{2D} & \textit{25D} \\
         \midrule
\multirow{4}{*}{Baseline} & - & Random & $19.952$\sstd{$0.119$} & $20.026$\sstd{$0.062$} & $19.832$\sstd{$0.202$} & $19.985$\sstd{$0.091$} \\
& - & Optimization & $94.369$\sstd{$0.022$} & $60.589$\sstd{$0.064$} & $93.664$\sstd{$0.025$} & $60.824$\sstd{$0.023$} \\
& - & Langevin & $91.286$\sstd{$0.149$} & $50.845$\sstd{$0.366$} & $92.449$\sstd{$0.068$} & $52.216$\sstd{$0.170$} \\
& - & HMC & $81.387$\sstd{$0.742$} & $47.408$\sstd{$0.548$} & $81.098$\sstd{$0.246$} & $52.854$\sstd{$0.190$} \\
\bottomrule
    \end{tabular}
    \caption{\textbf{Fixed-Dimensional}. Results for estimating the parameters of nonlinear classification models with ReLU activation function and five classes, with the expected accuracy according to the posterior predictive as metric.}
    \vspace{-4mm}
    \label{tab:}
\end{table*}
\begin{table*}[t]
    \centering
    \small
    \def\arraystretch{1.25}
    \setlength{\tabcolsep}{5pt}
    \begin{tabular}{lcr cc cc}
        \toprule
         &  &  & \multicolumn{4}{c}{\textit{Accuracy} ($\uparrow$)} \\
         \cmidrule(lr){4-7}
         \textbf{Objective} & $q_\varphi$ & \textbf{Model} & \multicolumn{4}{c}{\textbf{Nonlinear Classification} $|$ \textbf{TanH - 2 class}} \\
         \cmidrule(lr){4-7}
         & & & \multicolumn{2}{c}{\textit{1-layer}} & \multicolumn{2}{c}{\textit{2-layers}} \\
         \cmidrule(lr){4-5}\cmidrule(lr){6-7}
         & & & \textit{2D} & \textit{25D} & \textit{2D} & \textit{25D} \\
         \midrule
\multirow{4}{*}{Baseline} & - & Random & $50.147$\sstd{$0.603$} & $49.963$\sstd{$0.083$} & $49.942$\sstd{$0.452$} & $49.978$\sstd{$0.113$} \\
& - & Optimization & $96.552$\sstd{$0.005$} & $75.228$\sstd{$0.029$} & $94.130$\sstd{$0.018$} & $69.052$\sstd{$0.029$} \\
& - & Langevin & $94.778$\sstd{$0.210$} & $68.787$\sstd{$0.258$} & $92.417$\sstd{$0.149$} & $62.722$\sstd{$0.180$} \\
& - & HMC & $91.674$\sstd{$0.200$} & $67.415$\sstd{$0.667$} & $88.218$\sstd{$0.202$} & $62.323$\sstd{$0.262$} \\
\bottomrule
    \end{tabular}
    \caption{\textbf{Fixed-Dimensional}. Results for estimating the parameters of nonlinear classification models with TanH activation function and two classes, with the expected accuracy according to the posterior predictive as metric.}
    \vspace{-4mm}
    \label{tab:}
\end{table*}
\begin{table*}[t]
    \centering
    \small
    \def\arraystretch{1.25}
    \setlength{\tabcolsep}{5pt}
    \begin{tabular}{lcr cc cc}
        \toprule
         &  &  & \multicolumn{4}{c}{\textit{Accuracy} ($\uparrow$)} \\
         \cmidrule(lr){4-7}
         \textbf{Objective} & $q_\varphi$ & \textbf{Model} & \multicolumn{4}{c}{\textbf{Nonlinear Classification} $|$ \textbf{TanH - 5 class}} \\
         \cmidrule(lr){4-7}
         & & & \multicolumn{2}{c}{\textit{1-layer}} & \multicolumn{2}{c}{\textit{2-layers}} \\
         \cmidrule(lr){4-5}\cmidrule(lr){6-7}
         & & & \textit{2D} & \textit{25D} & \textit{2D} & \textit{25D} \\
         \midrule
\multirow{4}{*}{Baseline} & - & Random & $19.745$\sstd{$0.269$} & $20.037$\sstd{$0.075$} & $20.214$\sstd{$0.043$} & $19.864$\sstd{$0.089$} \\
& - & Optimization & $92.919$\sstd{$0.011$} & $49.972$\sstd{$0.070$} & $88.156$\sstd{$0.014$} & $39.410$\sstd{$0.043$} \\
& - & Langevin & $88.635$\sstd{$0.374$} & $39.758$\sstd{$0.212$} & $84.050$\sstd{$0.138$} & $31.472$\sstd{$0.010$} \\
& - & HMC & $81.057$\sstd{$0.277$} & $35.378$\sstd{$0.130$} & $75.305$\sstd{$0.134$} & $29.730$\sstd{$0.670$} \\
\bottomrule
    \end{tabular}
    \caption{\textbf{Fixed-Dimensional}. Results for estimating the parameters of nonlinear classification models with TanH activation function and five classes, with the expected accuracy according to the posterior predictive as metric.}
    \vspace{-4mm}
    \label{tab:apdx_nlc_5cl}
\end{table*}
\subsection{Meta Learning}
Meta-learning~\citep{hospedales2022metareview} aims to equip models with the ability to quickly learn from different tasks or data sets to generalize to new tasks in resource-constrained domains. This attribute is precious in practical scenarios where obtaining large amounts of task-specific data is impractical or costly. A simple way of obtaining this is through nonparametric or similarity-based models like k-Nearest Neighbours, where no training is involved. Thus, new tasks can be solved quickly based on a few examples by computing a similarity metric with these examples~\citep{koch2015siamese,vinyals2016matching,sung2018learning}. Another way of achieving this is through optimization-based setups, which use a nested optimization procedure. An inner step learns individual tasks from a shared initialization, whereas the outer loop computes the gradient of the whole inner process and moves the initialization in a way that allows for better generalization. Here, by relying on only a few iterations in the inner loop, the outer loop has the incentive to move the initialization to a point from which solutions to multiple tasks are reachable~\citep{finn2017model}. Given the similarities between meta-learning and hierarchical Bayesian inference~\citep{grant2018recasting}, our approach can be considered as a kind of meta-learning framework; however, the line between meta-learning and Bayesian posterior inference is quite blurry as any amortized approach for the latter can be seen as a case of the former.

\subsection{Neural Processes}
A notable approach in meta-learning related to our research is neural processes (NP), which excel in learning scenarios with few examples. 
NPs~\citep{garnelo2018conditional,garnelo2018neural,kim2019attentive,pakman2020neural,gordon2019convolutional} can be seen as a more flexible and powerful extension of Gaussian processes that leverage a neural network-based encoder-decoder architecture for learning to model a distribution over functions that approximate a stochastic process.
However, while we are interested in approximating the posterior distribution over the parameters, NPs are used to approximate the posterior predictive distribution to make predictions based on observed data. Similar to our setup, NPs rely on amortized VI for obtaining the predictive posterior. Still, instead of working with a known probabilistic model, they train the probabilistic model primarily for prediction-based tasks through approaches analogous to variational expectation maximization. Thus, they cannot provide an explicit posterior over the parameters, but they are suitable for tasks where only predictive posteriors are essential, such as those in supervised learning. NPs, in their most basic form, accomplish this by training for the objective:
\begin{align}
    \label{eq:np}
    \arg\max_{\mtheta, \varphi} \sE_{\gD \sim \chi}\sE_{\vz \sim q_\varphi(\cdot | \gD)}\left[\log \frac{p_\mtheta(\gD, \vz)}{q_\varphi(\vz | \gD)}\right]
\end{align}
where $\vz \in \sR^p$ is an arbitrary latent variable often uninterpretable, and the parameters of the probabilistic model $\mtheta$ do not get a Bayesian treatment. In particular, NPs are more suited to modeling datasets of the form $\gD = \{\vx_i, \vy_i\}_{i=1}^n$, where all probabilities in \Eqref{eq:np} are conditioned on the input $\vx$'s, and only the predictive over $\vy$'s is modeled, and $p_\mtheta$ is modeled as a Neural Network.

These approaches can be seen as quite related to ICL, where the exchangeable architecture backbone is switched from DeepSets to Transformers. Similar to ICL, they do not provide control over the solution space as they aim to model either the posterior predictive or an arbitrary latent space. While this leads to good predictive performance on various tasks, they cannot be freely applied to problems that pose certain constraints on the underlying probabilistic model. In such cases, estimating the actual parameters is important to enforce constraints in the parameter space as well as for interpretability, which we already discussed in the ICL section.

\subsection{Simulation-Based Inference}
In the case of simulation-based inference~\citep{cranmer2020sbireview}, when the likelihood $p(\vx|\mtheta)$ is intractable, BayesFlow~\citep{radev2020bayesflow} and similar methods~\citep{lorch2022amortized} provide a solution framework to amortize Bayesian inference of parameters in complex models. Starting from the forward KL divergence between the true and approximate posteriors, the resulting objective is to optimize for parameters of the approximate posterior distribution that maximize the posterior probability of data-generating parameters $\mtheta$ given observed data $\gD$ for all $\mtheta$ and $\gD$. Density estimation of the approximate posterior can then be done using the change-of-variables formula and a conditional invertible neural network that parameterizes the approximate posterior distribution. 
\begin{equation}
    \argmin_\varphi \sK\sL[p(\mtheta|\gD) || q_\varphi(\mtheta|\gD)] = \argmin_{\varphi=\{ \nu,\psi \}} \sE_{(\mtheta, \gD) \sim p(\mtheta, \gD)} \left[ - \log p_\vz(f_\nu(\mtheta; h_\psi(\gD))) - \log \left|\det J_{f_\nu}\right| \right]
\end{equation}
Since their goal is to learn a global estimator for the probabilistic mapping from $\gD$ to data generating $\mtheta$, the information about the observed dataset is encoded in the output of a summary network $h_\psi$. It is used as conditional input to the normalizing flow $f_\nu$. Although the likelihood function does not need to be known, the method requires access to paired observations $(\vx, \mtheta)$ for training, which is sometimes unavailable. This approach is equivalent to the \textit{Forward KL} setup in our experiments when trained with DeepSets and Normalizing Flows. Current research has also leveraged score-based generative models for SBI which can condition on a dataset by learning a score model conditional only on single observations~\citep{geffner2023compositional}.

\subsection{Amortization in Gaussian Processes}
Gaussian Processes (GPs) define a class of probabilistic models that do enjoy tractable likelihood. However, inference in such systems is slow and sensitive to the choice of kernel function that defines the covariance matrix. Similar to meta learning and neural processes, current research also focuses on estimating the kernel function in GPs by leveraging permutation invariant architectures like transformers~\citep{liu2020task,simpson2021kernel,bitzer2023amortized}. Additionally, often these approaches amortize based on point estimates and are leveraged when considering GPs for regression problems, and it is not straightforward to extend them to classification or unsupervised learning. In contrast, our approach is more general and can work for all problems that define a differentiable likelihood function. Additionally, our approach also approximates the Bayesian posterior distribution over the parameters of interest, as opposed to point estimates.

\subsection{Mode Collapse in Variational Inference}
Reverse KL based methods have been widely known to suffer from mode collapse due to the nature of the optimization objective~\citep{bishop2006pattern}, which implies that even if the approximate distribution possesses the ability to represent multiple modes, optimization is often sub-optimal and the distribution ends up covering only a small handful of them. Improving normalizing flow based methods with repulsive terms or through the lens of natural gradient optimization procedure for a mixture approximate distribution~\citep{arenz2022unified,lin2020handling} is an important topic of research, and we believe it would be quite an important future work to experimentally validate if they help in learning multi-modality in amortized posterior inference problems that are studied in this work.
\section{Architectures respecting Exchangeability}
\label{appdx:exchangeability}
In this section, we highlight how DeepSets and Transformer models satisfy the dataset exchangeability criteria, which is essential in modeling the posterior distribution over the parameters of any probabilistic model relying on \textit{iid} data. 

\subsection{DeepSets}
DeepSets~\citep{Zaheer2017deepsets} operate on arbitrary sets $\gX = \{x_1, ... x_N\} \subset \mathbb{R}^d$ of fixed dimensionality $d$ by first mapping each individual element $\vx_i \in \gX$ to some high-dimensional space using a nonlinear transform, which is parameterized as a multi-layered neural network with parameters $\varphi_1$
\begin{align}
    \vz_i = f_{\varphi_1}(\vx_i)
\end{align}
After having obtained this high-dimensional embedding of each element of the set, it applies an aggregation function $a(\cdot)$, which is a permutation invariant function that maps a set of elements $\gZ = \{\vz_1, ..., \vz_N\} \in \mathbb{R}^z$ to an element $\vh \in \mathbb{R}^z$,
\begin{align}
    \vh = a(\gZ)
\end{align}
Thus, the outcome does not change under permutations of $\gZ$. Finally, another nonlinear transform, parameterized by a multi-layered neural network with parameters $\varphi_2$, is applied to the outcome $\vh$ to provide the final output.
\begin{align}
    \vo = g_{\varphi_2}(\vh)
\end{align}
For our experiments, we then use the vector $\vo$ to predict the parameters of a parametric family of distributions (e.g., Gaussian or Flows) using an additional nonlinear neural network. As an example, for the Gaussian case, we consider the distribution $\gN(\cdot | \mmu, \mSigma)$, where
\begin{align}
    \mmu:= \mmu_{\varphi_3}(\vo) \quad\text{and}\quad\mSigma := \mSigma_{\varphi_4}(\vo)
\end{align}
which makes $\mmu$ implicitly a function of the original input set $\gX$. To understand why the posterior distribution modeled in this fashion does not change when the inputs are permuted, let us assume that $\Pi$ is a permutation over the elements of $\gX$. If we look at one of the parameters of the posterior distribution, e.g., $\mmu$, we can see that
\begin{align}
    \mmu(\Pi \gX) &= \mmu_{\varphi_3}\left(g_{\varphi_2}\left(a\left(\{f_{\varphi_1}(\vx_{\Pi(i)})\}_{i=1}^N\right)\right)\right) \\
    &= \mmu_{\varphi_3}\left(g_{\varphi_2}\left(a\left(\{f_{\varphi_1}(\vx_i)\}_{i=1}^N\right)\right)\right) \\
    &= \mmu(\gX)
\end{align}
which simply follows from the fact that $a(\cdot)$ is a permutation invariant operation, e.g., sum or mean. We can also provide similar reasoning for the other parameters (e.g., $\mSigma$). This shows that DeepSets can be used to model the posterior distribution over parameters of interest as it respects the exchangeability criteria (\textit{iid} observations) assumptions in the data through its permutation invariant structure.

\subsection{Transformers}
Similarly, we can look at Transformers~\citep{vaswani2017attention} as candidates for respecting the exchangeability conditions in the data. In particular, we consider transformer systems without positional encodings and consider an additional [CLS] token, denoted by $\vc\in\mathbb{R}^d$, to drive the prediction. If we look at the application of a layer of transformer model, it can be broken down into two components.

\textbf{Multi-Head Attention}. Given a query vector obtained from $\vc$ and keys and values coming from our input set $\gX \subset \mathbb{R}^d$, we can model the update of the context $\vc$ as
\begin{align}
    \hat{\vc}(\gX) = \text{Softmax}\left(\vc^T \mW_Q \mW_K^T \mX^T\right) \mX \mW_V
\end{align}
where $\mW_Q \in \mathbb{R}^{d\times k}, \mW_K \in \mathbb{R}^{d\times k}, \mW_V \in \mathbb{R}^{d\times k}$ and $\mX \in \mathbb{R}^{N\times d}$ denotes a certain ordering of the elements in $\gX$. Further, $\hat{\vc}$ is the updated vector after attention, and Softmax is over the rows of $\mX$. Here, we see that if we were to apply a permutation to the elements in $\mX$, the outcome would remain the same. In particular
\begin{align}
    \hat{\vc}(\Pi \mX) &= \text{Softmax}\left(\vc^T \mW_Q \mW_K^T \mX^T \Pi^T\right) \Pi \mX \mW_V \\
    &= \text{Softmax}\left(\vc^T \mW_Q \mW_K^T \mX^T\right) \Pi^T\Pi \mX \mW_V \\
    &= \text{Softmax}\left(\vc^T \mW_Q \mW_K^T \mX^T\right) \mX \mW_V \\
    &= \hat{\vc}(\mX) 
\end{align}
which follows because Softmax is an equivariant function,  i.e., applying Softmax on a permutation of columns is equivalent to applying Softmax first and then permuting the columns correspondingly. Thus, we see that the update to the [CLS] token $\vc$ is permutation invariant. This output is then used independently as input to a multi-layered neural network with residual connections, and the entire process is repeated multiple times without weight sharing to simulate multiple layers. Since all the individual parts are permutation invariant w.r.t permutations on $\gX$, the entire setup ends up being permutation invariant. Obtaining the parameters of a parametric family of distribution for posterior estimation then follows the same recipe as DeepSets, with $\vo$ replaced by $\vc$.
\section{Probabilistic Models}
\label{appdx:probabilistic_models}
This section details the various candidate probabilistic models used in our experiments for amortized computation of Bayesian posteriors over the parameters. Here, we explain the parameters associated with the probabilistic model over which we want to estimate the posterior and the likelihood and prior that we use for experimentation.

\textbf{Mean of Gaussian (GM):} As a proof of concept, we consider the simple setup of estimating the posterior distribution over the mean of a Gaussian distribution $p(\mmu | \gD)$ given some observed data. In this case, prior and likelihood defining the probabilistic model $p(\vx, \mtheta)$ (with $\mtheta$ being the mean $\mmu$) are given by:
\begin{align}
    p(\mmu) &= \gN\left(\mmu | \mathbf{0}, \mathbf{I}\right)\\
    p(\vx | \mmu) &= \gN\left(\vx | \mmu, \mSigma\right) 
\end{align}
and $\mSigma$ is known beforehand and defined as a unit variance matrix. 

\begin{table*}[t]
    \centering
    \small
    \def\arraystretch{1.25}
    \setlength{\tabcolsep}{3pt}
\begin{tabular}{lcr cc cccc}
        \toprule
         &  &  & \multicolumn{6}{c}{\textit{$L_2$ Loss} ($\downarrow$)} \\
        \cmidrule(lr){4-9}
        \textbf{Objective} & $q_\varphi$ & \textbf{Model} & \multicolumn{2}{c}{\textbf{Gaussian}} & \multicolumn{4}{c}{\textbf{GMM}} \\
        \cmidrule(lr){4-5}\cmidrule(lr){6-9}
         & & & \textit{2D} & \textit{100D} & \textit{2D-2cl} & \textit{2D-5cl} & \textit{5D-2cl} & \textit{5D-5cl} \\
        \midrule
\multirow{4}{*}{Baseline} & - & Random & $6.297$\sstd{$0.017$} & $298.238$\sstd{$0.228$} & $2.078$\sstd{$0.134$} & $0.626$\sstd{$0.037$} & $4.659$\sstd{$0.034$} & $1.632$\sstd{$0.004$} \\
& - & Optimization & $2.020$\sstd{$0.000$} & $100.885$\sstd{$0.000$} & $0.175$\sstd{$0.002$} & $0.121$\sstd{$0.002$} & $0.427$\sstd{$0.000$} & $0.323$\sstd{$0.002$} \\
& - & Langevin & $2.036$\sstd{$0.004$} & $101.917$\sstd{$0.042$} & $0.178$\sstd{$0.002$} & $0.123$\sstd{$0.002$} & $0.440$\sstd{$0.002$} & $0.340$\sstd{$0.006$} \\
& - & HMC & $2.044$\sstd{$0.008$} & $102.015$\sstd{$0.009$} & $0.189$\sstd{$0.013$} & $0.132$\sstd{$0.004$} & $0.462$\sstd{$0.013$} & $0.423$\sstd{$0.009$} \\
\cmidrule{2-9}

\multirow{3}{*}{Fwd-KL} & \multirow{6}{*}{\rotatebox[origin=c]{90}{Gaussian}} & GRU &$2.300$\sstd{$0.105$} & $133.224$\sstd{$0.579$} & $1.119$\sstd{$0.150$} & $0.473$\sstd{$0.012$} & $2.360$\sstd{$0.017$} & $1.208$\sstd{$0.002$} \\
\bottomrule
    \end{tabular}
    \caption{\textbf{Variable-Dimensional}. Results for estimating the mean of a Gaussian (Gaussian) and means of a Gaussian mixture model (GMM) with the expected $L_2$ loss according to the posterior predictive as metric.}
    \vspace{-4mm}
    \label{tab:variable_apdx_gaussian}
\end{table*}
\begin{table*}[t]
    \centering
    \small
    \def\arraystretch{1.25}
    \setlength{\tabcolsep}{3pt}
    \begin{tabular}{lcr cc cccc}
        \toprule
         &  &  & \multicolumn{2}{c}{\textit{$L_2$ Loss} ($\downarrow$)} & \multicolumn{2}{c}{\textit{Accuracy} ($\uparrow$)} \\
         \cmidrule(lr){4-5}\cmidrule(lr){6-9}
         \textbf{Objective} & $q_\varphi$ & \textbf{Model} & \multicolumn{2}{c}{\textbf{Linear Regression}} & \multicolumn{4}{c}{\textbf{Linear Classification}} \\
         \cmidrule(lr){4-5}\cmidrule(lr){6-9}
         & & & \textit{2D} & \textit{100D} & \textit{2D-2cl} & \textit{2D-5cl} & \textit{100D-2cl} & \textit{100D-5cl} \\
         \midrule
\multirow{4}{*}{Baseline} & - & Random & $4.272$\sstd{$0.068$} & $200.836$\sstd{$0.609$} & $50.125$\sstd{$0.264$} & $20.078$\sstd{$0.065$} & $50.005$\sstd{$0.061$} & $20.033$\sstd{$0.082$} \\
& - & Optimization & $0.258$\sstd{$0.000$} & $20.127$\sstd{$0.003$} & $97.301$\sstd{$0.000$} & $91.752$\sstd{$0.000$} & $71.231$\sstd{$0.010$} & $42.345$\sstd{$0.001$} \\
& - & Langevin & $0.263$\sstd{$0.002$} & $21.781$\sstd{$0.953$} & $95.441$\sstd{$0.209$} & $86.445$\sstd{$0.496$} & $65.469$\sstd{$0.513$} & $32.668$\sstd{$0.145$} \\
& - & HMC & $0.263$\sstd{$0.000$} & $17.774$\sstd{$0.120$} & $92.961$\sstd{$0.228$} & $78.793$\sstd{$0.314$} & $62.602$\sstd{$0.171$} & $30.055$\sstd{$0.506$} \\
\cmidrule{2-9}

\multirow{3}{*}{Fwd-KL} & \multirow{6}{*}{\rotatebox[origin=c]{90}{Gaussian}} & GRU & \highlight{$0.271$\sstd{$0.004$}} & $139.396$\sstd{$1.012$} & $79.467$\sstd{$0.711$} & $65.124$\sstd{$0.861$} & $57.872$\sstd{$0.157$} & $22.677$\sstd{$0.081$} \\
\bottomrule
    \end{tabular}
    \caption{\textbf{Variable-Dimensional}. Results for estimating the parameters of linear regression (LR) and classification (LC) models with the expected $L_2$ loss and accuracy according to the posterior predictive as metrics.}
    \vspace{-4mm}
    \label{tab:}
\end{table*}
\begin{table*}[t]
    \centering
    \small
    \def\arraystretch{1.25}
    \setlength{\tabcolsep}{5pt}
    \begin{tabular}{lcr cccc}
        \toprule
         &  &  & \multicolumn{4}{c}{\textit{$L_2$ Loss} ($\downarrow$)} \\
         \cmidrule(lr){4-7}
         & & & \multicolumn{2}{c}{\textit{1-layer}} & \multicolumn{2}{c}{\textit{2-layers}} \\
         \cmidrule(lr){4-5}\cmidrule(lr){6-7}
         \textbf{Objective} & $q_\varphi$ & \textbf{Model} & \textit{1D} & \textit{50D} & \textit{1D} & \textit{50D} \\
         \midrule
\multirow{4}{*}{Baseline} & - & Random & $73.006$\sstd{$0.171$} & $1704.335$\sstd{$9.330$} & $998.890$\sstd{$16.317$} & $27799.887$\sstd{$165.653$} \\
& - & Optimization & $0.359$\sstd{$0.002$} & $309.162$\sstd{$0.204$} & $3.078$\sstd{$0.057$} & $4894.141$\sstd{$4.290$} \\
& - & Langevin & $0.308$\sstd{$0.002$} & \textsc{N/A} & \textsc{N/A} & \textsc{N/A} \\
& - & HMC & $0.381$\sstd{$0.005$} & $303.857$\sstd{$2.487$} & $7.999$\sstd{$0.595$} & $12905.846$\sstd{$9.903$} \\
\cmidrule{2-7}

\multirow{3}{*}{Fwd-KL} & \multirow{6}{*}{\rotatebox[origin=c]{90}{Gaussian}} & GRU &$51.448$\sstd{$0.029$} & $1346.462$\sstd{$6.833$} & $754.388$\sstd{$32.650$} & $21121.735$\sstd{$112.458$} \\
\bottomrule
    \end{tabular}
    \caption{\textbf{Variable-Dimensional}. Results for estimating the parameters of nonlinear regression models with ReLU activation function, with the expected $L_2$ loss according to the posterior predictive as metric.}
    \vspace{-4mm}
    \label{tab:}
\end{table*}
\begin{table*}[t]
    \centering
    \small
    \def\arraystretch{1.25}
    \setlength{\tabcolsep}{5pt}
    \begin{tabular}{lcr cccc}
        \toprule
         &  &  & \multicolumn{4}{c}{\textit{$L_2$ Loss} ($\downarrow$)} \\
         \cmidrule(lr){4-7}
         & & & \multicolumn{2}{c}{\textit{1-layer}} & \multicolumn{2}{c}{\textit{2-layers}} \\
         \cmidrule(lr){4-5}\cmidrule(lr){6-7}
         \textbf{Objective} & $q_\varphi$ & \textbf{Model} & \textit{1D} & \textit{50D} & \textit{1D} & \textit{50D} \\
         \midrule
\multirow{4}{*}{Baseline} & - & Random & $33.972$\sstd{$0.273$} & $56.891$\sstd{$0.213$} & $50.843$\sstd{$0.270$} & $55.383$\sstd{$0.124$} \\
& - & Optimization & $0.330$\sstd{$0.000$} & $23.257$\sstd{$0.029$} & $0.672$\sstd{$0.007$} & $34.900$\sstd{$0.120$} \\
& - & Langevin & $0.296$\sstd{$0.003$} & $26.292$\sstd{$0.266$} & $0.356$\sstd{$0.006$} & $36.598$\sstd{$0.354$} \\
& - & HMC & $0.404$\sstd{$0.005$} & $19.937$\sstd{$0.259$} & $0.654$\sstd{$0.014$} & $32.884$\sstd{$0.448$} \\
\cmidrule{2-7}

\multirow{3}{*}{Fwd-KL} & \multirow{6}{*}{\rotatebox[origin=c]{90}{Gaussian}} & GRU &$34.235$\sstd{$0.240$} & $56.947$\sstd{$0.339$} & $50.493$\sstd{$0.676$} & $55.633$\sstd{$0.396$} \\
\bottomrule
    \end{tabular}
    \caption{\textbf{Variable-Dimensional}. Results for estimating the parameters of nonlinear regression models with TanH activation function, with the expected $L_2$ loss according to the posterior predictive as metric.}
    \vspace{-4mm}
    \label{tab:}
\end{table*}
\begin{table*}[t]
    \centering
    \small
    \def\arraystretch{1.25}
    \setlength{\tabcolsep}{5pt}
    \begin{tabular}{lcr cc cc}
        \toprule
         &  &  & \multicolumn{4}{c}{\textit{Accuracy} ($\uparrow$)} \\
         \cmidrule(lr){4-7}
         & & & \multicolumn{2}{c}{\textit{1-layer}} & \multicolumn{2}{c}{\textit{2-layers}} \\
         \cmidrule(lr){4-5}\cmidrule(lr){6-7}
         \textbf{Objective} & $q_\varphi$ & \textbf{Model} & \textit{2D} & \textit{50D} & \textit{2D} & \textit{50D} \\
         \midrule
\multirow{4}{*}{Baseline} & - & Random & $49.951$\sstd{$0.287$} & $49.904$\sstd{$0.281$} & $50.040$\sstd{$0.467$} & $50.044$\sstd{$0.239$} \\
& - & Optimization & $96.762$\sstd{$0.034$} & $76.139$\sstd{$0.033$} & $96.810$\sstd{$0.009$} & $78.225$\sstd{$0.129$} \\
& - & Langevin & $96.077$\sstd{$0.027$} & $70.142$\sstd{$0.229$} & $96.564$\sstd{$0.113$} & $71.328$\sstd{$0.410$} \\
& - & HMC & $91.734$\sstd{$0.152$} & $67.986$\sstd{$0.372$} & $91.336$\sstd{$0.591$} & $71.825$\sstd{$0.507$} \\
\cmidrule{2-7}

\multirow{3}{*}{Fwd-KL} & \multirow{6}{*}{\rotatebox[origin=c]{90}{Gaussian}} & GRU &$59.551$\sstd{$0.199$} & $58.637$\sstd{$0.250$} & $60.247$\sstd{$0.645$} & $58.862$\sstd{$0.065$} \\
\bottomrule
    \end{tabular}
    \caption{\textbf{Variable-Dimensional}. Results for estimating the parameters of nonlinear classification models with ReLU activation function and two classes, with the expected accuracy according to the posterior predictive as metric.}
    \vspace{-4mm}
    \label{tab:}
\end{table*}
\begin{table*}[t]
    \centering
    \small
    \def\arraystretch{1.25}
    \setlength{\tabcolsep}{5pt}
    \begin{tabular}{lcr cc cc}
        \toprule
         &  &  & \multicolumn{4}{c}{\textit{Accuracy} ($\uparrow$)} \\
         \cmidrule(lr){4-7}
         & & & \multicolumn{2}{c}{\textit{1-layer}} & \multicolumn{2}{c}{\textit{2-layers}} \\
         \cmidrule(lr){4-5}\cmidrule(lr){6-7}
         \textbf{Objective} & $q_\varphi$ & \textbf{Model} & \textit{2D} & \textit{50D} & \textit{2D} & \textit{50D} \\
         \midrule
\multirow{4}{*}{Baseline} & - & Random & $19.847$\sstd{$0.352$} & $20.052$\sstd{$0.066$} & $19.874$\sstd{$0.222$} & $20.028$\sstd{$0.106$} \\
& - & Optimization & $94.607$\sstd{$0.011$} & $56.091$\sstd{$0.158$} & $93.873$\sstd{$0.028$} & $60.253$\sstd{$0.053$} \\
& - & Langevin & $90.815$\sstd{$0.341$} & $46.072$\sstd{$0.225$} & $91.849$\sstd{$0.088$} & $49.808$\sstd{$0.337$} \\
& - & HMC & $81.145$\sstd{$0.303$} & $44.561$\sstd{$0.309$} & $79.559$\sstd{$0.483$} & $50.967$\sstd{$0.436$} \\
\cmidrule{2-7}

\multirow{3}{*}{Fwd-KL} & \multirow{6}{*}{\rotatebox[origin=c]{90}{Gaussian}} & GRU &$30.960$\sstd{$0.638$} & $32.164$\sstd{$0.151$} & $31.017$\sstd{$0.397$} & $32.224$\sstd{$0.108$} \\
\bottomrule
    \end{tabular}
    \caption{\textbf{Fixed-Dimensional}. Results for estimating the parameters of nonlinear classification models with ReLU activation function and five classes, with the expected accuracy according to the posterior predictive as metric.}
    \vspace{-4mm}
    \label{tab:}
\end{table*}
\begin{table*}[t]
    \centering
    \small
    \def\arraystretch{1.25}
    \setlength{\tabcolsep}{5pt}
    \begin{tabular}{lcr cc cc}
        \toprule
         &  &  & \multicolumn{4}{c}{\textit{Accuracy} ($\uparrow$)} \\
         \cmidrule(lr){4-7}
         & & & \multicolumn{2}{c}{\textit{1-layer}} & \multicolumn{2}{c}{\textit{2-layers}} \\
         \cmidrule(lr){4-5}\cmidrule(lr){6-7}
         \textbf{Objective} & $q_\varphi$ & \textbf{Model} & \textit{2D} & \textit{50D} & \textit{2D} & \textit{50D} \\
         \midrule
\multirow{4}{*}{Baseline} & - & Random & $50.278$\sstd{$0.337$} & $50.028$\sstd{$0.064$} & $50.188$\sstd{$0.479$} & $49.982$\sstd{$0.084$} \\
& - & Optimization & $96.943$\sstd{$0.012$} & $68.086$\sstd{$0.016$} & $94.444$\sstd{$0.024$} & $63.950$\sstd{$0.022$} \\
& - & Langevin & $95.143$\sstd{$0.094$} & $61.694$\sstd{$0.340$} & $92.719$\sstd{$0.016$} & $57.447$\sstd{$0.437$} \\
& - & HMC & $92.489$\sstd{$0.338$} & $59.963$\sstd{$0.202$} & $87.548$\sstd{$0.094$} & $56.319$\sstd{$0.882$} \\
\cmidrule{2-7}

\multirow{3}{*}{Fwd-KL} & \multirow{6}{*}{\rotatebox[origin=c]{90}{Gaussian}} & GRU &$50.274$\sstd{$0.337$} & $50.023$\sstd{$0.060$} & $50.187$\sstd{$0.477$} & $49.992$\sstd{$0.072$} \\
\bottomrule
    \end{tabular}
    \caption{\textbf{Variable-Dimensional}. Results for estimating the parameters of nonlinear classification models with TanH activation function and two classes, with the expected accuracy according to the posterior predictive as metric.}
    \vspace{-4mm}
    \label{tab:}
\end{table*}
\begin{table*}[t]
    \centering
    \small
    \def\arraystretch{1.25}
    \setlength{\tabcolsep}{5pt}
    \begin{tabular}{lcr cc cc}
        \toprule
         &  &  & \multicolumn{4}{c}{\textit{Accuracy} ($\uparrow$)} \\
         \cmidrule(lr){4-7}
         & & & \multicolumn{2}{c}{\textit{1-layer}} & \multicolumn{2}{c}{\textit{2-layers}} \\
         \cmidrule(lr){4-5}\cmidrule(lr){6-7}
         \textbf{Objective} & $q_\varphi$ & \textbf{Model} & \textit{2D} & \textit{50D} & \textit{2D} & \textit{50D} \\
         \midrule
\multirow{4}{*}{Baseline} & - & Random & $20.041$\sstd{$0.136$} & $20.002$\sstd{$0.079$} & $19.914$\sstd{$0.111$} & $19.954$\sstd{$0.005$} \\
& - & Optimization & $92.059$\sstd{$0.012$} & $40.722$\sstd{$0.027$} & $88.848$\sstd{$0.005$} & $34.136$\sstd{$0.041$} \\
& - & Langevin & $88.357$\sstd{$0.309$} & $30.941$\sstd{$0.097$} & $83.788$\sstd{$0.170$} & $26.538$\sstd{$0.206$} \\
& - & HMC & $79.161$\sstd{$0.292$} & $27.508$\sstd{$0.379$} & $74.987$\sstd{$0.130$} & $25.377$\sstd{$0.246$} \\
\cmidrule{2-7}

\multirow{3}{*}{Fwd-KL} & \multirow{6}{*}{\rotatebox[origin=c]{90}{Gaussian}} & GRU &$20.264$\sstd{$0.139$} & $20.158$\sstd{$0.056$} & $20.215$\sstd{$0.138$} & $20.093$\sstd{$0.028$} \\
\bottomrule
    \end{tabular}
    \caption{\textbf{Variable-Dimensional}. Results for estimating the parameters of nonlinear classification models with TanH activation function and five classes, with the expected accuracy according to the posterior predictive as metric.}
    \vspace{-4mm}
    \label{tab:variable_apdx_nlc_5cl}
\end{table*}
\textbf{Linear Regression (LR):} We then look at the problem of estimating the posterior over the weight vector for Bayesian linear regression given a dataset $p(\vw, b | \gD)$, where the underlying model $p(\gD, \mtheta)$ is given by:
\begin{align}
    p(\vw) &= \gN(\vw | \mathbf{0}, \mathbf{I})\\
    p(b) &= \gN(b | 0, 1)\\
    p(y | \vx, \vw, b) &= \gN\left(y | \vw^T\vx + b, \sigma^2\right) \, ,
\end{align}
and with $\sigma^2 = 0.25$ known beforehand. Inputs $\vx$ are generated from $p(\vx) = \gN(\mathbf{0}, I)$.

\textbf{Linear Classification (LC):}
We now consider a setting where the true posterior cannot be obtained analytically as the likelihood and prior are not conjugate. In this case, we consider the underlying probabilistic model by:
\begin{align}
    p(\mW) &= \gN\left(\mW | \mathbf{0}, \mathbf{I}\right)\\
    p(y | \vx, \mW) &= \mathrm{Categorical}\left(y  \;\vline\; \frac{1}{\tau}\;\mW\vx\right)\, ,
\end{align}
where $\tau$ is the known temperature term which is kept as $0.1$ to ensure peaky distributions, and $\vx$ is being generated from $p(\vx) = \gN(\mathbf{0}, I)$.

\textbf{Nonlinear Regression (NLR):}
Next, we tackle the more complex problem where the posterior distribution is multi-modal and obtaining multiple modes or even a single good one is challenging. For this, we consider the model as a Bayesian Neural Network (BNN) for regression with fixed hyper-parameters like the number of layers, dimensionality of the hidden layer, etc. Let the BNN denote the function $f_\mtheta$ where $\mtheta$ are the network parameters such that the estimation problem is to approximate $p(\mtheta | \gD)$. Then, for regression, we specify the probabilistic model using:
\begin{align}
    p(\mtheta) &= \gN\left(\mtheta | \mathbf{0}, \mathbf{I}\right)\\
    p(y | \vx, \mtheta) &= \gN\left(y | f_\mtheta(\vx), \sigma^2\right) \, ,
\end{align}
where $\sigma^2 = 0.25$ is a known quantity and $\vx$ being generated from $p(\vx) = \gN(\mathbf{0}, I)$.
 
\textbf{Nonlinear Classification (NLC):}
Like in Nonlinear Regression, we consider BNNs with fixed hyper-parameters for classification problems with the same estimation task of approximating $p(\mtheta | \gD)$. In this formulation, we consider the probabilistic model as:
\begin{align}
    p(\mtheta) &= \gN\left(\mtheta | \mathbf{0}, \mathbf{I}\right)\\
    p(y | \vx, \mtheta) &= \mathrm{Categorical}\left(y \;\vline\; \frac{1}{\tau}\;f_\mtheta(\vx)\right)
\end{align}
where $\tau$ is the known temperature term which is kept as $0.1$ to ensure peaky distributions, and $\vx$ is being generated from $p(\vx) = \gN(\mathbf{0}, I)$.

\textbf{Gaussian Mixture Model (GMM):}
While we have mostly looked at predictive problems, where the task is to model some predictive variable $y$ conditioned on some input $\vx$, we now look at a well-known probabilistic model for unsupervised learning, Gaussian Mixture Model (GMM), primarily used to cluster data. Consider a $K$-cluster GMM with:
\begin{align}
    p(\mmu_k) &= \gN\left(\mmu_k | \mathbf{0}, \mathbf{I}\right)\\
    p(\vx | \mmu_{1:K}) &= \sum_{k=1}^K \pi_k \gN\left(\vx | \mmu_k, \mSigma_k\right) \, .
\end{align}
 We assume $\mSigma_k$ and $\pi_k$ to be known and set $\mSigma_k$ to be an identity matrix and the mixing coefficients to be equal, $\pi_k = 1/K$, for all clusters $k$ in our experiments. 
\section{Metrics}
\label{appdx:metrics}
In this section, we provide details about the metrics considered for the different tasks. We generally look at two main metrics for benchmarking performance: $L_2$ loss and Accuracy. For estimating the mean of a Gaussian distribution, the $L_2$ loss is defined as
\begin{align}
    GM_{L_2} &= \mathbb{E}_{\gD \sim \chi}\mathbb{E}_{\mmu \sim q_\varphi(\cdot | \gD)}\left[\sum_{i=1}^{N_\gD} (\vx_i - \mmu)^2\right]
\end{align}
where $\gD = \{\vx_i\}_{i=1}^{N_\gD}$. Intuitively, this captures the quality of the estimation of the mean parameter by measuring how far the observations are from it. Lower value implies better estimation of the mean parameter. Similarly, for estimating the means of a Gaussian Mixture Model, we rely on a similar metric but we also find the cluster closest to the observation, which can be defined as
\begin{align}
    GMM_{L_2} &= \mathbb{E}_{\gD \sim \chi}\mathbb{E}_{\mmu_k \sim q_\varphi(\cdot | \gD)}\left[\sum_{i=1}^{N_\gD} (\vx_i - \mmu_{\text{Match}\left(\vx_i, \{\mmu_1, ... \mmu_K\}\right)})^2\right] \\
    \text{Match}(\vx, \{\mmu_1, ..., \mmu_K\} &= \arg\min_k (\vx - \mmu_k)^2
\end{align}
which intuitively captures the distance of observations from the cluster closest to them. Next, we define the metric for evaluating (non-)linear regression models as
\begin{align}
    (N-)LR_{L_2} &= \mathbb{E}_{\gD \sim \chi}\mathbb{E}_{\mtheta \sim q_\varphi(\cdot | \gD)}\left[\sum_{i=1}^{N_\gD} (y_i - \text{Mode}\left[p(y_i | \vx_i, \mtheta)\right])^2\right]
\end{align}
Finally, for the (non-)linear classification setups, we define the accuracy metric as
\begin{align}
    (N-)LC_{Accuracy} &= \mathbb{E}_{\gD \sim \chi}\mathbb{E}_{\mtheta \sim q_\varphi(\cdot | \gD)}\left[\frac{100}{N_\gD} \times \sum_{i=1}^{N_\gD} \delta(y_i, \text{Mode}\left[p(y_i | \vx_i, \mtheta)\right])\right]
\end{align}
where $\delta(a, b) = 1$ if and only if $a = b$. Thus this metric captures the accuracy of the posterior predictive distribution. Another metric that we use to test the quality of the posterior is the symmetric KL divergence, defined as
\begin{align}
    \text{Symmetric }\mathbb{KL}(p(\mtheta || \gD), q_\varphi(\mtheta | \gD)) &= \frac{1}{2}\mathbb{KL}(p(\mtheta || \gD) || q_\varphi(\mtheta | \gD)) + \frac{1}{2}\mathbb{KL}(q_\varphi(\mtheta | \gD) || p(\mtheta || \gD))
\end{align}
\section{Architecture Details}
\label{appdx:architecture}
In this section, we outline the two candidate architectures that we consider for the backbone of our amortized variational inference model. We discuss the specifics of the architectures and the hyperparameters used for our experiments.

\subsection{Transformer}
\label{subsec:transformer}
We use a transformer model~\citep{vaswani2017attention} as a permutation invariant architecture by removing positional encodings from the setup and using multiple layers of the encoder model. We append the set of observations with a [CLS] token before passing it to the model and use its output embedding to predict the parameters of the variational distribution. Since no positional encodings or causal masking is used in the whole setup, the final embedding of the [CLS] token becomes invariant to permutations in the set of observations, thereby leading to permutation invariance in the parameters of $q_\varphi$.

We use $4$ encoder layers with a $256$ dimensional attention block and $1024$ feed-forward dimensions, with $4$ heads in each attention block for our Transformer models to make the number of parameters comparative to the one of the DeepSets model.

\subsection{DeepSets}
\label{subsec:deepsets}
Another framework that can process set-based input is Deep Sets~\citep{Zaheer2017deepsets}. In our experiments, we used an embedding network that encodes the input into representation space, a mean aggregation operation, which ensures that the representation learned is invariant concerning the set ordering, and a regression network. The latter's output is either used to directly parameterize a diagonal Gaussian or as conditional input to a normalizing flow, representing a summary statistics of the set input.

For DeepSets, we use $4$ layers each in the embedding network and the regression network, with a mean aggregation function, ReLU activation functions, and $627$ hidden dimensions to make the number of parameters comparable to those in the Transformer model.

\subsection{RNN}
For the recurrent neural network setup, we use the Gated Recurrent Unit (GRU). Similar to the above setups, we use a $4$-layered GRU model with $256$ hidden dimensions. While such an architecture is not permutation invariant, by training on tasks that require such invariance could encourage learning of solution structure that respects this invariance.

\subsection{Normalizing Flows}
\label{subsec:flows}
Assuming a Gaussian posterior distribution as the approximate often leads to poor results as the true posterior distribution can be far from the Gaussian shape. To allow for more flexible posterior distributions, we use normalizing flows~\citep{kingma2018glow,kobyzev2020normalizing,papamakarios2021normalizing,rezende2015variational} for approximating $q_\varphi(\mtheta | \gD)$ conditioned on the output of the summary network $h_\psi$. Specifically, let $g_\nu: \vz \mapsto \mtheta$ be a diffeomorphism parameterized by a conditional invertible neural network (cINN) with network parameters $\nu$ such that $\mtheta = g_\nu(\vz; h_\psi(\gD))$. With the change-of-variables formula it follows that $p(\mtheta)=p(\vz)\left|\det \frac{\partial}{\partial\vz}g_\nu(\vz; h_\psi(\gD))\right|^{-1} = p(\vz)|\det J_\nu(\vz; h_\psi(\gD))|^{-1}$, where $J_\nu$ is the Jacobian matrix of $g_\nu$. Further, integration by substitution gives us $d\mtheta = |\det J_\nu(\vz; h_\psi(\gD)| d\vz$ to rewrite the objective from eq. \ref{eq:arkl} as:
\begin{align}
    &\argmin_\varphi \sE_{\gD \sim \chi} \sK\sL[q_\varphi(\mtheta|\gD) || p(\mtheta|\gD)]\\
    &= \argmin_\varphi \sE_{\gD \sim \chi} \sE_{\mtheta \sim q_\varphi(\mtheta|\gD)} \left[ \log q_\varphi(\mtheta|\gD) - \log p(\mtheta, \gD) \right]\\
    &= \argmin_{\varphi=\{\psi, \nu\}} \sE_{\gD \sim \chi} \sE_{\vz \sim p(\vz)} \left[ \log \frac{q_\nu (\vz|h_\psi(\gD))}{\left| \det J_\nu(\vz; h_\psi(\gD)) \right|} - \log p(g_\nu(\vz; h_\psi(\gD)), \gD) \right]
\end{align}
As shown in BayesFlow \citep{radev2020bayesflow}, the normalizing flow $g_\nu$ and the summary network $h_\psi$ can be trained simultaneously. The $\mathrm{AllInOneBlock}$ coupling block architecture of the FrEIA Python package \citep{Ardizzone2018freia}, which is very similar to the RNVP style coupling block \citep{Dinh2017rnvp}, is used as the basis for the cINN. $\mathrm{AllInOneBlock}$ combines the most common architectural components, such as ActNorm, permutation, and affine coupling operations.

For our experiments, $6$ coupling blocks define the normalizing flow network, each with a $1$ hidden-layered non-linear feed-forward subnetwork with ReLU non-linearity and $128$ hidden dimensions.
\section{Experimental Details}
\label{appdx:experiment}
Unless specified, we obtain a stream of datasets for all our experiments by simply sampling from the assumed probabilistic model, where the number of observations $n$ is sampled uniformly in the range $[64, 128]$. For efficient mini-batching over datasets with different cardinalities, we sample datasets with maximum cardinality $(128)$ and implement different cardinalities by masking out different numbers of observations for different datasets whenever required. 

To evaluate both our proposed approach and the baselines, we compute an average of the predictive performances across $25$ different posterior samples for each of the $100$ fixed test datasets for all our experiments. 
That means for our proposed approach, we sample $25$ different parameter vectors from the approximate posterior that we obtain. For MCMC, we rely on $25$ MCMC samples, and for optimization, we train $25$ different parameter vectors where the randomness comes from initialization. 
For the optimization baseline, we perform a quick hyperparameter search over the space $\{0.01, 003, 0.001, 0.0003, 0.0001, 0.00003\}$ to pick the best learning rate that works for all of the test datasets and then use it to train for $1000$ iterations using the Adam optimizer~\citep{kingma2014adam}. For the MCMC baseline, we use the open-sourced implementation of Langevin-based MCMC sampling\footnote{\href{https://github.com/alisiahkoohi/Langevin-dynamics}{https://github.com/alisiahkoohi/Langevin-dynamics}} where we leave a chunk of the starting samples as burn-in and then start accepting samples after a regular interval (to not make them correlated). The details about the burn-in time and the regular interval for acceptance are provided in the corresponding experiments' sections below.

For our proposed approach of amortized inference, we do not consider explicit hyperparameter optimization and simply use a learning rate of $1\mathrm{e}\text{-}4$ with the Adam optimizer. For all experiments, we used linear scaling of the KL term in the training objectives as described in~\citep{higgins2017betavae}, which we refer to as warmup. Furthermore, training details for each experiment can be found below. 

\subsection{Fixed-Dim}
\label{appdx:details_fixed_dim}
In this section, we provide the experimental details relevant to reproducing the results of Section~\ref{sec:experiments}. All the models are trained with streaming data from the underlying probabilistic model, such that every iteration of training sees a new set of datasets. Training is done with a batch size of $128$, representing the number of datasets seen during one optimization step. Evaluations are done with $25$ samples and we ensure that the test datasets used for each probabilistic model are the same across all the compared methods, i.e., baselines, forward KL, and reverse KL. We train the amortized inference model and the forward KL baselines for the following different probabilistic models:

\textbf{Mean of Gaussian (GM):} We train the amortization models over $20,000$ iterations for both the $2$-dimensional as well as the $100$-dimensional setup. We use a linear warmup with $5000$ iterations over which the weight of the KL term in our proposed approach scales linearly from $0$ to $1$. We use an identity covariance matrix for the data-generating process, but it can be easily extended to the case of correlated or diagonal covariance-based Gaussian distributions.

\textbf{Gaussian Mixture Model (GMM):} We train the mixture model setup for $200,000$ iterations with $50,000$ iterations of warmup. We mainly experiment with $2$-dimensional and $5$-dimensional mixture models, with $2$ and $5$ mixture components for each setup. While we do use an identity covariance matrix for the data-generating process, again, it can be easily extended to other cases.

\textbf{Linear Regression (LR):} The amortization models for this setup are trained for $50,000$ iterations with $12,500$ iterations of warmup. The feature dimensions considered for this task are $1$ and $100$ dimensions, and the predictive variance $\sigma^2$ is assumed to be known and set as $0.25$.

\textbf{Nonlinear Regression (NLR):} We train the setup for $100,000$ iterations with $25,000$ iterations consisting of warmup. The feature dimensionalities considered are $1$-dimensional and $25$-dimensional, and training is done with a known predictive variance similar to the LR setup. For the probabilistic model, we consider both a $1$-layered and a $2$-layered multi-layer perceptron (MLP) network with 32 hidden units in each, and either a \textsc{relu} or \textsc{tanh} activation function.

\textbf{Linear Classification (LC):} We experiment with $2$-dimensional and $100$-dimensional setups with training done for $50,000$ iterations, out of which $12,500$ are used for warmup. Further, we train for both binary classification as well as a $5$-class classification setup.

\textbf{Nonlinear Classification (NLC):} We experiment with $2$-dimensional and $25$-dimensional setups with training done for $100,000$ iterations, out of which $2,5000$ are used for warmup. Further, we train for both binary classification as well as a $5$-class classification setup. For the probabilistic model, we consider both a $1$-layered and a $2$-layered multi-layer perceptron (MLP) network with 32 hidden units in each, and either a \textsc{relu} or \textsc{tanh} activation function.

\begin{table*}[t]
    \centering
    \footnotesize	    
    \def\arraystretch{1.25}
    \setlength{\tabcolsep}{5pt}
    \begin{tabular}{lcr ccc cccc}
        \cmidrule[\heavyrulewidth]{1-9}
         &  &  & \multicolumn{6}{c}{\textit{$L_2$ Loss} ($\downarrow$)} \\
        \cmidrule(lr){4-9}
        \textbf{Objective} & $q_\varphi$ & \textbf{Model} & \multicolumn{3}{c}{\textbf{Linear Model $|$ MLP-TanH Data}} & \multicolumn{3}{c}{\textbf{MLP-TanH Model $|$ Linear Data}} & $\leftarrow\chi_{real}$ \\
        \cmidrule(lr){4-6}\cmidrule(lr){7-9}
        & & & \textit{LR} & \textit{NLR} & \textit{GP} & \textit{LR} & \textit{NLR} & \textit{GP} & $\leftarrow\chi_{sim}$ \\
        \cmidrule{1-9}
\multirow{4}{*}{Baseline} & - & Random & - & $17.761$\sstd{$0.074$}  & -  & $17.847$\sstd{$0.355$} & -  & -  \\
& - & Optimization & - & $1.213$\sstd{$0.000$} & -  & $0.360$\sstd{$0.001$} & -  & -  \\
& - & Langevin & - & $1.218$\sstd{$0.002$} & -  & $0.288$\sstd{$0.001$} & -  & -  \\
& - & HMC & - & $1.216$\sstd{$0.002$} & -  & $0.275$\sstd{$0.001$} & -  & -  \\
\cmidrule{2-9}
\multirow{3}{*}{Fwd-KL} & \multirow{6}{*}{\rotatebox[origin=c]{90}{Gaussian}} & GRU &$2.415$\sstd{$0.269$} & -  & -  & -  & $15.632$\sstd{$0.283$} & -  \\
& & DeepSets &$1.402$\sstd{$0.017$} & -  & -  & -  & $16.046$\sstd{$0.393$} & -  \\
& & Transformer &$2.216$\sstd{$0.097$} & -  & -  & -  & $15.454$\sstd{$0.246$} & -  \\
\cmidrule{3-9}
\multirow{3}{*}{Rev-KL}& & GRU &$1.766$\sstd{$0.044$} & $1.216$\sstd{$0.001$} & $4.566$\sstd{$0.199$} & $0.375$\sstd{$0.001$} & $0.386$\sstd{$0.002$} & $0.524$\sstd{$0.019$} \\
& & DeepSets &$1.237$\sstd{$0.006$} & $1.216$\sstd{$0.001$} & $3963.694$\sstd{$5602.411$} & $0.365$\sstd{$0.000$} & $0.377$\sstd{$0.003$} & $0.385$\sstd{$0.011$} \\
& & Transformer &$1.892$\sstd{$0.113$} & $1.226$\sstd{$0.001$} & $4.313$\sstd{$0.707$} & $0.367$\sstd{$0.006$} & $0.382$\sstd{$0.003$} & $0.458$\sstd{$0.048$} \\
\cmidrule{2-9}
\multirow{3}{*}{Fwd-KL} & \multirow{6}{*}{\rotatebox[origin=c]{90}{Flow}} & GRU &$2.180$\sstd{$0.024$} & -  & -  & -  & $9.800$\sstd{$0.473$} & -  \\
& & DeepSets &$1.713$\sstd{$0.244$} & -  & -  & -  & $15.253$\sstd{$0.403$} & -  \\
& & Transformer &$1.632$\sstd{$0.070$} & -  & -  & -  & $7.949$\sstd{$0.419$} & -  \\
\cmidrule{3-9}
\multirow{3}{*}{Rev-KL} & & GRU &$1.830$\sstd{$0.081$} & \highlight{$1.214$\sstd{$0.001$}} & $5.690$\sstd{$0.196$} & $0.346$\sstd{$0.004$} & $0.349$\sstd{$0.001$} & $0.520$\sstd{$0.015$} \\
& & DeepSets &$1.282$\sstd{$0.036$} & $1.218$\sstd{$0.001$} & $11.690$\sstd{$10.602$} & \highlight{$0.339$\sstd{$0.003$}} & $0.344$\sstd{$0.002$} & $0.397$\sstd{$0.026$} \\
& & Transformer &$1.471$\sstd{$0.016$} & $1.226$\sstd{$0.004$} & $5.194$\sstd{$0.320$} & $0.346$\sstd{$0.002$} & $0.347$\sstd{$0.001$} & $0.480$\sstd{$0.030$} \\
\cmidrule[\heavyrulewidth]{1-9}
    \end{tabular}
    \caption{\textbf{Model Misspecification}. Results for model misspecification under different training data $\chi_{sim}$, when evaluated under MLP-TanH and Linear Data ($\chi_{real}$), with the underlying model as a linear and MLP-TanH model respectively.}
    \vspace{-4mm}
    \label{tab:misspec_model}
\end{table*}
\begin{table*}[t]
    \centering
    \footnotesize	    
    \def\arraystretch{1.25}
    \setlength{\tabcolsep}{5pt}
    \begin{tabular}{lcr ccc cccc}
        \cmidrule[\heavyrulewidth]{1-9}
         &  &  & \multicolumn{6}{c}{\textit{$L_2$ Loss} ($\downarrow$)} \\
        \cmidrule(lr){4-9}
        \textbf{Objective} & $q_\varphi$ & \textbf{Model} & \multicolumn{3}{c}{\textbf{Linear Model $|$ GP Data}} & \multicolumn{3}{c}{\textbf{MLP-TanH Model $|$ GP Data}} & $\leftarrow\chi_{real}$ \\
        \cmidrule(lr){4-6}\cmidrule(lr){7-9}
        & & & \textit{LR} & \textit{NLR} & \textit{GP} & \textit{LR} & \textit{NLR} & \textit{GP}  & $\leftarrow\chi_{sim}$ \\
        \cmidrule{1-9}
\multirow{4}{*}{Baseline} & - & Random & -  & -  & $2.681$\sstd{$0.089$} &  -  & -  & $16.236$\sstd{$0.381$} \\
& - & Optimization & -  & -  & $0.263$\sstd{$0.000$} & -  & -  & $0.007$\sstd{$0.000$} \\
& - & Langevin & -  & -  & $0.266$\sstd{$0.001$} & -  & -  & $0.022$\sstd{$0.001$} \\
& - & HMC & -  & -  & $0.266$\sstd{$0.000$} & -  & -  & $0.090$\sstd{$0.002$} \\
\cmidrule{2-9}
\multirow{3}{*}{Fwd-KL} & \multirow{6}{*}{\rotatebox[origin=c]{90}{Gaussian}} & GRU &$0.268$\sstd{$0.000$} & -  & -  & -  & $14.077$\sstd{$0.368$} & -  \\
& & DeepSets &$0.269$\sstd{$0.001$} & -  & -  & -  & $14.756$\sstd{$0.280$} & -  \\
& & Transformer &$0.270$\sstd{$0.001$} & -  & -  & -  & $14.733$\sstd{$0.513$} & -  \\
\cmidrule{3-9}
\multirow{3}{*}{Rev-KL} & & GRU &$0.268$\sstd{$0.000$} & $0.269$\sstd{$0.000$} & $0.266$\sstd{$0.000$} & $0.334$\sstd{$0.005$} & $0.157$\sstd{$0.003$} & $0.080$\sstd{$0.003$} \\
& & DeepSets &$0.269$\sstd{$0.000$} & $0.269$\sstd{$0.000$} & \highlight{$0.265$\sstd{$0.000$}} & $0.331$\sstd{$0.003$} & $0.146$\sstd{$0.002$} & $0.063$\sstd{$0.000$} \\
& & Transformer &$0.269$\sstd{$0.000$} & $0.269$\sstd{$0.000$} & $0.267$\sstd{$0.000$} & $0.310$\sstd{$0.013$} & $0.155$\sstd{$0.006$} & $0.066$\sstd{$0.004$} \\
\cmidrule{2-9}
\multirow{3}{*}{Fwd-KL} & \multirow{6}{*}{\rotatebox[origin=c]{90}{Flow}} & GRU &$0.268$\sstd{$0.000$} & -  & -  & -  & $9.756$\sstd{$0.192$} & -  \\
& & DeepSets &$0.269$\sstd{$0.001$} & -  & -  & -  & $14.345$\sstd{$0.628$} & -  \\
& & Transformer &$0.269$\sstd{$0.000$} & -  & -  & -  & $8.557$\sstd{$0.561$} & -  \\
\cmidrule{3-9}
\multirow{3}{*}{Rev-KL} & & GRU &$0.268$\sstd{$0.000$} & $0.270$\sstd{$0.001$} & $0.266$\sstd{$0.000$} & $0.289$\sstd{$0.011$} & $0.120$\sstd{$0.004$} & $0.059$\sstd{$0.003$} \\
& & DeepSets &$0.269$\sstd{$0.000$} & $0.269$\sstd{$0.001$} & $0.266$\sstd{$0.000$} & $0.270$\sstd{$0.008$} & $0.115$\sstd{$0.002$} & $0.059$\sstd{$0.002$} \\
& & Transformer &$0.269$\sstd{$0.001$} & $0.270$\sstd{$0.000$} & $0.267$\sstd{$0.000$} & $0.293$\sstd{$0.008$} & $0.120$\sstd{$0.005$} & \highlight{$0.055$\sstd{$0.002$}} \\
\cmidrule[\heavyrulewidth]{1-9}
    \end{tabular}
    \caption{\textbf{Model Misspecification}. Results for model misspecification under different training data $\chi_{sim}$, when evaluated under GP Data ($\chi_{real}$), with the underlying model as a linear and MLP-TanH model respectively.}
    \vspace{-4mm}
    \label{tab:misspec_gp}
\end{table*}
\subsection{Variable-Dim}
\label{appdx:details_max_dim}
In this section, we provide the experimental details relevant to reproducing the results of Section~\ref{sec:experiments}. All the models are trained with streaming data from the underlying probabilistic model, such that every iteration of training sees a new set of datasets. Training is done with a batch size of $128$, representing the number of datasets seen during one optimization step. Further, we ensure that the datasets sampled resemble a uniform distribution over the feature dimensions, ranging from $1$-dimensional to the maximal dimensional setup. Evaluations are done with $25$ samples and we ensure that the test datasets used for each probabilistic model are the same across all the compared methods, i.e., baselines, forward KL, and reverse KL. We train the amortized inference model and the forward KL baselines for the following different probabilistic models:

\textbf{Mean of Gaussian (GM):} We train the amortization models over $50,000$ iterations using a linear warmup with $12,5000$ iterations over which the weight of the KL term in our proposed approach scales linearly from $0$ to $1$. We use an identity covariance matrix for the data-generating process, but it can be easily extended to the case of correlated or diagonal covariance-based Gaussian distributions. In this setup, we consider a maximum of $100$ feature dimensions.

\textbf{Gaussian Mixture Model (GMM):} We train the mixture model setup for $500,000$ iterations with $125,000$ iterations of warmup. We set the maximal feature dimensions as $5$ and experiment with $2$ and $5$ mixture components. While we do use an identity covariance matrix for the data-generating process, again, it can be easily extended to other cases.

\textbf{Linear Regression (LR):} The amortization models for this setup are trained for $100,000$ iterations with $25,000$ iterations of warmup. The maximal feature dimension considered for this task is $100$-dimensional, and the predictive variance $\sigma^2$ is assumed to be known and set as $0.25$.

\textbf{Nonlinear Regression (NLR):} We train the setup for $250,000$ iterations with $62,500$ iterations consisting of warmup. The maximal feature dimension considered is $100$-dimensional, and training is done with a known predictive variance similar to the LR setup. For the probabilistic model, we consider both a $1$-layered and a $2$-layered multi-layer perceptron (MLP) network with 32 hidden units in each, and either a \textsc{relu} or \textsc{tanh} activation function.

\textbf{Linear Classification (LC):} We experiment with a maximal $100$-dimensional setup with training done for $100,000$ iterations, out of which $25,000$ are used for warmup. Further, we train for both binary classification as well as a $5$-class classification setup.

\textbf{Nonlinear Classification (NLC):} We experiment with a maximal $100$-dimensional setup with training done for $250,000$ iterations, out of which $62,500$ are used for warmup. Further, we train for both binary classification as well as a $5$-class classification setup. For the probabilistic model, we consider both a $1$-layered and a $2$-layered multi-layer perceptron (MLP) network with 32 hidden units in each, and either a \textsc{relu} or \textsc{tanh} activation function.

\begin{figure*}
    \centering
    \captionsetup[subfigure]{font=scriptsize}
    \includegraphics[width=\textwidth]{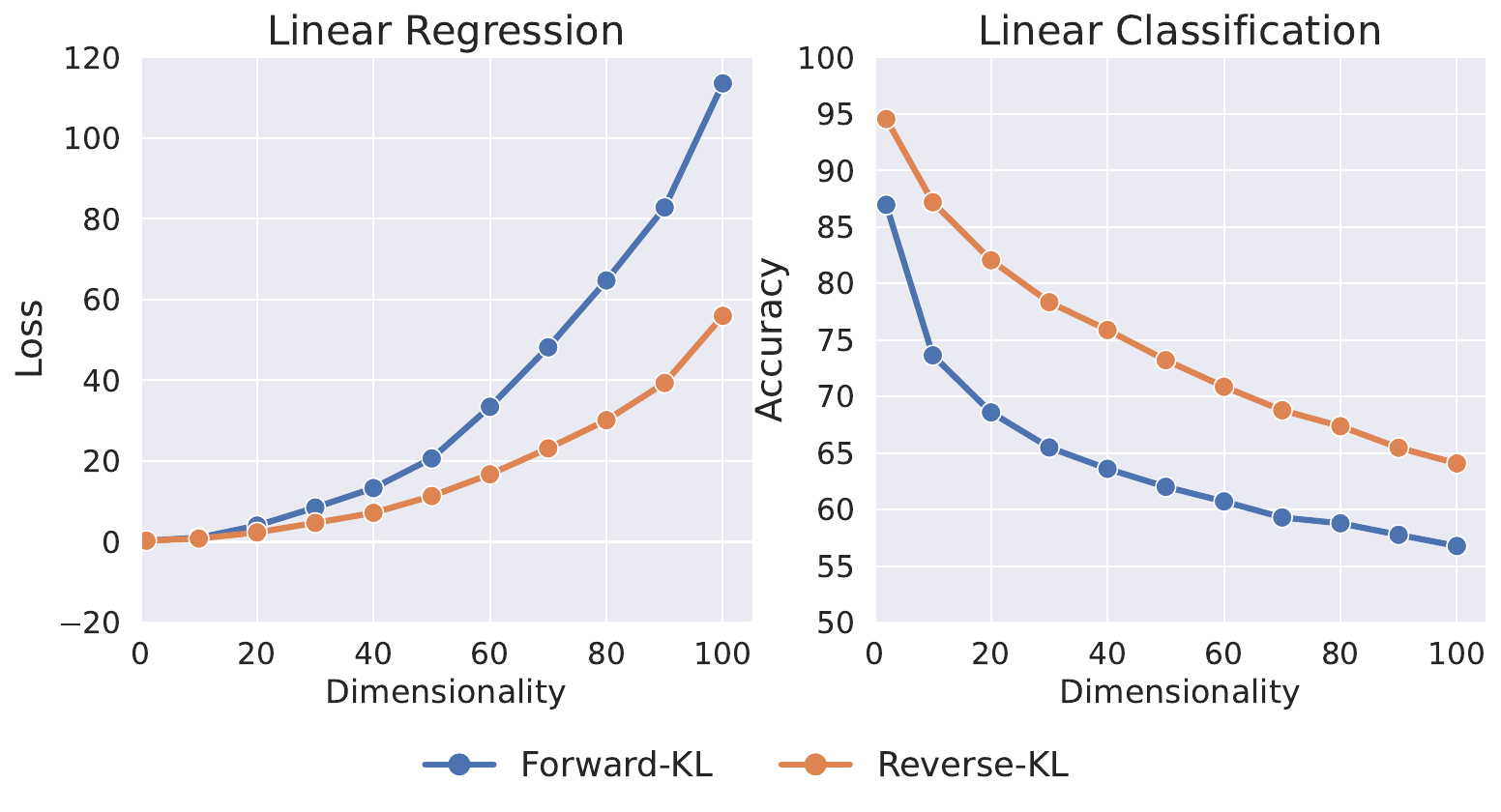}
    \vspace{-7mm}
    \caption{\textbf{Trends of Performance over different Dimensions in Variable Dimensionality Setup:} We see that our proposed reverse KL methodology outperforms the forward KL one.}
    \vspace{-5mm}
    \label{fig:dim_kl}
\end{figure*}

\begin{figure*}
    \centering
    \captionsetup[subfigure]{font=scriptsize}
    \includegraphics[width=\textwidth]{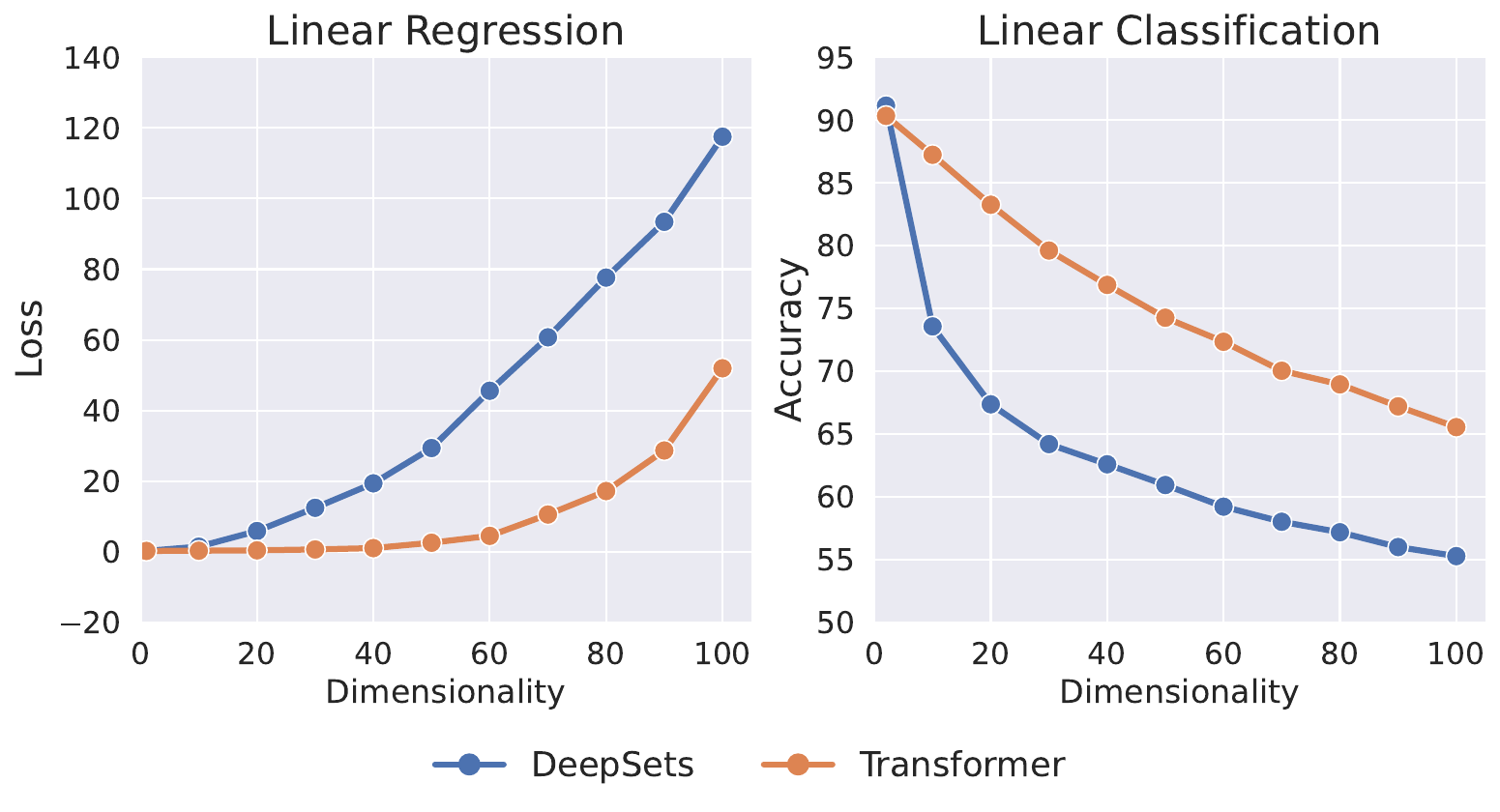}
    \vspace{-7mm}
    \caption{\textbf{Trends of Performance over different Dimensions in Variable Dimensionality Setup:} We see that transformer models generalize better to different dimensional inputs than DeepSets.}
    \vspace{-5mm}
    \label{fig:dim_kl}
\end{figure*}

\begin{figure*}
    \centering
    \captionsetup[subfigure]{font=scriptsize}
    \includegraphics[width=\textwidth]{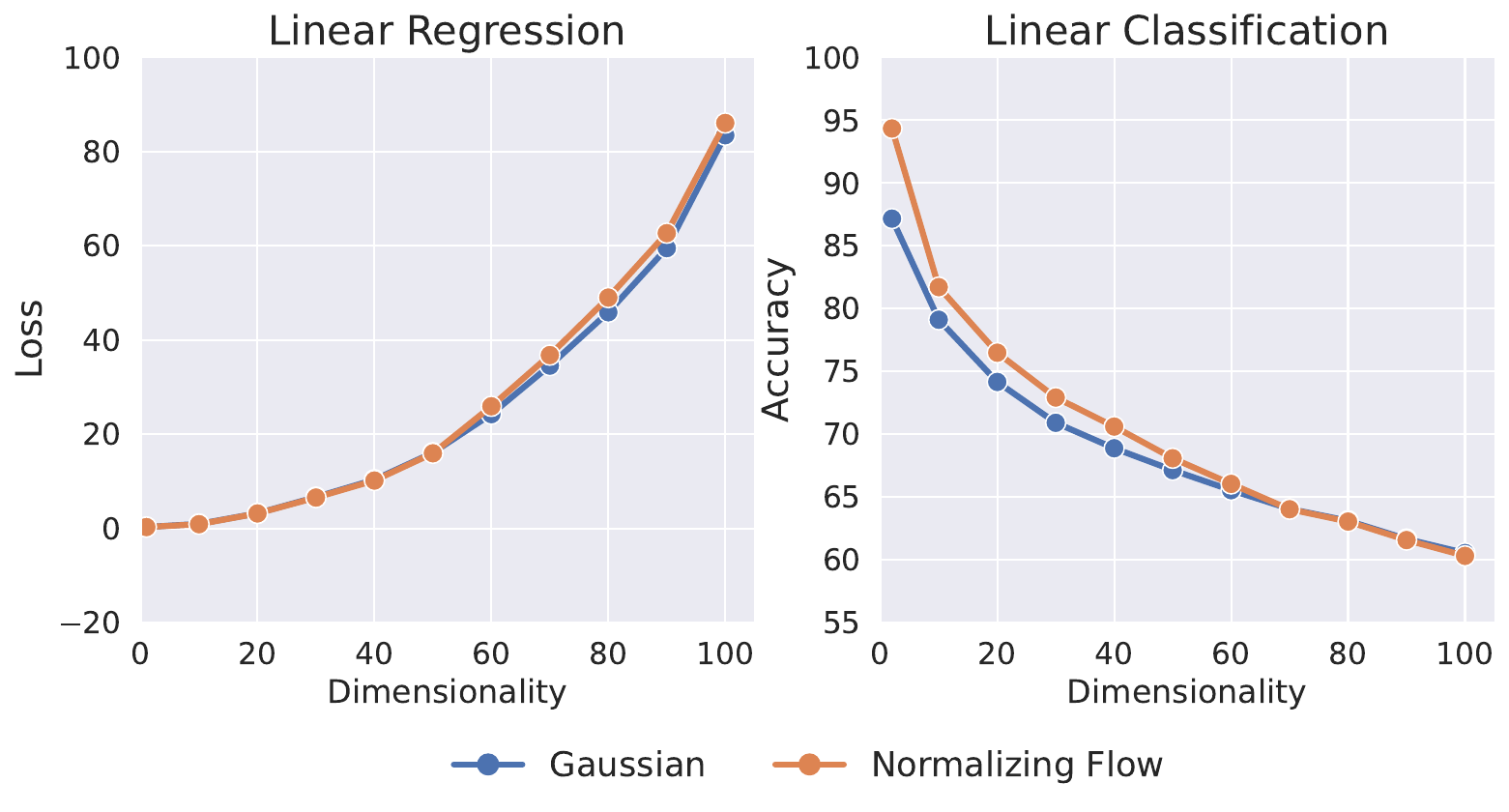}
    \vspace{-7mm}
    \caption{\textbf{Trends of Performance over different Dimensions in Variable Dimensionality Setup:} We see that normalizing flows leads to similar performances than Gaussian based variational approximation.}
    \vspace{-5mm}
    \label{fig:dim_kl}
\end{figure*}
\subsection{Model Misspecification}
\label{appdx:details_misspecification}
In this section, we provide the experimental details relevant to reproducing the results of Section~\ref{sec:experiments}.
All models during this experiment are trained with streaming data from the currently used dataset-generating function $\chi$, such that every iteration of training sees a new batch of datasets. Training is done with a batch size of $128$, representing the number of datasets seen during one optimization step. Evaluation for all models is done with $10$ samples from each dataset-generator used in the respective experimental subsection and we ensure that the test datasets are the same across all compared methods, i.e., baselines, forward KL, and reverse KL.

\textbf{Linear Regression Model:} The linear regression amortization models are trained following the training setting for linear regression fixed dimensionality, that is, $50,000$ training iterations with $12,500$ iterations of warmup. The feature dimension considered for this task is $1$-dimension. The model is trained separately on datasets from three different generators $\chi$: linear regression, nonlinear regression, and Gaussian processes, and evaluated after training on test datasets from all of them.
For training with datasets from the linear regression probabilistic model, the predictive variance $\sigma^2$ is assumed to be known and set as $0.25$. 
The same variance is used for generating datasets from the nonlinear regression dataset generator with $1$ layer, $32$ hidden units, and \textsc{tanh} activation function. 
Lastly, datasets from the Gaussian process-based generator are sampled similarly, using the GPytorch library~\cite{gardner2018gpytorch}, where datasets are sampled of varying cardinality, ranging from $64$ to $128$. We use a zero-mean Gaussian Process (GP) with a unit lengthscale radial-basis function (RBF) kernel serving as the covariance matrix. Further, we use a very small noise of $\sigma^2 = 1\mathrm{e}^{-6}$ in the likelihood term of the GP.
Forward KL training in this experiment can only be done when the amortization model and the dataset-generating function are the same: when we train on datasets from the linear regression-based $\chi$. Table \ref{tab:misspec_model} provides a detailed overview of the results.

\textbf{Nonlinear Regression Models:} The nonlinear regression amortization models are trained following the training setting for nonlinear regression fixed dimensionality, that is, $100,000$ training iterations with $25,000$ iterations of warmup. Here, we consider two single-layer perceptions with 32 hidden units with a \textsc{tanh} activation function. The feature dimensionality considered is $1$ dimension.
We consider the same dataset-generating functions as in the misspecification experiment for a linear regression model above. However, the activation function used in the nonlinear regression dataset generator matches the activation function of the currently trained amortization model. In this case, forward KL training is possible in the two instances when trained on datasets from the corresponding nonlinear regression probabilistic model. A more detailed overview of the results can be found in Table \ref{tab:misspec_model} and \ref{tab:misspec_gp}.

\begin{figure}
    \centering
    \includegraphics[width=\textwidth]{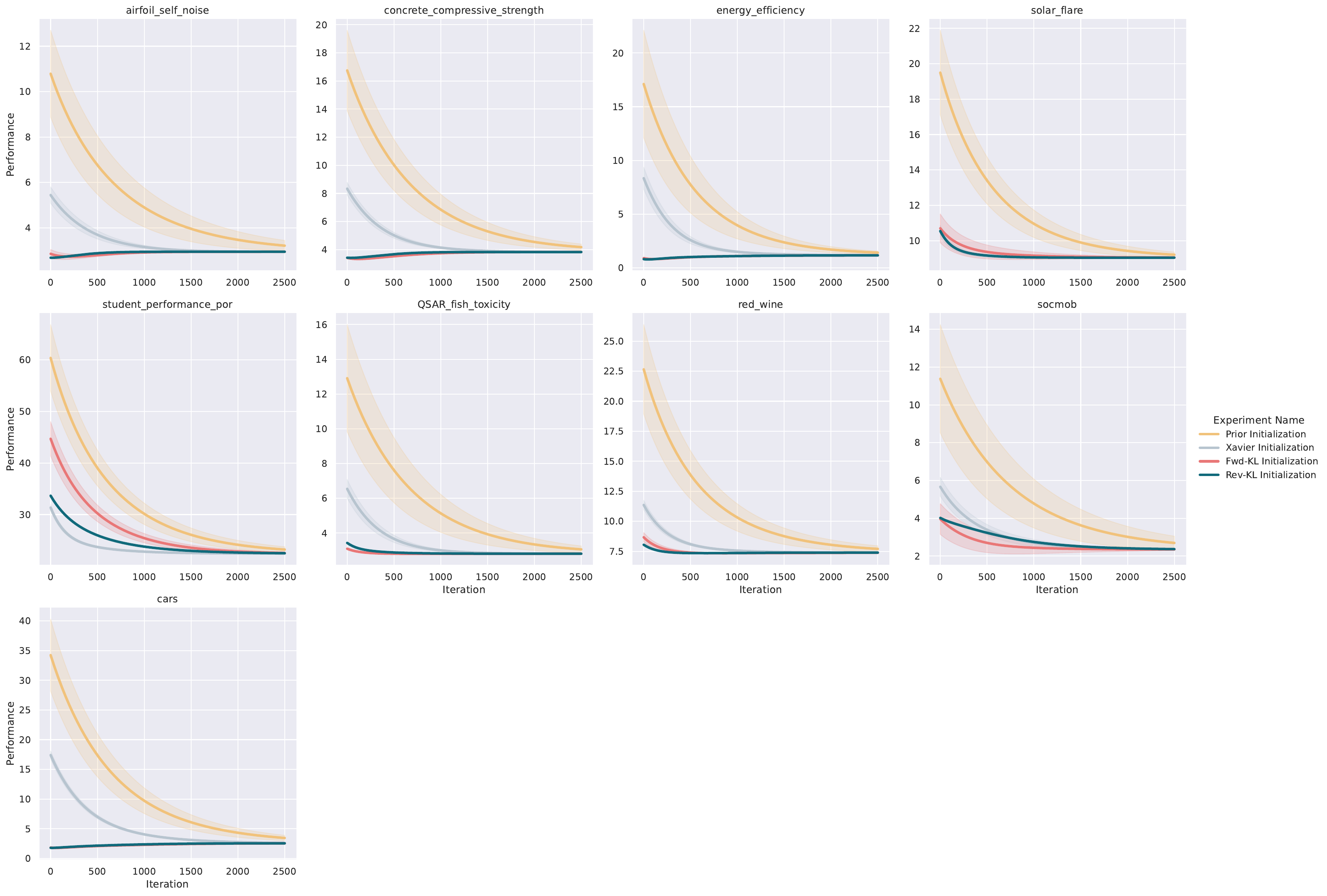}
    \caption{\textbf{Tabular Experiments $|$ Linear Regression with Diagonal Gaussian}: For every regression dataset from the OpenML platform considered, we initialize the parameters of a linear regression-based probabilistic model with the amortized inference models which were trained with a diagonal Gaussian assumption. The parameters are then further trained with maximum-a-posteriori (MAP) estimate with gradient descent. Reverse and Forward KL denote initialization with the correspondingly trained amortized model. Prior refers to a MAP-based optimization baseline initialized from the prior $\gN(0, I)$, whereas Xavier refers to initialization from the Xavier initialization scheme.}
    \label{fig:regression_linear_vanilla}
\end{figure}

\subsection{Tabular Experiments}
\label{appdx:details_tabular}
For the tabular experiments, we train the amortized inference models for (non-)linear regression (NLR/LR) as well as (non-)linear classification (NLC/LC) with $\vx \sim \mathcal{N}(\mathbf{0}, \mathbf{I})$ as opposed to $\vx \sim \gU(-\mathbf{1}, \mathbf{1})$ in the dataset generating process $\chi$, with the rest of the settings the same as \textsc{maximum-dim} experiments. For the nonlinear setups, we only consider the \textsc{relu} case as it has seen predominant success in deep learning. Further, we only consider a 1-hidden layer neural network with 32 hidden dimensions in the probabilistic model. 

After having trained the amortized inference models, both for forward and reverse KL setups, we evaluate them on real-world tabular datasets. We first collect a subset of tabular datasets from the OpenML platform as outlined in Appendix~\ref{appdx:datasets}. Then, for each dataset, we perform a 5-fold cross-validation evaluation where the dataset is chunked into $5$ bins, of which, at any time, $4$ are used for training and one for evaluation. This procedure is repeated five times so that every chunk is used for evaluation once.

For each dataset, we normalize the observations and the targets so that they have zero mean and unit standard deviation. For the classification setups, we only normalize the inputs as the targets are categorical. For both forward KL and reverse KL amortization models, we initialize the probabilistic model from samples from the amortized model and then further finetune it via dataset-specific maximum a posteriori optimization. We repeat this setup over $25$ different samples from the inference model. In contrast, for the optimization baseline, we initialize the probabilistic models' parameters from $\gN(0, I)$, which is the prior that we consider, and then train 25 such models with maximum a posteriori objective using Adam optimizer. 

While we see that the amortization models, particularly the reverse KL model, lead to much better initialization and convergence, it is important to note that the benefits vanish if we initialize using the Xavier-init initialization scheme. However, we believe that this is not a fair comparison as it means that we are considering a different prior now, while the amortized models were trained with $\gN(0, I)$ prior. We defer the readers to the section below for additional discussion and experimental results.

\begin{figure}
    \centering
    \includegraphics[width=\textwidth]{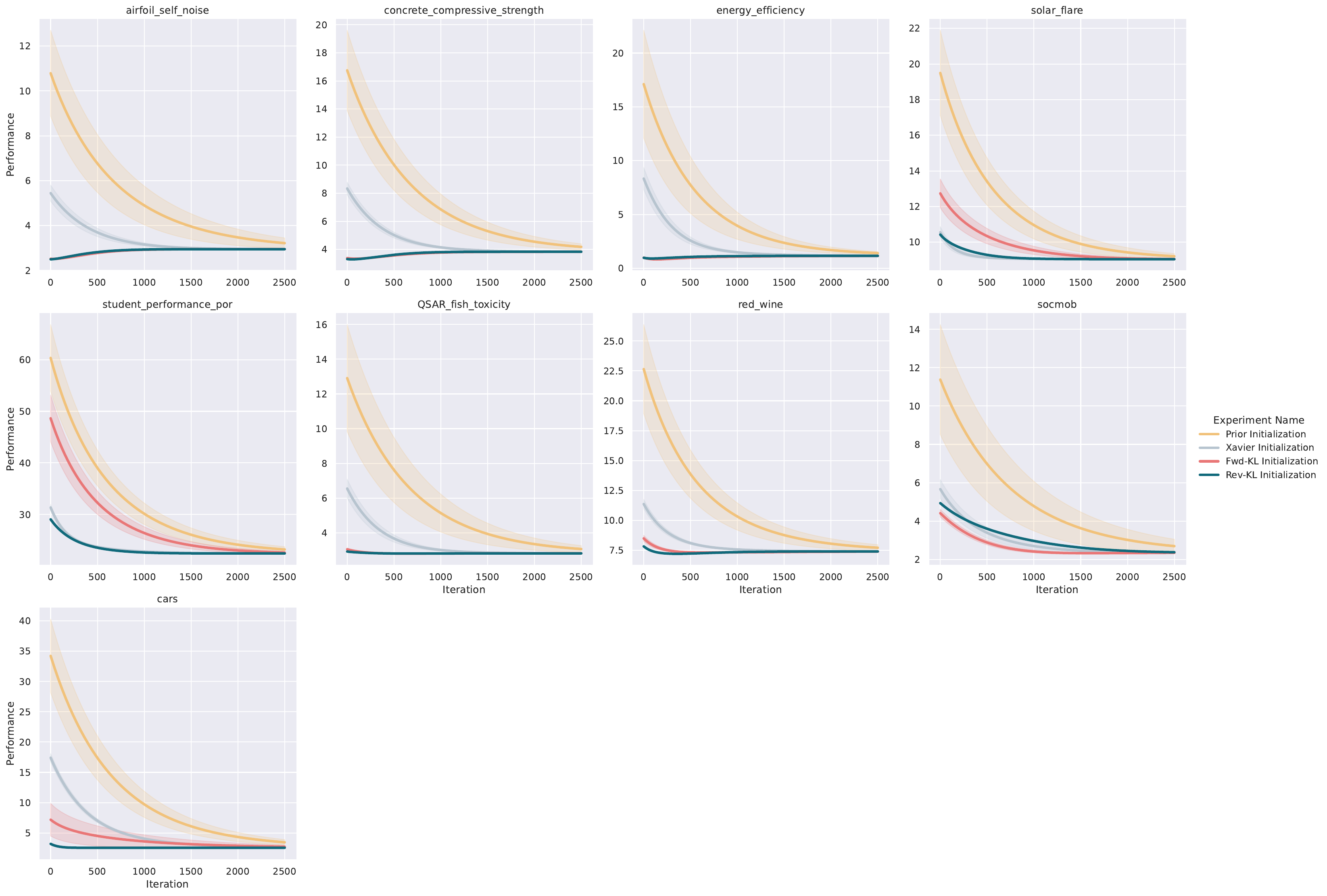}
    \caption{\textbf{Tabular Experiments $|$ Linear Regression with Normalizing Flow}: For every regression dataset from the OpenML platform considered, we initialize the parameters of a linear regression-based probabilistic model with the amortized inference models which were trained with a normalizing flow-based model. The parameters are then further trained with maximum-a-posteriori (MAP) estimate with gradient descent. Reverse and Forward KL denote initialization with the correspondingly trained amortized model. Prior refers to a MAP-based optimization baseline initialized from the prior $\gN(0, I)$, whereas Xavier refers to initialization from the Xavier initialization scheme.}
    \label{fig:regression_linear_flow}
\end{figure}
\section{OpenML Datasets}
\label{appdx:datasets}

For the tabular regression problems, we consider the suite of tasks outlined in \textit{OpenML-CTR23 - A curated tabular regression benchmarking suite}~\citep{fischer2023openmlctr23}, from which we further filter out datasets that have more than 2000 examples and 100 features. We also remove datasets with missing information and NaNs. Similarly, we consider the \textit{OpenML-CC18 Curated Classification benchmark
}~\citep{bischl2019openmlcc18} suite of tasks for classification and perform a similar filtering algorithm. We remove datasets with missing information and NaNs, as well as datasets with more than 2000 examples and 100 features. In addition, we also exclude datasets that are not made for binary classification. At the end of this filtering mechanism, we end up with 9 regression and 13 classification problems, and our dataset filtration pipeline is heavily inspired by~\cite{hollmann2022tabpfn}. We provide the datasets considered for both regression and classification below:

\textbf{Regression}: \textsc{airfoil\_self\_noise}, \textsc{concrete\_compressive\_strength}, \textsc{energy\_efficiency}, \textsc{solar\_flare}, \textsc{student\_performance\_por}, \textsc{QSAR\_fish\_toxicity}, \textsc{red\_wine}, \textsc{socmob} and \textsc{cars}.

\textbf{Classification}: \textsc{credit-g}, \textsc{diabetes}, \textsc{tic-tac-toe}, \textsc{pc4}, \textsc{pc3}, \textsc{kc2}, \textsc{pc1}, \textsc{banknote-authentication}, \textsc{blood-transfusion-service-center}, \textsc{ilpd}, \textsc{qsar-biodeg}, \textsc{wdbc} and \textsc{climate-model-simulation-crashes}.
\begin{figure}
    \centering
    \includegraphics[width=\textwidth]{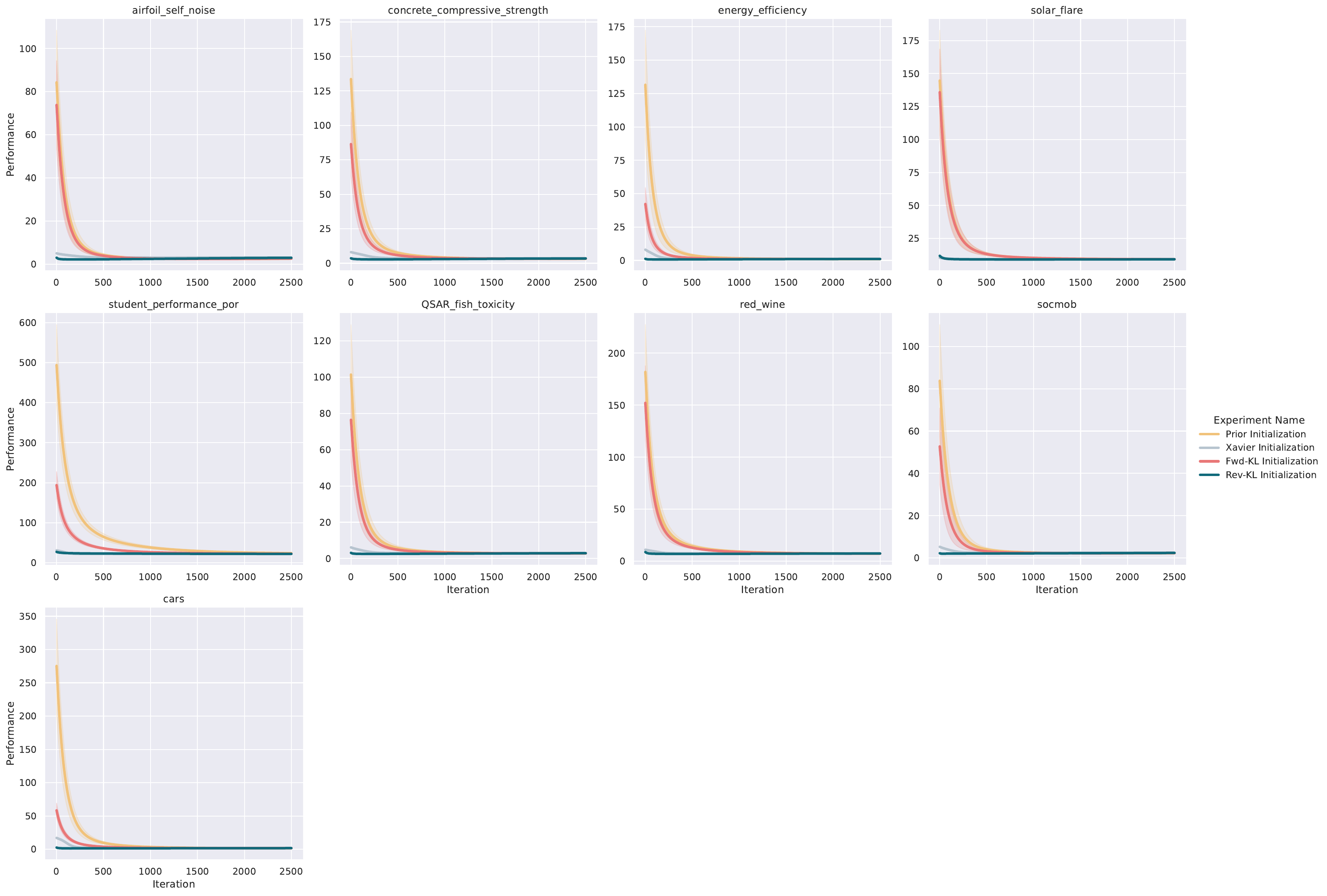}
    \caption{\textbf{Tabular Experiments $|$ Nonlinear Regression with Diagonal Gaussian}: For every regression dataset from the OpenML platform considered, we initialize the parameters of a nonlinear regression-based probabilistic model with the amortized inference models which were trained with a diagonal Gaussian assumption. The parameters are then further trained with maximum-a-posteriori (MAP) estimate with gradient descent. Reverse and Forward KL denote initialization with the correspondingly trained amortized model. Prior refers to a MAP-based optimization baseline initialized from the prior $\gN(0, I)$, whereas Xavier refers to initialization from the Xavier initialization scheme.}
    \vspace{-5mm}
    \label{fig:regression_nonlinear_vanilla}
\end{figure}
\section{Additional Experiments}
\label{appdx:results}
In this section, we outline the additional experiments we conducted in obtaining Bayesian posteriors for the different probabilistic models for different hyperparameters and their downstream uses. We provide a comprehensive account of the results in the relevant sections below.

\subsection{Fixed-Dim}
\label{appdx:results_fixdim}
While we highlighted the results with the Gaussian mixture model and classification settings with only 2 clusters/classes, we also conducted experiments with an increased number of clusters and classes, making the problem even more challenging. Tables~\ref{tab:apdx_gaussian}-\ref{tab:apdx_nlc_5cl} shows that both forward and reverse KL methods perform reasonably, with forward KL struggling more in challenging scenarios.

Next, we also consider harder tasks based on the Bayesian Neural Network (BNN) paradigm, where we consider nonlinear regression and classification setups with different activation functions: \textsc{tanh} and \textsc{relu} for a 1-layered and 2-layered BNN. We provide the results of our experiments in Tables~\ref{tab:apdx_gaussian}-\ref{tab:apdx_nlc_5cl}. The results indicate that forward KL approaches struggle a lot in such scenarios, often achieving performance comparable to random chance. On the contrary, we see that reverse KL-based amortization leads to performances often similar to dataset-specific optimization, thereby showing the superiority of our proposed method.

\begin{figure}
    \centering
    \includegraphics[width=\textwidth]{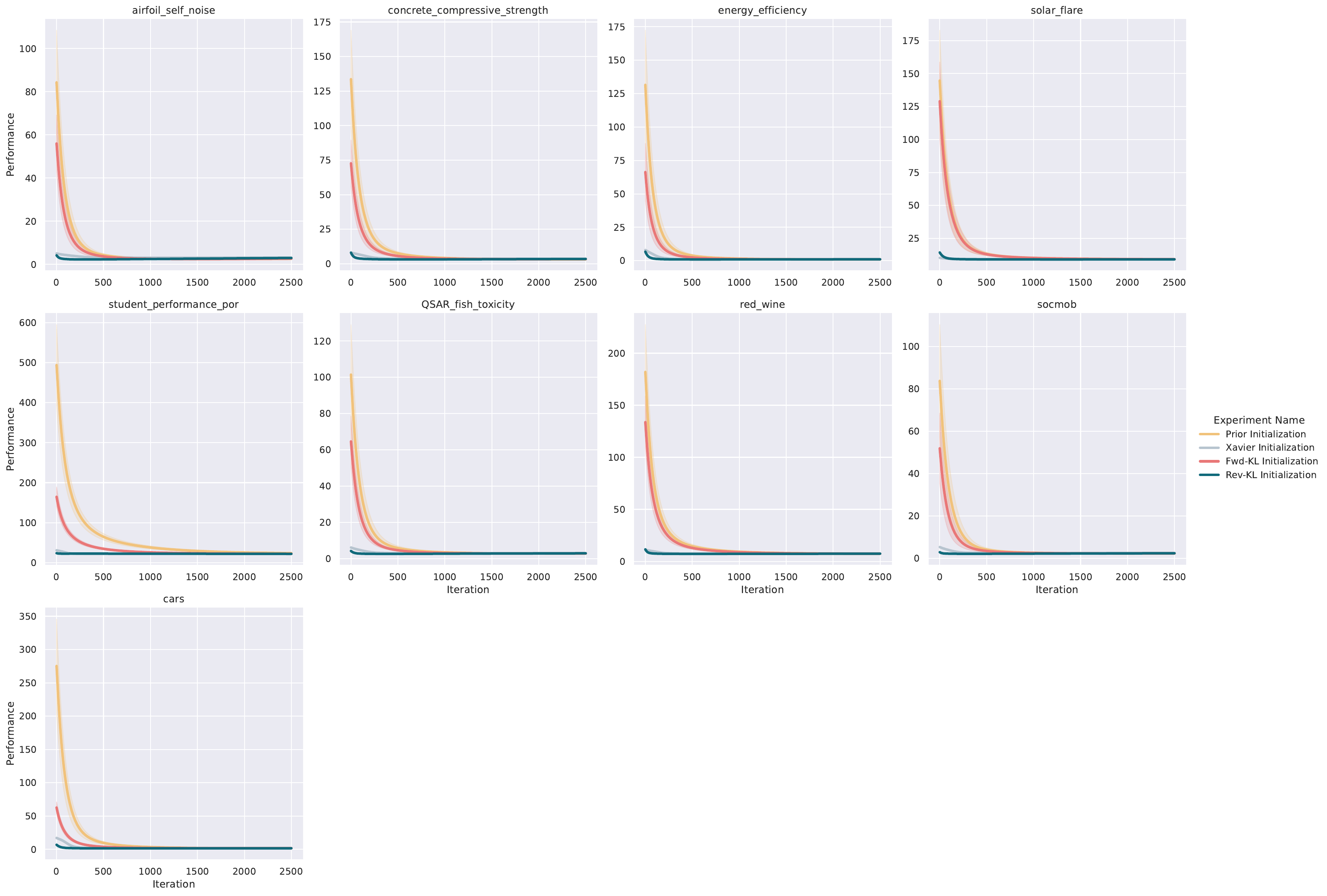}
    \caption{\textbf{Tabular Experiments $|$ Nonlinear Regression with Normalizing Flow}: For every regression dataset from the OpenML platform considered, we initialize the parameters of a nonlinear regression-based probabilistic model with the amortized inference models which were trained with a normalizing flow-based model. The parameters are then further trained with maximum-a-posteriori (MAP) estimate with gradient descent. Reverse and Forward KL denote initialization with the correspondingly trained amortized model. Prior refers to a MAP-based optimization baseline initialized from the prior $\gN(0, I)$, whereas Xavier refers to initialization from the Xavier initialization scheme.}
    \vspace{-5mm}
\label{fig:regression_nonlinear_flow}
\end{figure}
\subsection{Variable-Dim}
\label{appdx:results_maxdim}
Our experiments on variable dimensional datasets can be evaluated for arbitrary feature cardinality, of which we show a few examples in Section~\ref{sec:experiments}. In this section, we provide results for additional dimensionality setups. In particular, we refer the readers to Tables~\ref{tab:variable_apdx_gaussian}-\ref{tab:variable_apdx_nlc_5cl}, which contain experimental results w.r.t different dimensionalities (e.g. 50D setup), as well as different number of clusters and classes, respectively, for the GMM and LC setup. Throughout, we see that amortization leads to reasonable performance, and in particular, we see forward KL-based amortization starting to struggle in high-dimensional setups.

Again, to make the setup more challenging, we consider the Bayesian Neural Network (BNN) setup where we consider nonlinear regression and classification with different activation functions: \textsc{tanh} and \textsc{relu} for a 1-layered and 2-layered BNN, but which can now be tested for an arbitrary number of input features. Our experiments are highlighted in Tables~\ref{tab:variable_apdx_gaussian}-\ref{tab:variable_apdx_nlc_5cl}, for 1- and 2-layered BNN, among others. In such complex multi-modal and complicated setups, forward KL often performs comparable to random chance and thus does not lead to any good approximation of the true posterior distribution. On the other hand, our proposed method indeed leads to good predictive performance, often comparable to dataset-specific optimization routines.

\subsection{Model Misspecification}
\label{appdx:results_missspecification}
As a representative of the results on model misspecification (Section \ref{sec:experiments}), we highlighted training and evaluation of the amortization models with Transformer backbone on a subset of in-distribution and OoD data-generating functions (Table \ref{tab:misspecification}) to show superiority in generalization of reverse KL trained system vs. forward KL based ones on OoD data but also to highlight that training a misspecified amortization model on OoD datasets directly with our approach results in even better posterior predictive performance.

\subsection{Tabular Experiments}
\label{appdx:results_tabular}
\begin{figure}
    \centering
    \includegraphics[width=1.\textwidth]{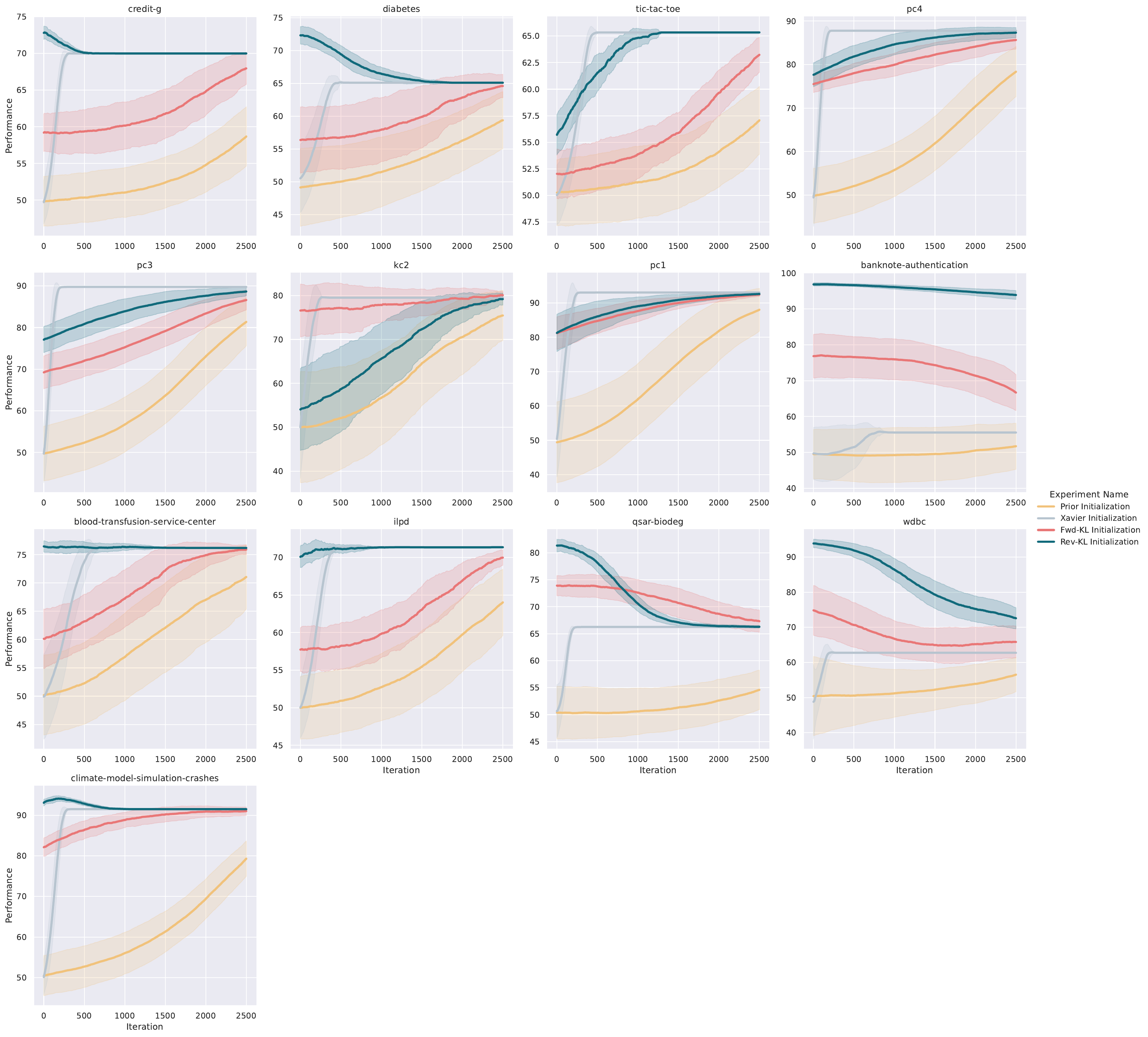}
    \caption{\textbf{Tabular Experiments $|$ Linear Classification with Diagonal Gaussian}: For every classification dataset from the OpenML platform considered, we initialize the parameters of a linear classification-based probabilistic model with the amortized inference models which were trained with a diagonal Gaussian assumption. The parameters are then further trained with maximum-a-posteriori (MAP) estimate with gradient descent. Reverse and Forward KL denote initialization with the correspondingly trained amortized model. Prior refers to a MAP-based optimization baseline initialized from the prior $\gN(0, I)$, whereas Xavier refers to initialization from the Xavier initialization scheme.}
    \vspace{-5mm}\label{fig:classification_linear_vanilla}
\end{figure}

\begin{figure}
    \centering    \includegraphics[width=1.\textwidth]{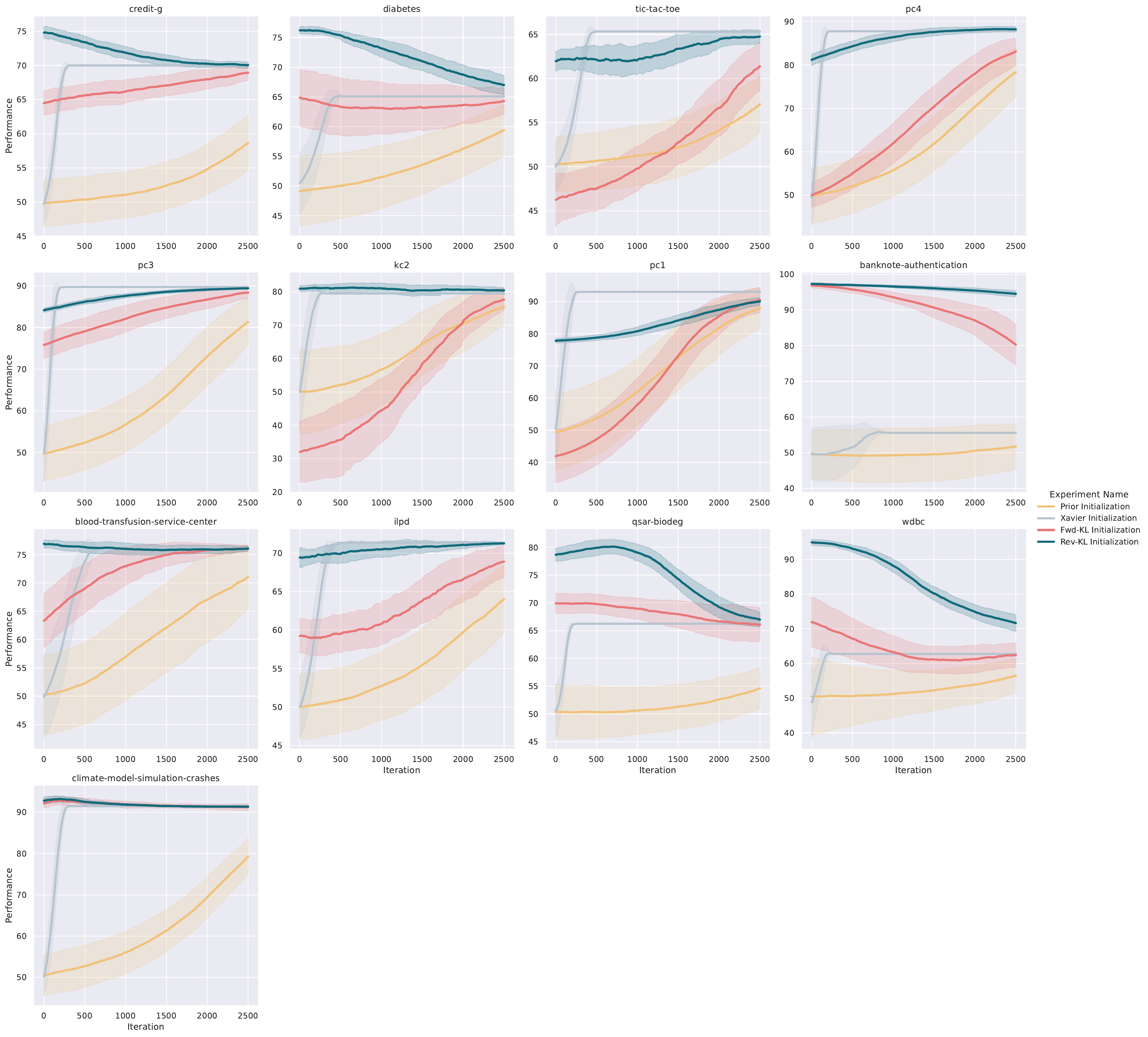}
    \caption{\textbf{Tabular Experiments $|$ Linear Classification with Normalizing Flow}: For every classification dataset from the OpenML platform considered, we initialize the parameters of a linear classification-based probabilistic model with the amortized inference models which were trained with a normalizing flow-based model. The parameters are then further trained with maximum-a-posteriori (MAP) estimate with gradient descent. Reverse and Forward KL denote initialization with the correspondingly trained amortized model. Prior refers to a MAP-based optimization baseline initialized from the prior $\gN(0, I)$, whereas Xavier refers to initialization from the Xavier initialization scheme.}
    \vspace{-5mm}
    \label{fig:classification_linear_flow}
\end{figure}

\begin{figure}
    \centering
    \includegraphics[width=1.\textwidth]{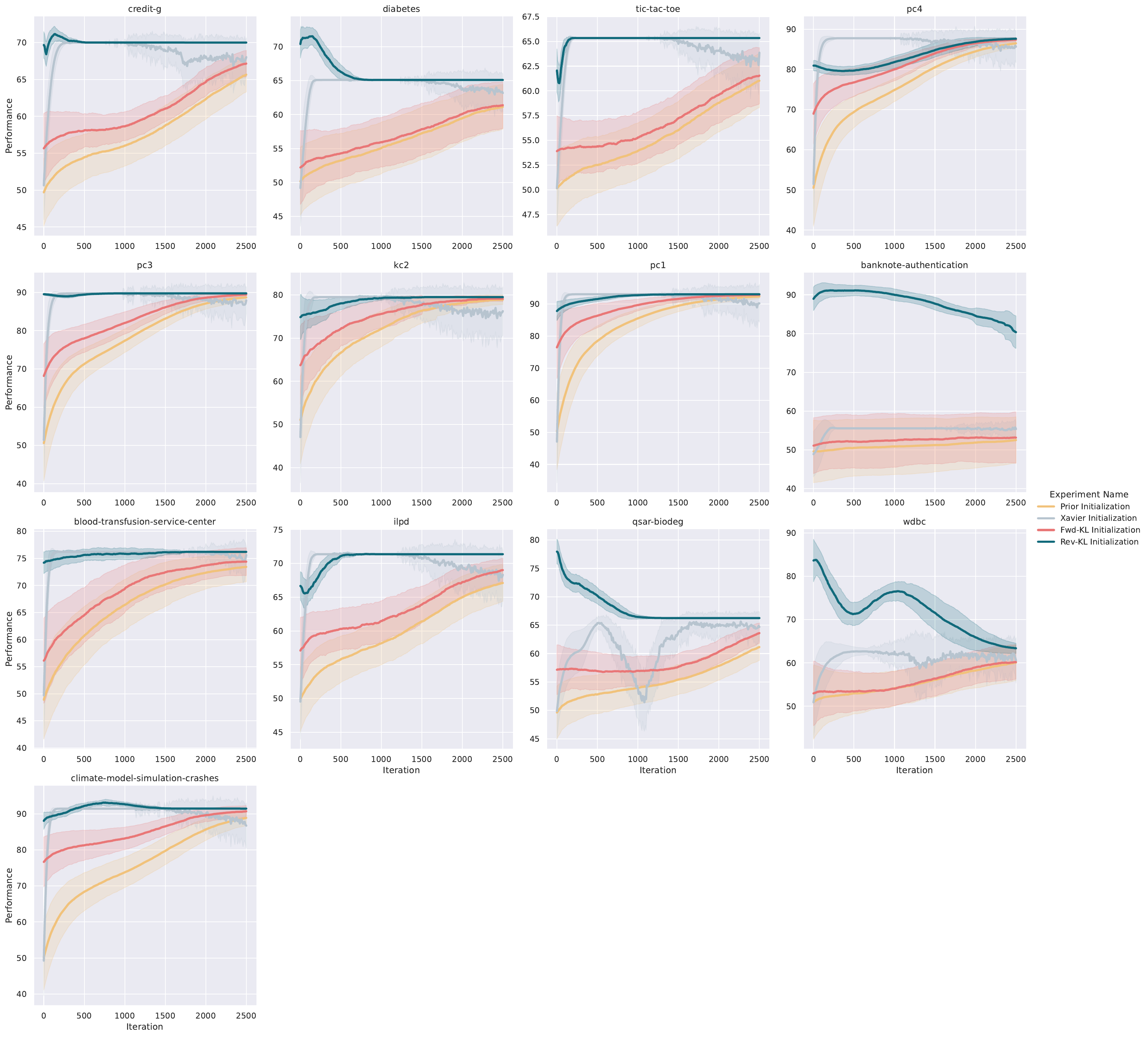}
    \caption{\textbf{Tabular Experiments $|$ Nonlinear Classification with Diagonal Gaussian}: For every classification dataset from the OpenML platform considered, we initialize the parameters of a nonlinear classification-based probabilistic model with the amortized inference models which were trained with a diagonal Gaussian assumption. The parameters are then further trained with maximum-a-posteriori (MAP) estimate with gradient descent. Reverse and Forward KL denote initialization with the correspondingly trained amortized model. Prior refers to a MAP-based optimization baseline initialized from the prior $\gN(0, I)$, whereas Xavier refers to initialization from the Xavier initialization scheme.}
    \vspace{-5mm}
    \label{fig:classification_nonlinear_vanilla}
\end{figure}

\begin{figure}
    \centering
    \includegraphics[width=1.\textwidth]{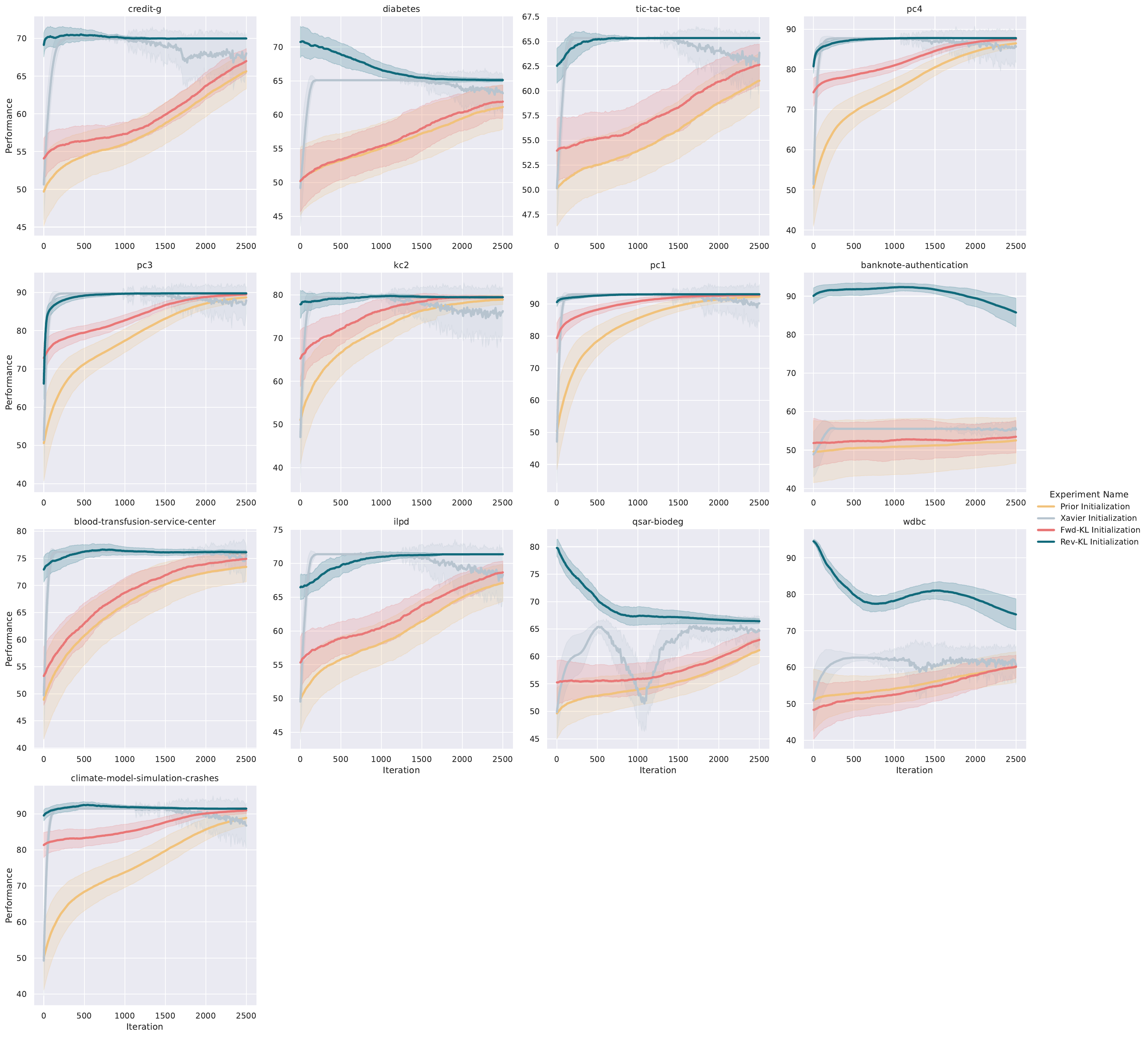}
    \caption{\textbf{Tabular Experiments $|$ Nonlinear Classification with Normalizing Flow}: For every classification dataset from the OpenML platform considered, we initialize the parameters of a linear classification-based probabilistic model with the amortized inference models which were trained with a normalizing flow-based model. The parameters are then further trained with maximum-a-posteriori (MAP) estimate with gradient descent. Reverse and Forward KL denote initialization with the correspondingly trained amortized model. Prior refers to a MAP-based optimization baseline initialized from the prior $\gN(0, I)$, whereas Xavier refers to initialization from the Xavier initialization scheme.}
    \vspace{-5mm}
    \label{fig:classification_nonlinear_flow}
\end{figure}

As a case of extreme OoD generalization, we test our amortized models trained to handle variable feature dimensions on the suite of regression and classification problems that we filtered out from the OpenML platform, as outlined in Appendix~\ref{appdx:datasets}. We consider both linear and nonlinear probabilistic models to tackle the regression and binary classification setups, which lead to predicting the parameters of a linear regression/classification model and a small nonlinear neural network based on \textsc{relu} activation function. Further, we also perform the analysis with a diagonal Gaussian assumption and a normalizing flow-based amortization model trained with both a forward and reverse KL objective. We provide the results on the regression problems in (a) linear model with diagonal Gaussian assumption (Figure~\ref{fig:regression_linear_vanilla}), (b) linear model with normalizing flow (Figure~\ref{fig:regression_linear_flow}), (c) nonlinear model with diagonal Gaussian assumption (Figure~\ref{fig:regression_nonlinear_vanilla}), and (d) nonlinear model with normalizing flow (Figure~\ref{fig:regression_nonlinear_flow}). The results of the classification problems are shown in (a) linear model with diagonal Gaussian assumption (Figure~\ref{fig:classification_linear_vanilla}), (b) linear model with normalizing flow (Figure~\ref{fig:classification_linear_flow}), (c) nonlinear model with diagonal Gaussian assumption (Figure~\ref{fig:classification_nonlinear_vanilla}), and (d) nonlinear model with normalizing flow (Figure~\ref{fig:classification_nonlinear_flow}).
Our experiments indicate that initializing with amortized models leads to better performance and training than models trained via maximum a-posteriori approach and initialized with the prior, i.e., $\gN(0, I)$. 

We do provide an additional baseline of initializing with \textsc{Xavier-init} initialization, which often leads to faster convergence; however, as we consider the prior to be a unit normal, this is an unfair baseline as we assume the weights to be initialized from a different prior. We leave the work of computing Bayesian posteriors with different priors and testing an amortized Bayesian model with \textsc{Xavier-init} prior for the future.

In addition to those experiments, we also conducted a broader range of experiments utilizing DeepSets as the backbone, various OoD data-generating functions for training and evaluation of the reverse KL system, and an additional nonlinear regression model with \textsc{relu} activation function. For a comprehensive description of these experiments and the complete setup, please refer to Section \ref{appdx:details_misspecification}.
We considered two probabilistic models, including a linear regression model and a nonlinear regression models utilizing the \textsc{tanh} activation function. The detailed results for each model can be found in Tables \ref{tab:misspec_model} and \ref{tab:misspec_gp}.

In all experiments, reverse KL outperforms forward KL trained amortization models in in-distribution performance and excels in posterior prediction on OoD datasets. 
Although the significant difference in posterior prediction performance of forward vs. reverse KL in cases where the underlying model is nonlinear was already mentioned in previous experiments, here, reverse KL-trained models also excel in evaluations of posterior prediction for the linear regression model.
Although only by a margin, in the case of approximating the posterior of the simpler linear regression model, a diagonal Gaussian-shaped posterior shows the best posterior prediction results when evaluated on OoD datasets from the nonlinear regression dataset generating function.
In almost all other experiments, the posterior prediction performance could be enhanced when we used the normalizing flow based posterior. 
A definitive conclusion cannot be drawn regarding the superiority of one backbone over the other, i.e. between DeepSets or Transformer. However, amortization models with DeepSets as the backbone tend towards better generalization regarding OoD datasets.

\end{document}